%% file: main.tex
\documentclass[10pt,twocolumn,letterpaper]{article}

\usepackage{cvpr}              

\input{preamble}

\definecolor{cvprblue}{rgb}{0.21,0.49,0.74}
\usepackage[pagebackref,breaklinks,colorlinks,citecolor=cvprblue]{hyperref}

\usepackage{tikz}
\usepackage{graphicx}
\usepackage{algorithm,algorithmicx,algpseudocode}
\usepackage{makecell}
\usepackage{multirow}

\newtoggle{hqfigures}
\togglefalse{hqfigures}  

\newtheorem{lemma}{Lemma}

\newcommand{\StartMenu}{\includegraphics[scale=0.04]{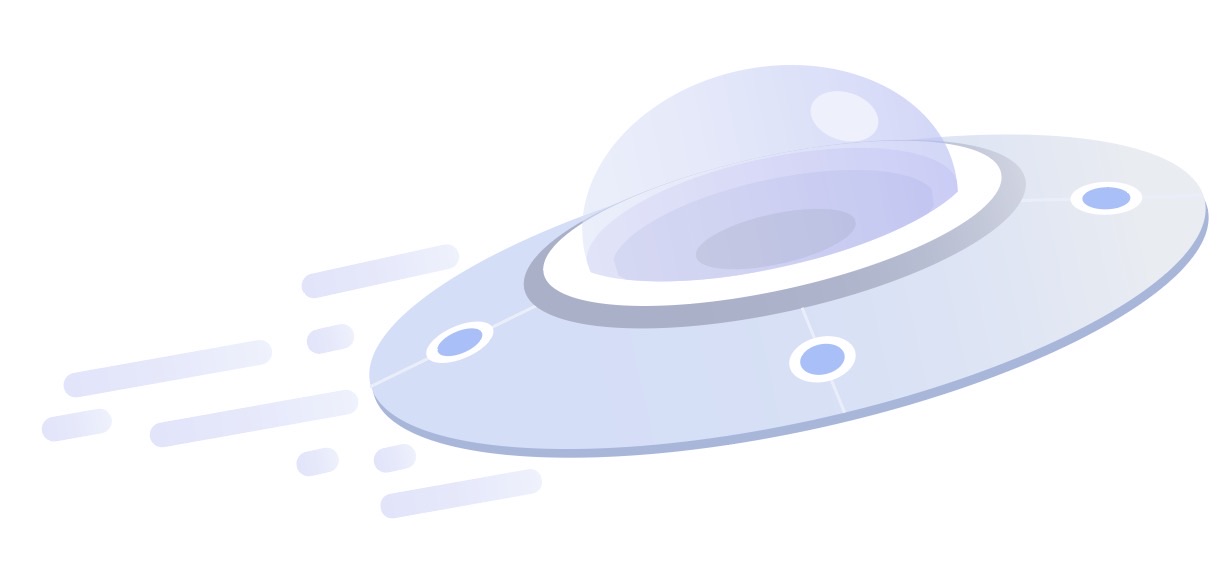}}%
\newcommand{\UFO}[1]{%
   \StartMenu
   \foreach \x in {#1} {%
   \texttt{\textbf{\x}}%
   }%
}%
\newcommand\blfootnote[1]{%
  \begingroup
  \renewcommand\thefootnote{}\footnote{#1}%
  \addtocounter{footnote}{-1}%
  \endgroup
}
\def\paperID{8504} 
\def\confName{CVPR}
\def\confYear{2024}

\title{\UFO{UFOGen}: You Forward Once Large Scale Text-to-Image Generation via Diffusion GANs}

\author{Yanwu Xu*$^{[1,2]}$\footnote[2]{}, Yang Zhao$^{[1]}$\footnote[2]{}, Zhisheng Xiao$^{[1]}$\footnote[2]{}, Tingbo Hou$^{[1]}$\\
\tt\small $^1$ Google \\
\tt\small \{yanwuxu,yzhaoeric,zsxiao,tingbo\}@google.com \\
\tt\small $^2$ Department of Electrical Computer Engineering, Boston University \\
\tt\small yanwuxu@bu.edu \\
}

\begin{document}
\maketitle

\input{sec/abstract}
\input{sec/intro_v2}
\input{sec/related_works_v1}

\input{sec/background_v1}

\input{sec/method_v3}
\input{sec/experiments}
\input{sec/conclusion}

\clearpage
{
    \small
    \bibliographystyle{ieeenat_fullname}
    \bibliography{main}
}
\onecolumn
\appendix
\input{sec/appendix}

\end{document}

%% file: preamble.tex
\usepackage[dvipsnames]{xcolor}


%% file: sec/abstract.tex
\begin{abstract}
Text-to-image diffusion models have demonstrated remarkable capabilities in transforming text prompts into coherent images, yet the computational cost of the multi-step inference remains a persistent challenge. To address this issue, we present UFOGen, a novel generative model designed for ultra-fast, one-step text-to-image generation. In contrast to conventional approaches that focus on improving samplers or employing distillation techniques for diffusion models, UFOGen adopts a hybrid methodology, integrating diffusion models with a GAN objective. Leveraging a newly introduced diffusion-GAN objective and initialization with pre-trained diffusion models, UFOGen excels in efficiently generating high-quality images conditioned on textual descriptions in a single step. Beyond traditional text-to-image generation, UFOGen showcases versatility in applications. Notably, UFOGen stands among the pioneering models enabling one-step text-to-image generation and diverse downstream tasks, presenting a significant advancement in the landscape of efficient generative models.
\blfootnote{*Work done as a student researcher of Google, $\dagger$ indicates equal contribution.}
\end{abstract}

%% file: sec/intro_v2.tex
\section{Introduction}

\begin{figure*}[!htp]
  \centering
    \includegraphics[width=0.9\textwidth]{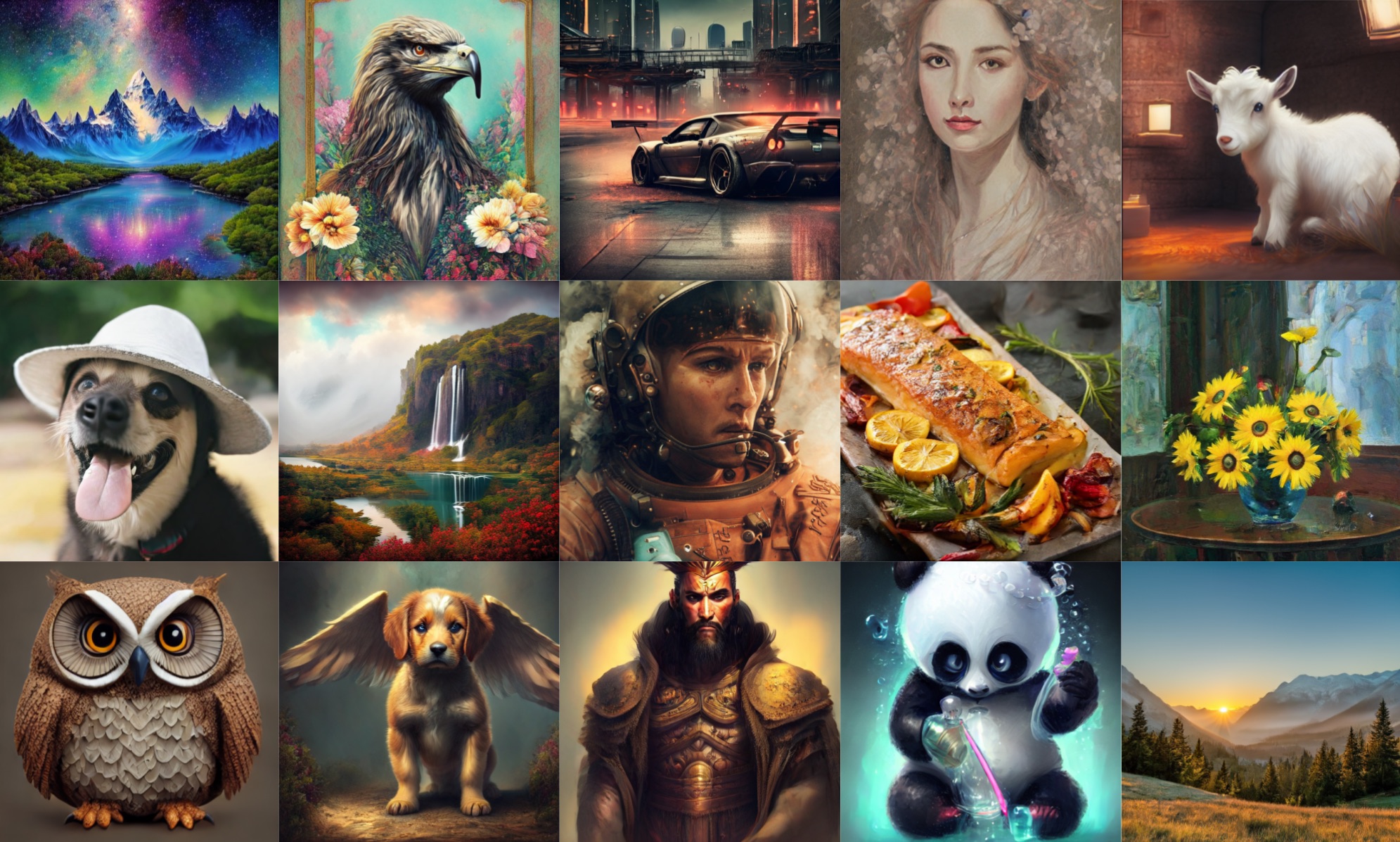}
  \caption{Images generated by our UFOGen Model with \textbf{1 sampling step}. The model is trained by fine-tuning Stable Diffusion 1.5 with our introduced techniques.}\label{fig: teaser}
\end{figure*}

Diffusion models~\cite{sohl2015deep, ho2020denoising, song2020score} has recently emerged as a powerful class of generative models, demonstrating unprecedented results in many generative modeling tasks~\cite{imagen, rombach2022high, kong2020diffwave, cai2020learning, huang2022riemannian, xu2021geodiff}. In particular, they have shown the remarkable ability to synthesize high-quality images conditioned on texts~\cite{balaji2022ediffi, ramesh2022hierarchical, imagen, nichol2022glide, rombach2022high, xue2023raphael}. Beyond the text-to-image synthesis tasks, large-scale text-to-image models serve as foundational building blocks for various downstream applications, including personalized generation~\cite{ruiz2023dreambooth, gal2022image, kumari2023multi, chen2023subject}, controlled generation~\cite{zhang2023adding, mou2023t2i} and image editing~\cite{hertz2022prompt, brooks2023instructpix2pix, yang2023paint}. Yet, despite their impressive generative quality and wide-ranging utility, diffusion models have a notable limitation: they rely on iterative denoising to generate final samples, which leads to slow generation speeds. The slow inference and the consequential computational demands of large-scale diffusion models pose significant impediments to their deployment.

In the seminal work by Song \etal~\cite{song2020score}, it was revealed that sampling from a diffusion model is equivalent to solving the probability flow ordinary differential equation (PF-ODE) associated with the diffusion process. Presently, the majority of research aimed at enhancing the sampling efficiency of diffusion models centers on the ODE formulation. One line of work seeks to advance numerical solvers for the PF-ODE, with the intention of enabling the solution of the ODE with greater discretization size, ultimately leading to fewer requisite sampling steps~\cite{song2020denoising, lu2022dpm, lu2023dpm,bao2022analytic}. However, the inherent trade-off between step size and accuracy still exists. Given the highly non-linear and complicated trajectory of the PF-ODE, it would be extremely difficult to reduce the number of required sampling steps to a minimal level. Even the most advanced solvers~\cite{lu2022dpm,lu2023dpm} can generate images within 10 to 20 sampling steps, and further reduction leads to a noticeable drop in image quality. An alternative approach seeks to distill the PF-ODE trajectory from a pre-trained diffusion model. For instance, progressive distillation~\cite{salimans2021progressive, meng2023distillation, li2023snapfusion} tries to condense multiple discretization steps of the PF-ODE solver into a single step by explicitly aligning with the solver's output. Similarly, consistency distillation~\cite{song2023consistency,luo2023latent} works on learning consistency mappings that preserve point consistency along the ODE trajectory. These methods have demonstrated the potential to significantly reduce the number of sampling steps. However, due to the intrinsic complexity of the ODE trajectory, they still struggle in the extremely small step regime, especially for large-scale text-to-image diffusion models. 

The pursuit of developing ultra-fast large-scale diffusion models that requires just one or two sampling steps, remains a challenging open problem. We assert that to achieve this ambitious objective, fundamental adjustments are necessary in the formulation of diffusion models, as the current ODE-based approach seems intrinsically constrained for very few steps sampling, as elucidated earlier. In this work, we introduce a novel one-step text-to-image generative model, representing a fusion of GAN and diffusion model elements. Our inspiration stems from previous work that successfully incorporated GANs into the framework of diffusion models~\cite{ssidms,ddgans,zheng2022truncated, wang2022diffusion}, which have demonstrated the capacity to generate images in as few as four steps when trained on small-scale datasets. These models diverge from the traditional ODE formulation by leveraging adversarial loss for learning the denoising distribution, rather than relying on KL minimization. Section \ref{sec_background} offers a comprehensive review of existing diffusion-GAN hybrid models. 

Despite the promising outcomes of earlier diffusion GAN hybrid models, achieving one-step sampling and extending their utility to text-to-image generation remains a non-trivial challenge. In this research, we introduce innovative techniques to enhance diffusion GAN models, resulting in an ultra-fast text-to-image model capable of producing high-quality images in a single sampling step. In light of this achievement, we have named our model \textbf{UFOGen}, an acronym denoting ``You Forward Once" Generative model. A detailed exposition of UFOGen is presented in Section~\ref{sec_method}. Our UFOGen model excels at generating high-quality images in just one inference step. Notably, when initialized with a pre-trained Stable Diffusion model~\cite{rombach2022high}, our method efficiently transforms Stable Diffusion into a one-step inference model while largely preserving the quality of generated content. See Figure~\ref{fig: teaser} for a showcase of text-conditioned images generated by UFOGen. To the best of our knowledge, our model stands among the pioneers to achieve a reduction in the number of required sampling steps for text-to-image diffusion models to just one.

Our work presents several significant contributions:
\begin{enumerate}
    \item We introduce UFOGen, a powerful generative model capable of producing high-quality images conditioned on text descriptions in a single inference step.
    \item We present an efficient and simplified training process, enabling the fine-tuning of pre-existing large-scale diffusion models, like Stable Diffusion, to operate as one-step generative models.
    \item Our model's versatility extends to applications such as image-to-image and controllable generation, thereby unlocking the potential for one-step inference across various generative scenarios.
\end{enumerate}

%% file: sec/related_works_v1.tex
\section{Related Works}

\vspace{2mm} \noindent \textbf{Text-to-image Diffusion Models} Denoising diffusion models~\cite{sohl2015deep, ho2020denoising, song2020score} are trained to reconstruct data from corrupted inputs. 
The simplicity of the training objective makes denoising diffusion models well-suited for scaling up generative models. Researchers have made numerous efforts to train diffusion models on large datasets containing image-text pairs~\cite{schuhmann2022laion} for the text-to-image generation task~\cite{nichol2022glide, ramesh2022hierarchical, imagen, balaji2022ediffi, rombach2022high,xue2023raphael}. Among these, latent diffusion models, such as the popular Stable Diffusion model~\cite{rombach2022high, podell2023sdxl}, have gained substantial attention in the research community due to their simplicity and efficiency compared to pixel-space counterparts. 

\vspace{2mm} \noindent \textbf{Accelerating Diffusion Models} The notable issue of slow generation speed has motivated considerable efforts towards enhancing the sampling efficiency of diffusion models. These endeavors can be categorized into two primary approaches. The first focuses on the development of improved numerical solvers ~\cite{song2020denoising,lu2022dpm, lu2023dpm,bao2022analytic,karras2022elucidating}. The second approach explores the concept of knowledge distillation~\cite{hinton2015distilling}, aiming at condensing the sampling trajectory of a numerical solver into fewer steps~\cite{salimans2021progressive, meng2023distillation, song2023consistency, berthelot2023tract, li2023snapfusion,luo2023latent}. However, both of these approaches come with significant limitations, and thus far, they have not demonstrated the ability to substantially reduce the sampling steps required for text-to-image diffusion models to a truly minimal level.

\vspace{2mm} \noindent \textbf{Text-to-image GANs} As our model has GAN~\cite{goodfellow2014generative} as one of its component, we provide a brief overview of previous attempts of training GANs for text-to-image generation. Early GAN-based text-to-image models were primarily confined to small-scale datasets~\cite{reed2016generative,zhang2017stackgan, xu2018attngan, tao2022df}. Later, with the evolution of more sophisticated GAN architectures~\cite{karnewar2020msg, karras2020analyzing, sauer2022stylegan}, GANs trained on large datasets have shown promising results in the domain of text-to-image generation~\cite{zhou2022towards, sauer2023stylegan, kang2023gigagan}. Comparatively, our model has several distinct advantages. Firstly, to overcome the well-known issues of training instability and mode collapse, text-to-image GANs have to incorporate multiple auxiliary losses and complex regularization techniques, which makes training and parameter tuning extremely intricate. This complexity is particularly exemplified by GigaGAN~\cite{kang2023gigagan}, currently regarded as the most powerful GAN-based models. In contrast, our model offers a streamlined and robust training process, thanks to the diffusion component. Secondly, our model's design allows us to seamlessly harness pre-trained diffusion models for initialization, significantly enhancing the efficiency of the training process. Lastly, our model exhibits greater flexibility when it comes to downstream applications (see Section \ref{sec exp app}), an area in which GAN-based models have not explored.

\vspace{2mm} \noindent \textbf{Recent Progress on Few-step Text-to-image Generation} While developing our model, we noticed some concurrent work on few-step text-to-image generation. Latent Consistency Model~\cite{luo2023latent} extends the idea of consistency distillation~\cite{song2023consistency} to Stable Diffusion, leading to 4-step sampling with reasonable quality. However, further reducing the sampling step results in significant quality drop. InstaFlow~\cite{liu2023insta} achieves text-to-image generation in a single sampling step. Similar to our model, InstaFlow tackles the slow sampling issue of diffusion models by introducing improvements to the model itself. Notably, they extend Rectified Flow models~\cite{liu2022flow,liu2022rectified} to create a more direct trajectory in the diffusion process. In direct comparison to InstaFlow, our model outperforms in terms of both quantitative metrics and visual quality. Moreover, our approach presents the added benefits of a streamlined training pipeline and improved training efficiency. InstaFlow requires multiple stages of fine-tuning, followed by a subsequent distillation stage. In contrast, our model only need one single fine-tuning stage with a minimal number of training iterations.

%% file: sec/background_v1.tex
\section{Background} \label{sec_background}
\vspace{2mm} \noindent \textbf{Diffusion Models} 
Diffusion models~\cite{sohl2015deep, ho2020denoising} is a family of generative models that progressively inject Gaussian noises into the data, and then generate samples from noise via a reverse denoising process. Diffusion models define a forward process that corrupts data $x_0 \sim q(x_0)$ in $T$ steps with variance schedule $\beta_t$: $q(x_t|x_{t-1}) := \mathcal{N}(x_t; \sqrt{1-\beta_t}x_{t-1}, \beta_t\textbf{I})$. The parameterized reversed diffusion process aims to gradually recover cleaner data from noisy observations: $p_{\theta}(x_{t-1}|x_t) := \mathcal{N}(x_{t-1}; \mu_\theta(x_t, t), \sigma_t^2\textbf{I})$.

The model $p_\theta(x_{t-1}|x_t)$ is parameterized as a Gaussian distribution, because when the denoising step size from $t$ to $t-1$ is sufficiently small, the true denoising distribution $q(x_{t-1}|x_t)$ is a Gaussian \cite{feller2015retracted}. To train the model, one can minimize the negative ELBO objective ~\cite{ho2020denoising, kingma2021variational}:
\begin{align}\label{eq:ddpm_matching}
    \mathcal{L}=\mathbb{E}_{t,q(x_0)q(x_t|x_0)}\text{KL}(q(x_{t-1}|x_t, x_0)||p_\theta(x_{t-1}|x_t)), 
\end{align}
where $q(x_{t-1}|x_t,x_0)$ is Gaussian posterior distribution derived in~\cite{ho2020denoising}. 


\vspace{2mm} \noindent \textbf{Diffusion-GAN Hybrids}
The idea of combining diffusion models and GANs is first explored in~\cite{ddgans}. The main motivation is that, when the denoising step size is large, the true denoising distribution $q(x_{t-1}|x_t)$ is no longer a Gaussian. Therefore, instead of minimizing KL divergence with a parameterized Gaussian distribution, they parameterized $p_{\theta}(x_{t-1}'|x_t)$ as a conditional GAN to minimize the adversarial divergence between model $p_{\theta}(x_{t-1}'|x_t)$ and $q(x_{t-1}|x_t)$:
\begin{align}\label{eq:ddgan_jsd}
\min_\theta\mathbb{E}_{q(x_t)}\Bigl[D_{adv}(q(x_{t-1}|x_t)||p_\theta(x_{t-1}'|x_t))\Bigr].
\end{align}
The objective of Denoising Diffusion GAN (DDGAN) in ~\cite{ddgans} can be expressed as:
\begin{align}\label{eq:ddgan}
    \min_\theta&\max_{D_\phi} \mathbb{E}_{q(x_t)}\Bigl[\mathbb{E}_{q(x_{t-1}|x_t)}[-\log(D_{\phi}(x_{t-1}, x_t, t))] \nonumber \\
    &+ \mathbb{E}_{p_{\theta}(x_{t-1}'|x_t)}[-\log(1-D_{\phi}(x_{t-1}', x_t, t))] \Bigr],
\end{align}
where $D_{\phi}$ is the conditional discriminator network, and the expectation over the unknown distribution $q(x_{t-1}|x_t)$ can be approximated by sampling from $q(x_0)q(x_{t-1}|x_0)q(x_t|x_{t-1})$. The flexibility of a GAN-based denoising distribution surpasses that of a Gaussian parameterization, enabling more aggressive denoising step sizes. Consequently, DDGAN successfully achieves a reduction in the required sampling steps to just four.

Nonetheless, the utilization of a purely adversarial objective in DDGAN introduces training instability, as documented by the findings in ~\cite{ssidms}. In response to this challenge, the authors in ~\cite{ssidms} advocated matching the joint distribution $q(x_{t-1}, x_t)$ and $p_{\theta}(x_{t-1}, x_t)$, as opposed to the conditional distribution as outlined in Equation \ref{eq:ddgan_jsd}. ~\cite{ssidms} further demonstrated that the joint distribution matching can be disassembled into two components: matching marginal distributions using adversarial divergence and matching conditional distributions using KL divergence:
\begin{align}\label{eq:siddms_match}
\min_\theta & \mathbb{E}_{q(x_t)} \Bigl[D_{adv}(q(x_{t-1})||p_\theta(x_{t-1})) \nonumber\\
&+ \lambda_{KL}\text{KL}(p_\theta(x_t|x_{t-1})||q(x_t|x_{t-1}))\Bigr].
\end{align}
The objective of adversarial divergence minimization is similar to Equation \ref{eq:ddgan} except that the discriminator does not take $x_t$ as part of its input. The KL divergence minimization translates into a straightforward reconstruction objective, facilitated by the Gaussian nature of the diffusion process (see Appendix \ref{app:kl derivation} for a derivation). This introduction of a reconstruction objective plays a pivotal role in enhancing the stability of the training dynamics. As observed in ~\cite{ssidms}, which introduced Semi-Implicit Denoising Diffusion Models (SIDDMs), this approach led to markedly improved results, especially on more intricate datasets.

%% file: sec/method_v3.tex
\section{Methods} \label{sec_method}
In this section, we present a comprehensive overview of the enhancements we have made in our diffusion-GAN hybrid models, ultimately giving rise to the UFOGen model. These improvements are primarily focused on two critical domains: 1) enabling one step sampling, as detailed in Section \ref{sec_method one step}, and 2) scaling-up for text-to-image generation, as discussed in Section \ref{sec_method scale up}.  

\subsection{Enabling One-step Sampling for UFOGen} \label{sec_method one step}
Diffusion-GAN hybrid models are tailored for training with a large denoising step size. However, attempting to train these models with just a single denoising step (i.e., $x_{T-1} = x_0$) effectively reduces the training to that of a conventional GAN. Consequently, prior diffusion-GAN models were unable to achieve one-step sampling. In light of this challenge, we conducted an in-depth examination of the SIDDM ~\cite{ssidms} formulation and implemented specific modifications in the generator parameterization and the reconstruction term within the objective. These adaptations enabled UFOGen to perform one-step sampling, while retaining training with several denoising steps.

\vspace{2mm} \noindent \textbf{Parameterization of the Generator}\label{sec generator param}
In diffusion-GAN models, the generator should produce a sample of $x_{t-1}$. However, instead of directly outputting $x_{t-1}$, the generator of DDGAN and SIDDM is parameterized by $p_{\theta}(x_{t-1}|x_t) = q(x_{t-1}|x_t, x_0=G_{\theta}(x_t, t))$. In other words, first $x_0$ is predicted using the denoising generator $G_{\theta}(x_t, t)$, and then, $x_{t-1}$ is sampled using the Gaussian posterior distribution $q(x_{t-1}|x_t, x_0)$ derived in ~\cite{ho2020denoising, ddgans}. Note that this parameterization is mainly for practical purposes, as discussed in ~\cite{ddgans}, and alternative parameterization would not break the model formulation. 

We propose another plausible parameterization for the generator: $p_{\theta}(x_{t-1}) = q(x_{t-1}|x_0=G_{\theta}(x_t, t))$. The generator still predicts $x_0$, but we sample $x_{t-1}$ from the forward diffusion process $q(x_{t-1}|x_0)$ instead of the posterior. As we will show later, this design allows distribution matching at $x_0$, paving the path to one-step sampling. 

\vspace{2mm} \noindent \textbf{Improved Reconstruction Loss at $x_0$} \label{sec improved rec}
We argue that with the new generator parameterization, the objective of SIDDM in Equation \ref{eq:siddms_match} indirectly matches the distribution at $x_0$. To see this, we analyze the adversarial objective and KL objective in Equation \ref{eq:siddms_match} separately. The first term minimizes adversarial divergence $D_{adv}\left(q(x_{t-1})||p_\theta(x_{t-1}')\right)$, where $q(x_{t-1})$ and $p_{\theta}(x_{t-1}')$ can both be seen as the corruption of a distribution at $x_0$ by the same Gaussian kernel. Specifically, since $q(x_{t-1}) = \mathbb{E}_{q(x_0)}[q(x_{t-1}|x_0)]$, given a sample $x_0 \sim q(x_0)$, we have $q(x_t) = \mathcal{N}(x_{t-1};\sqrt{\Bar{\alpha}_{t-1}}x_0, (1-\Bar{\alpha}_{t-1})\textbf{I})$, according to the forward diffusion formulation \cite{ho2020denoising}. Similarly, $p_{\theta}(x_{t-1}')$ has the same form except that $x_0$ is produced by the generator. As a result, adversarial distribution matching on $q(x_{t-1})$ and $p_{\theta}(x_{t-1}')$ will also encourage the matching between $q(x_0)$ and $p_{\theta}(x_0')$, which is the distribution over $x_0$ produced by the generator. A formal explanation will be presented in Appendix~\ref{app adv equivalent}. 

The second term in the objective minimizes the KL divergence between $p_\theta(x_t|x_{t-1}')$ and $q(x_t|x_{t-1})$, which, as derived in Appendix \ref{app:kl derivation}, can be simplified to the following reconstruction term:
\begin{align}\label{eq: recon xt-1}
    \mathbb{E}_{q(x_t)}\Bigl[\frac{(1-\beta_t)||x_{t-1}' - x_{t-1}||^2}{2\beta_t}\Bigr]. 
\end{align}
Based on above analysis on $x_{t-1}'$ and $x_{t-1}$, it is easy to see that minimizing this reconstruction loss will essentially matches $x_0$ and $x_0'$ as well (a straightforward derivation is provided in Appendix~\ref{app recon equivalent}). 
\begin{figure}[tp!]
  \centering
    \includegraphics[width=\columnwidth]{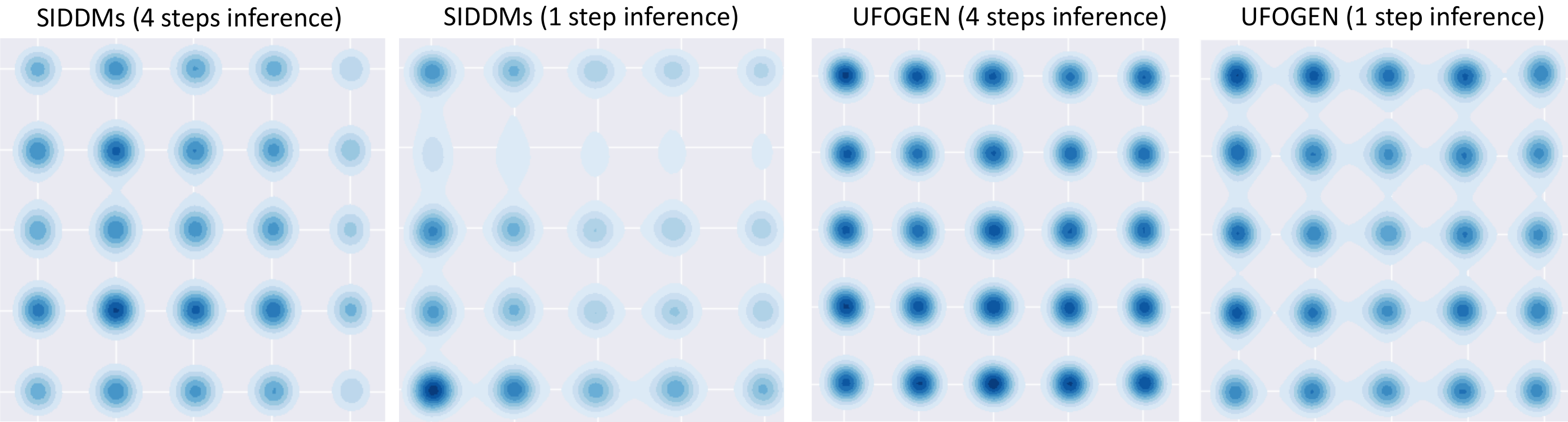}
  \caption{Results of training with UFOGen loss versus the original loss of SIDDM on 25-Gaussian toy data. With the modified objective, UFO enables one-step sampling.}\label{fig 2dtoy}
\end{figure}

Per our analysis, both terms in the SIDDM objective in Equation \ref{eq:siddms_match} implicitly matches the distribution at $x_0$, which suggests that one-step sampling is possible. However, empirically we observe that one-step sampling from SIDDM does not work well even on 2-D toy dataset (See Figure \ref{fig 2dtoy}). We conjecture that this is due to the variance introduced in the additive Gaussian noise when sampling $x_{t-1}$ with $x_0$. To reduce the variance, we propose to replace the reconstruction term in Equation \ref{eq: recon xt-1} with the reconstruction at clean sample $||x_0 - x_0'||^2$, so that the matching at $x_0$ becomes explicit. We observe that with this change, we can obtain samples in one step, as shown in Figure \ref{fig 2dtoy}.

\vspace{2mm} \noindent \textbf{Training and Sampling of UFOGen}
To put things together, we present the complete training objective and strategy for the UFOGen model. UFOGen is trained with the following objective:
\begin{align}\label{eq:ufo obj}
    &\min_\theta\max_{D_\phi}\mathbb{E}_{q(x_0)q(x_{t-1}|x_0), p_{\theta}(x_{0}')p_{\theta}(x_{t-1}'|x_0')} \Bigl[ \nonumber\\
    &[\log(D_{\phi}(x_{t-1}, t))]+ [\log(1-D_{\phi}(x'_{t-1}, t))]  \nonumber\\
    &+ \lambda_{KL}\gamma_t\left\lVert x_0-x_0'\right\rVert^2 \Bigl],
\end{align}
where $\gamma_t$ is a time-dependent coefficient. The objective consists of an adversarial loss to match noisy samples at time step $t-1$, and a reconstruction loss at time step $0$. Note that the reconstruction term is essentially the training objective of diffusion models ~\cite{ho2020denoising, song2020score}, and therefore the training of UFOGen model can also be interpreted as training a diffusion model with adversarial refinement. The training scheme of UFOGen is presented in Algorithm \ref{alg:training}.

Despite the straightforward nature of the modifications to the training objective, these enhancements have yielded impressive outcomes, particularly evident in the context of one-step sampling, where we simply sample $x_{T}\sim \mathcal{N}(0, \textbf{I})$ and produce sample $x_0'=G_{\theta}(x_T)$.

\begin{figure}[tp]
  \centering
    \includegraphics[width=0.48\textwidth]{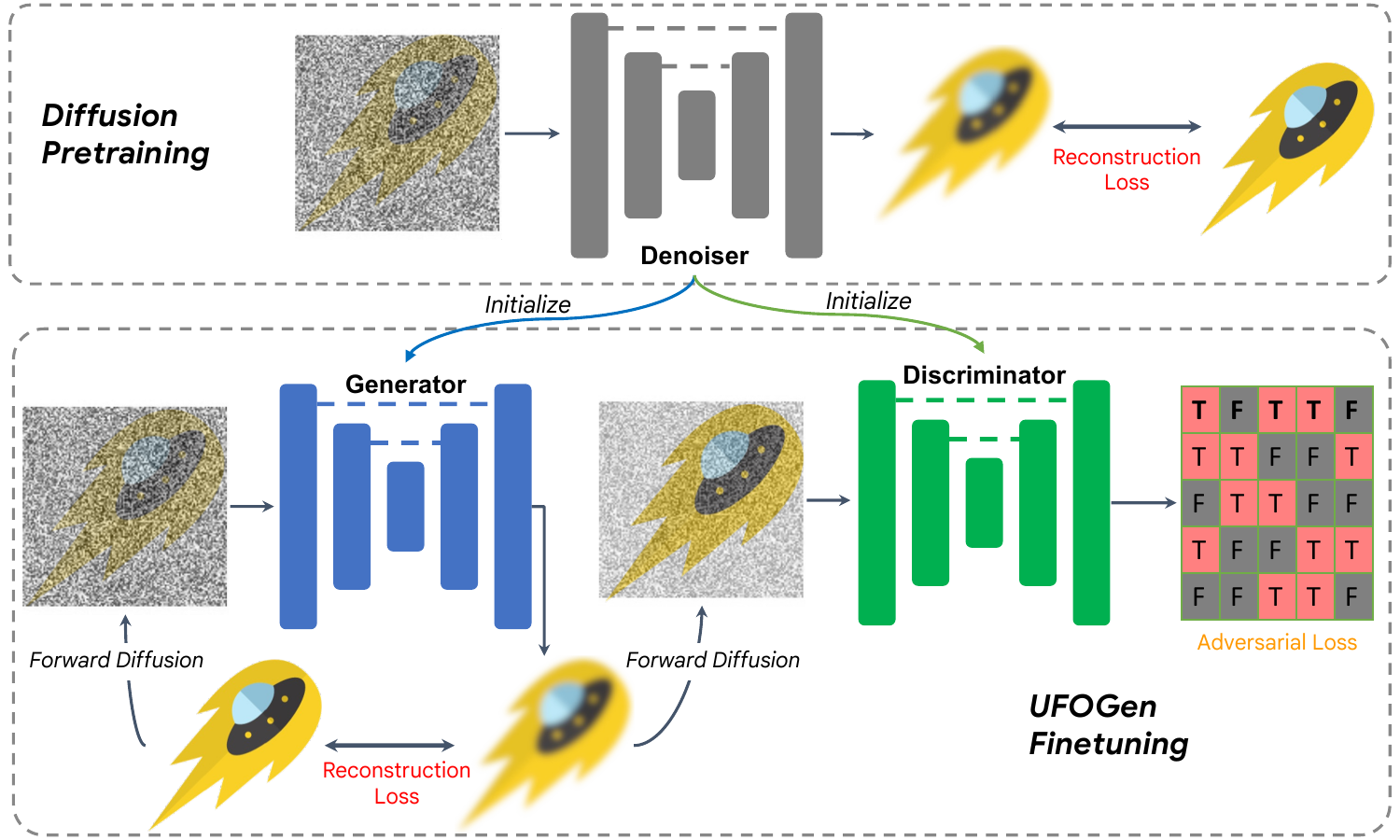}
  \caption{Illustration of the training strategy for UFOGen model.}
\label{fig:ufo_scheme}
\end{figure}

\subsection{Leverage Pre-trained Diffusion Models} \label{sec_method scale up}
Our objective is developing an ultra-fast text-to-image model. However, the transition from an effective UFOGen recipe to web-scale data presents considerable challenges. Training diffusion-GAN hybrid models for text-to-image generation encounters several intricacies. Notably, the discriminator must make judgments based on both texture and semantics, which govern text-image alignment. This challenge is particularly pronounced during the initial stage of training. Moreover, the cost of training text-to-image models can be extremely high, particularly in the case of GAN-based models, where the discriminator introduces additional parameters. Purely GAN-based text-to-image models~\cite{sauer2023stylegan, kang2023gigagan} confront similar complexities, resulting in highly intricate and expensive training.

To surmount the challenges of scaling-up diffusion-GAN hybrid models, we propose the utilization of pre-trained text-to-image diffusion models, notably the Stable Diffusion model~\cite{rombach2022high}. Specifically, our UFOGen model is designed to employ a consistent UNet structure for both its generator and discriminator. This design enables seamless initialization with the pre-trained Stable Diffusion model. We posit that the internal features within the Stable Diffusion model contain rich information of the intricate interplay between textual and visual data. This initialization strategy significantly streamlines the training of UFOGen. Upon initializing UFOGen's generator and discriminator with the Stable Diffusion model, we observe stable training dynamics and remarkably fast convergence. The complete training strategy of UFOGen is illustrated in Figure~\ref{fig:ufo_scheme}.

\algrenewcommand\algorithmicindent{0.5em}%
\begin{figure}[t]
\begin{minipage}[t]{0.495\textwidth}
\begin{algorithm}[H]
  \caption{UFOGen Training} \label{alg:training}
  \small
  \begin{algorithmic}[1]
   \Require Generator $G_{\theta}$, discriminator $D_{\phi}$, loss coefficient $\lambda_{KL}$
    \Repeat
      \State Sample $x_0\sim q(x_0), t-1\sim \text{Uniform}({0, . . . , T-1})$.
      \State Sample $x_{t-1}\sim q(x_{t-1}|x_0), x_t\sim q(x_t|x_{t-1})$
      \State Sample $ x_{t-1}' \sim q(x_{t-1}|x_0')$, where $ x_0'=G_\theta(x_t,t)$
      \State Update $D_{\phi}$ with gradient 
      
      \noindent$\nabla_{\phi} \left(-\log\left(D_\phi(x_{t-1},t-1)\right)-\log\left(1-D_\phi(x_{t-1}',t-1)\right)\right) $
      \State Update $G_{\theta}$ with gradient 
      
      \noindent$\nabla_\theta \left(-\log(D_\phi(x_{t-1}',t-1) + \lambda_{KL}\gamma_t||x_0-x_0'||_2^2\right)$
    \Until{converged}
  \end{algorithmic}
\end{algorithm}
\end{minipage}
\hfill
\end{figure}

%% file: sec/experiments.tex
\section{Experiments}
In this section, we evaluate our proposed UFOGen model for the text-to-image synthesis problem. In Section \ref{sec exp t2i}, we start with briefly introducing our experimental setup, followed by comprehensive evaluations of UFOGen model on the text-to-image task, both quantitatively and qualitatively. We conduct ablation studies in Section \ref{sec exp ablation}, highlighting the effectiveness of our modifications introduced in Section \ref{sec_method}. In Section \ref{sec exp app}, we present qualitative results for downstream applications of UFOGen. 

\subsection{Text-to-image Generation}\label{sec exp t2i}
\vspace{2mm} \noindent \textbf{Configuration for Training and Evaluation} For experiments on text-to-image generation, we follow the scheme proposed in Section \ref{sec_method scale up} to initialize both the generator and discriminator with the pre-trained Stable Diffusion 1.5\footnote{\url{https://huggingface.co/runwayml/stable-diffusion-v1-5}} model ~\cite{rombach2022high}. We train our model on the LAION-Aesthetics-6+ subset of LAION-5B \cite{schuhmann2022laion}. More training details are provided in Appendix \ref{app exp details}. For evaluation, we adopt the common practice that uses zero-shot FID \cite{heusel2017gans} on MS-COCO ~\cite{lin2014microsoft}, and CLIP score with ViT-g/14 backbone \cite{radford2021learning}. 

\vspace{2mm} \noindent \textbf{Main Results} To kick-start our evaluation, we perform a comparative analysis in Table \ref{tab:full-5k}, bench-marking UFOGen against other few-step sampling models that share the same Stable Diffusion backbone. Our baselines include Progressive Distillation \cite{meng2023distillation} and its variant \cite{li2023snapfusion}, which are previously the state-of-the-art for few-step sampling of SD, as well as the concurrent work of InstaFlow \cite{liu2023insta}. Latent Consistency Model (LCM) \cite{luo2023latent} is excluded, as the metric is not provided in their paper. Analysis of the results presented in Table \ref{tab:full-5k} reveals the superior performance of our single-step UFOGen when compared to Progressive Distillation across one, two, or four sampling steps, as well as the CFG-Aware distillation \cite{li2023snapfusion} in eight steps. Furthermore, our method demonstrates advantages in terms of both FID and CLIP scores over the single-step competitor, InstaFlow-0.9B, which share the same network structure of SD with us. Impressively, our approach remains highly competitive even when compared to InstaFlow-1.7B with stacked UNet structures, which effectively doubles the parameter count.

\begin{table}[t]
\scriptsize
    \centering
    \begin{tabular}{lcccc}
    \toprule
    Method  & \#Steps & Time (s) & FID-5k & CLIP \\ \midrule
    \multirow{2}{*}{DPM Solver~\cite{lu2022dpm}} & 25 & 0.88  & \textbf{20.1}  & 0.318  \\ 
     & 8 & 0.34  &  31.7 & \textbf{0.320}  \\ 
    \midrule
    \multirow{3}{*}{Progressive Distillation~\cite{meng2023distillation}} &1 & 0.09 & 37.2 & 0.275 \\ 
    & 2 & 0.13  & 26.0 & 0.297  \\
    & 4 & 0.21  & 26.4 & 0.300 \\  \midrule
    CFG-Aware Distillation~\cite{li2023snapfusion} & 8& 0.34 & 24.2 & 0.30 \\
    \midrule 
    InstaFlow-0.9B & 1 & 0.09 & 23.4  &0.304  \\
    InstaFlow-1.7B & 1 & 0.12 & \textbf{22.4}  & 0.309 \\
    \midrule
    \textbf{UFOGen} & 1 & 0.09  & 22.5  & \textbf{0.311} \\
    \bottomrule
    \end{tabular}
    \caption{Comparison of FID on MSCOCO-2017 5k and CLIP score. All models are based on SD. Numbers of progressive distillation and InstaFlow are cited from \cite{liu2023insta}.
    }
    \label{tab:full-5k}
\end{table}

\input{sec/figure_tables/main_compare}

The results depicted in Table \ref{tab:full-5k} may suggest that InstaFlow remains a strong contender in one-step generation alongside UFOGen. However, we argue that relying solely on the MS-COCO zero-shot FID score for evaluating visual quality might not be the most reliable metric, a concern highlighted in prior research such as \cite{podell2023sdxl,kirstain2023pick} and discussed by \cite{betzalel2022study}. 
Consequently, we believe that qualitative assessments can provide more comprehensive insights. We present qualitative comparisons involving InstaFlow and LCM\footnote{InstaFlow (\url{https://huggingface.co/spaces/XCLiu/InstaFlow}) and LCM (\url{https://huggingface.co/spaces/SimianLuo/Latent_Consistency_Model})} in Table \ref{tab:main_comparison}. The comparisons allow for a clear-cut conclusion: UFOGen's one-step image generation surpasses InstaFlow by a substantial margin in terms of image quality. Notably, UFOGen also demonstrates significant advantages when contrasted with the 2-step LCM, as showed by the evident blurriness present in LCM's samples. Furthermore, even when compared to the samples generated by the 4-step LCM, our generated images exhibit distinct characteristics, including sharper textures and finer details. We do not present results of single-step LCM, as we observe that it fail to generate any textures (see Appendix \ref{app failed lcm-1}). Additional examples of the comparison are provided in Appendix \ref{app extra comparison}, where we display multiple images generated by each model for different prompts. We provide additional qualitative samples of UFOGen in Appendix \ref{app extra t2i results}.

\begin{table}[t]
\scriptsize
    \centering
    \begin{tabular}{lcccl}
    \toprule
     Method & Type & Time (s) & \# Param. & FID-30k \\ \midrule
     DALLE~\cite{ramesh2021zero} & AR  &- & 12B & 27.5 \\ 
    Parti-20B~\cite{yu2022scaling} & AR  &- & 20B & 7.23 \\ 
    Make-A-Scene~\cite{gafni2022make} & AR  &25.0 & - & 11.84 \\ \midrule
    GLIDE~\cite{nichol2022glide} &  Diff  &  15.0 & 5B & 12.24 \\
    DALLE 2~\cite{ramesh2022hierarchical} & Diff  &  - & 5.5B & 10.39 \\
    Imagen~\cite{ho2022imagen} &  Diff  & 9.1 & 3B & 7.27 \\
    eDiff-I~\cite{balaji2022ediffi} &  Diff  & 32.0 & 9B & 6.95 \\
    SD~\cite{rombach2022high} &  Diff  & 2.9 & 0.9B & 9.62 \\ \midrule
    LAFITE~\cite{zhou2022towards} & GAN  & 0.02 & 75M & 26.94 \\    
    StyleGAN-T~\cite{sauer2023stylegan} & GAN  & 0.10 & 1B & 13.90 \\
    GigaGAN~\cite{kang2023scaling} & GAN  & 0.13 & 1B & 9.09 \\ \midrule
    Muse-3B~\cite{chang2023muse} & - &  1.3 & 3B & 7.88 \\    
    InstaFlow~\cite{liu2023insta}& - &  0.09&0.9B&13.10\\
    \midrule
    \textbf{UFOGen (Ours)} & -  & 0.09 & 0.9B & 12.78 \\
    \bottomrule
    \end{tabular}
    \caption{Comparison of FID on MSCOCO 2014 with 30k images. Numbers of other models are cited from \cite{liu2023insta}. Inference time measurement follows the setting of \cite{gigagan}.}
    \label{tab:full-30k}
\end{table}

For completeness, we extend our comparison to encompass a diverse array of text-to-image generative models in Table \ref{tab:full-30k}. While the results in Table \ref{tab:full-30k} are not directly comparable due to substantial variations in model architecture, parameter count, and training data, it is noteworthy that UFOGen is a competitive contender among the contemporary landscape of text-to-image models, offering the advantage of remarkable speed over auto-regressive or diffusion models, thanks to its inherent one-step generation capability.

Based on both quantitative and qualitative assessments, we assert that UFOGen stands as a powerful text-to-image generative model, capable of producing sharp and visually appealing images that align well with the provided text conditioning, all in a single step. Our evaluation underscores its capacity to produce superior sample quality when contrasted with competing diffusion-based methods designed for a few-step generation process. 


\subsection{Ablation Studies}\label{sec exp ablation}
Ablation studies have been conducted to offer deeper insights into the effectiveness of our training strategies. As outlined in Table~\ref{tab:main_ablation}, we compare the training of diffusion-GAN hybrid models using the SIDDM objective \cite{ssidms} against the proposed UFOGen objective in Section \ref{sec_method one step}. The results validate our assertions, demonstrating that the modifications in the UFOGen objective facilitate one-step sampling. We additionally provide qualitative samples, and an supplementary ablation study on the denoising step size during training in Appendix \ref{app extra ablation}.

\begin{table}[t]
\scriptsize
    \centering
    \begin{tabular}{lccc}
    \toprule
    Method & \#Steps  & FID-5k & CLIP \\ \midrule
    \multirow{2}{*}{SIDDM~\cite{ssidms}} & 4 & 21.7 & 0.306 \\
      & 1  & 28.0 & 0.289 \\\midrule
    \multirow{2}{*}{UFOGen} & 4 & 22.1  & 0.307 \\
    & 1 & 22.5 & 0.311 \\
    \bottomrule
    \end{tabular}
    \caption{Ablation study comparing the SIDDM objective with our UFOGen objective, incorporating the introduced modifications detailed in Section \ref{sec_method one step}.
    }
    \label{tab:main_ablation}
\end{table}

\subsection{Applications}\label{sec exp app}
A promising aspect of text-to-image diffusion models is their versatility as foundational components for various applications, whether fine-tuned or utilized as is. In this section, we showcase UFOGen's ability to extend beyond text-to-image generation, while benefiting from its unique advantage of single-step generation. Specifically, we explore two applications of UFOGen: image-to-image \cite{meng2021sdedit} generation and controllable generation \cite{zhang2023adding, mou2023t2i}.
\input{sec/figure_tables/main_i2i}

Table \ref{tab:main_i2i} showcases UFOGen's image-to-image generation outcomes. Following SDEdit \cite{meng2021sdedit}, we introduce a suitable amount of noise to the input data, and let UFOGen to execute single-step generation based on the given prompt. Our observations affirm that UFOGen adeptly produces samples that adhere to the specified conditions of both the prompt and the input image. 

To facilitate controllable generation, we conduct fine-tuning of UFOGen by incorporating an additional adapter network, akin to the approach outlined in \cite{mou2023t2i}. This adapter network takes control signals as input to guide the generation process. In our exploration, we employ two types of control signals: depth maps and canny edges. The results are presented in Table \ref{tab:main_control}. Post fine-tuning, UFOGen exhibits the ability to generate high-quality samples that align with both the provided prompt and control signal.

Our results highlight UFOGen can work on diverse generation tasks in a single step, a distinctive feature that, to the best of our knowledge, sets our model apart. Unlike GAN-based text-to-image models \cite{sauer2023stylegan,kang2023gigagan}, which lack the ability to handle zero-shot image-to-image generation tasks as they do not generate samples through denoising, UFOGen excels in this context. Moreover, our model succeeds in controllable generation, a domain that earlier GAN-based models have not explored due to the complexities of fine-tuning and adding supplementary modules to the StyleGAN architecture. Consequently, the flexibility of our model in addressing various downstream tasks positions it uniquely among one-step text-to-image models. Additional results of the applications are provided in Appendix \ref{app extra application}.

%% file: sec/figure_tables/main_compare.tex
\setlength{\tabcolsep}{3pt} 
\renewcommand{\arraystretch}{1.0} 
\begin{table*}[!t]
    \centering
    \footnotesize
    \scalebox{1.0}{
    \begin{tabular}{ccccc}
        \textbf{SD (50 steps)} & \textbf{InstaFlow (1 step)} & \textbf{LCM (2 steps)} & \textbf{LCM (4 steps)} & \textbf{UFOGen (1 step)}\\
        \includegraphics[width=0.16\textwidth]{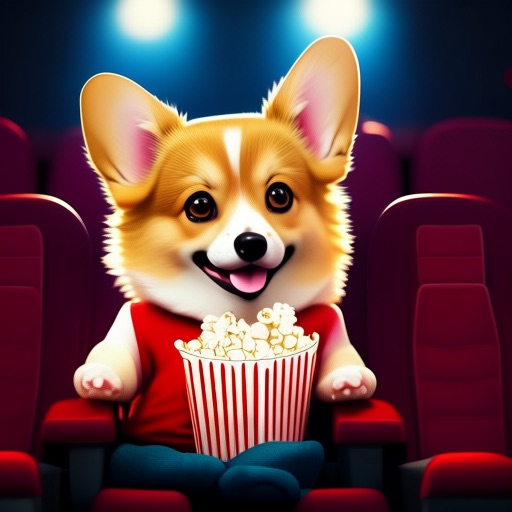} &  
        \includegraphics[width=0.16\textwidth]{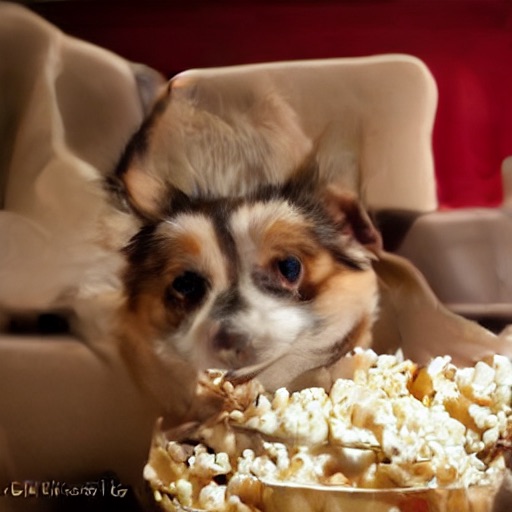} &
        \includegraphics[width=0.16\textwidth]{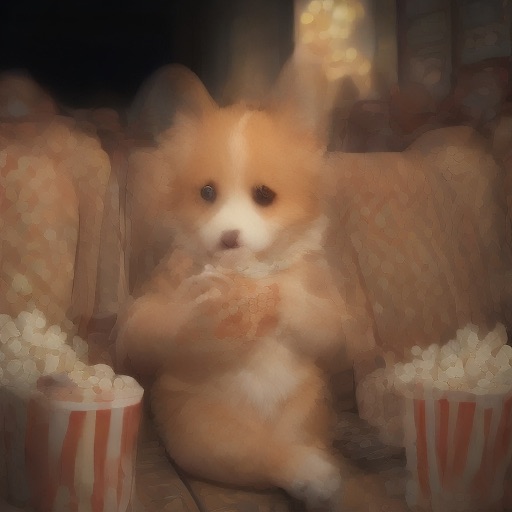} &
        \includegraphics[width=0.16\textwidth]{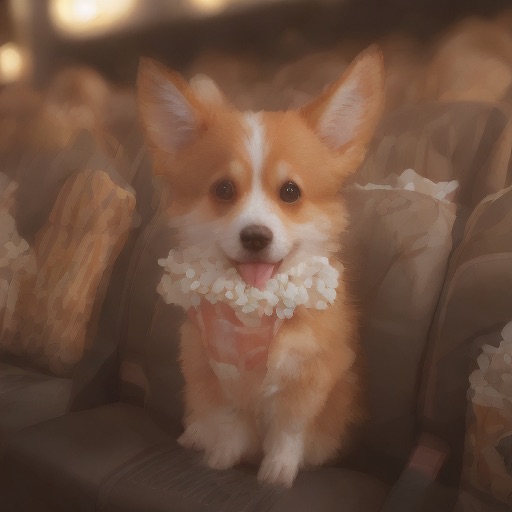} &  
        \includegraphics[width=0.16\textwidth]{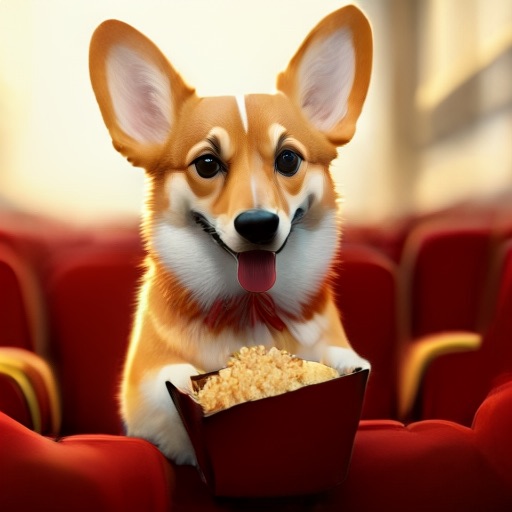} \\  
        \multicolumn{5}{c}{\textit{Cute small corgi sitting in a movie theater eating popcorn, unreal engine.}} \\

        \includegraphics[width=0.16\textwidth]{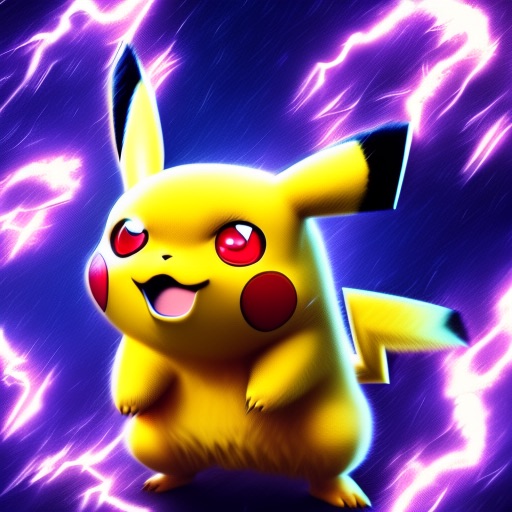} &  
        \includegraphics[width=0.16\textwidth]{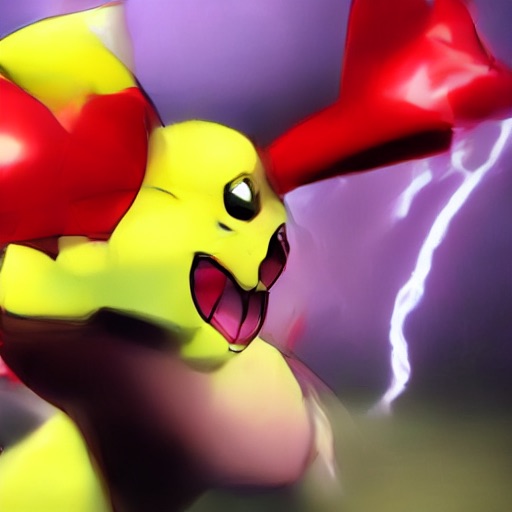} &
        \includegraphics[width=0.16\textwidth]{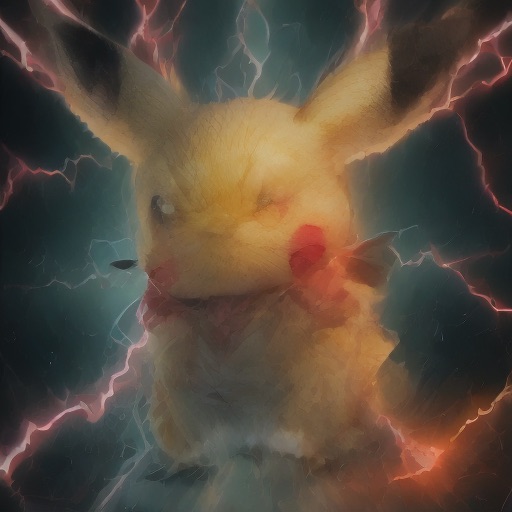} &
        \includegraphics[width=0.16\textwidth]{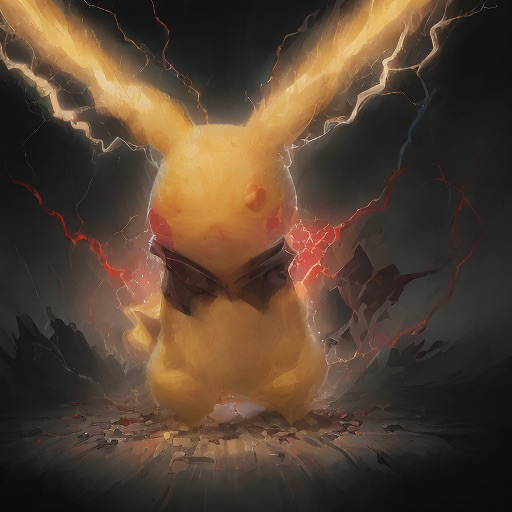} &
        \includegraphics[width=0.16\textwidth]{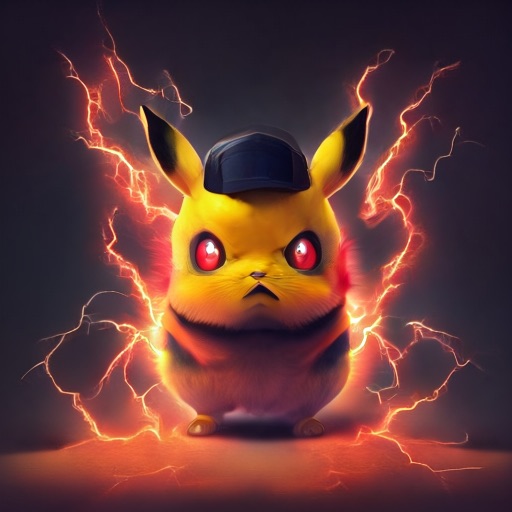} \\  
        \multicolumn{5}{c}{\textit{A Pikachu with an angry expression and red eyes, with lightning around it, hyper realistic style.}} \\

        \includegraphics[width=0.16\textwidth]{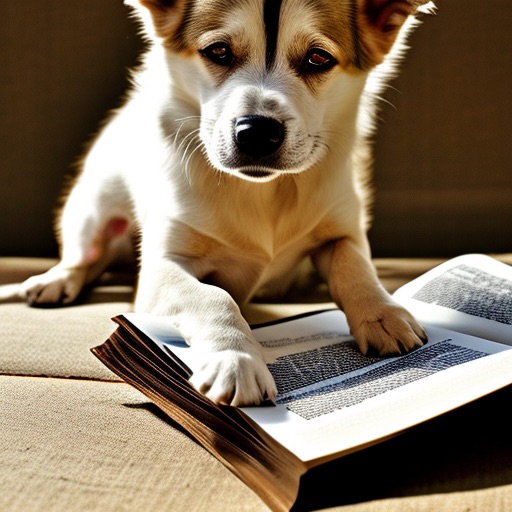} &  
        \includegraphics[width=0.16\textwidth]{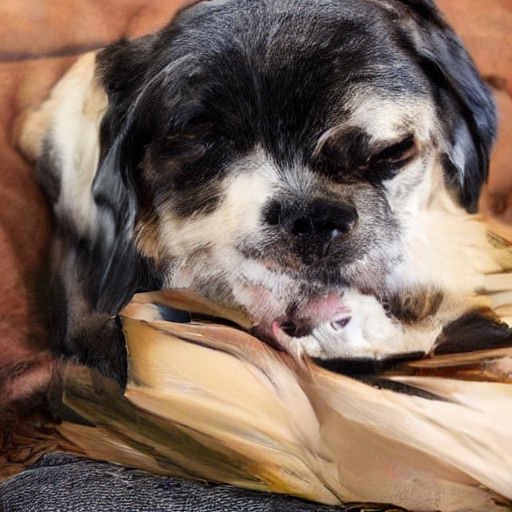} &
        \includegraphics[width=0.16\textwidth]{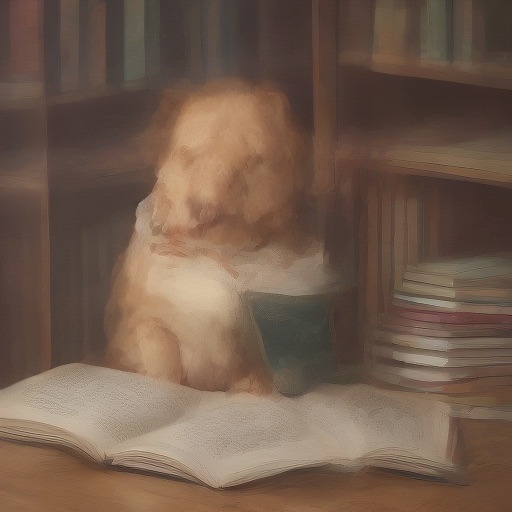} &
        \includegraphics[width=0.16\textwidth]{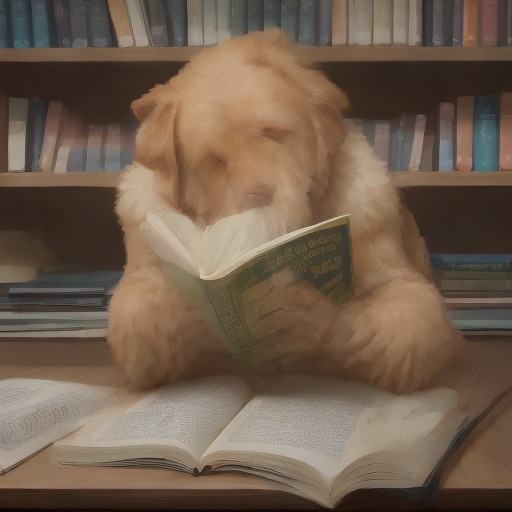} &  
        \includegraphics[width=0.16\textwidth]{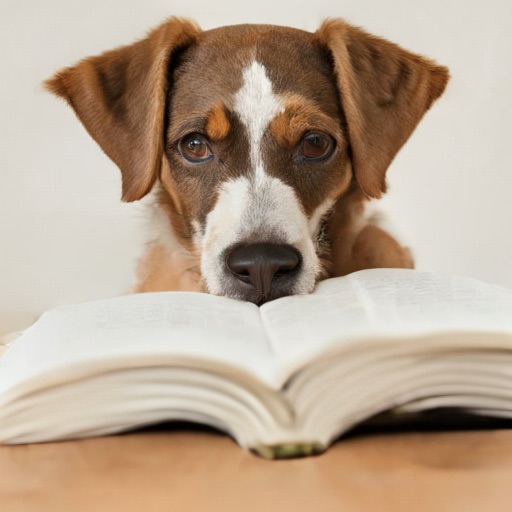} \\  
        \multicolumn{5}{c}{\textit{A dog is reading a thick book.}} \\

        \includegraphics[width=0.16\textwidth]{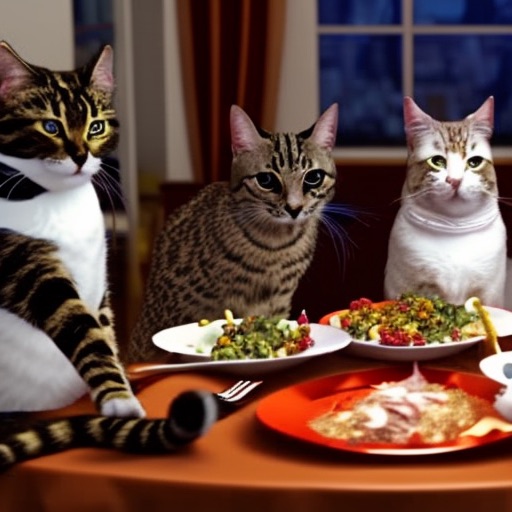} &  
        \includegraphics[width=0.16\textwidth]{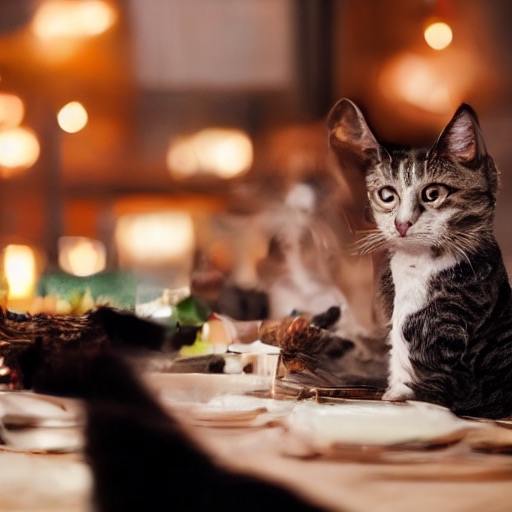} &
        \includegraphics[width=0.16\textwidth]{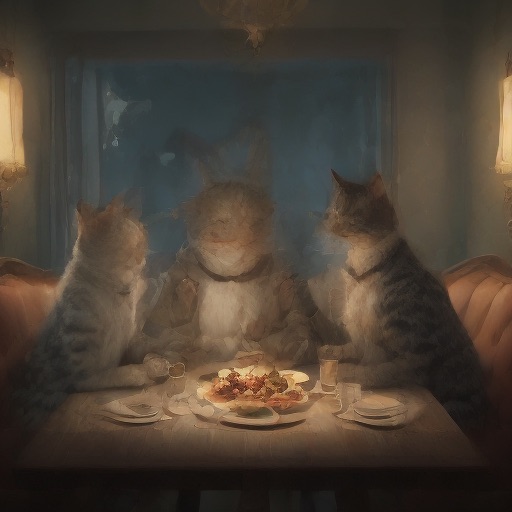} &
        \includegraphics[width=0.16\textwidth]{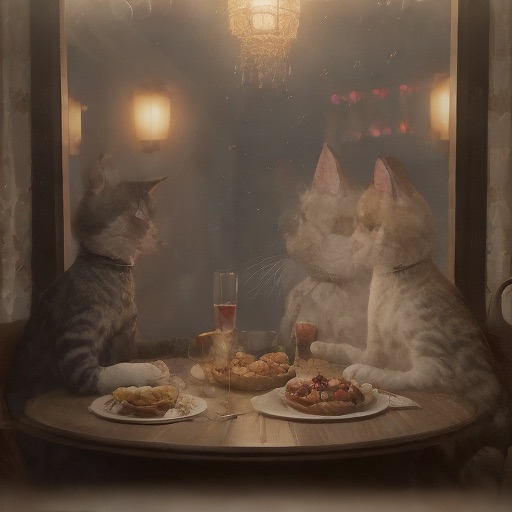} &  
        \includegraphics[width=0.16\textwidth]{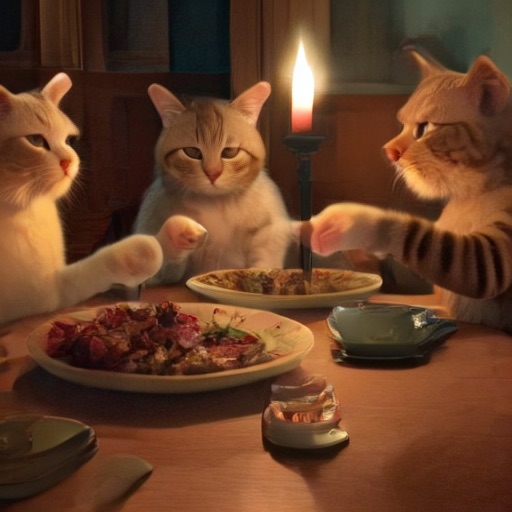} \\  
        \multicolumn{5}{c}{\textit{Three cats having dinner at a table at new years eve, cinematic shot, 8k.}} \\

        \includegraphics[width=0.16\textwidth]{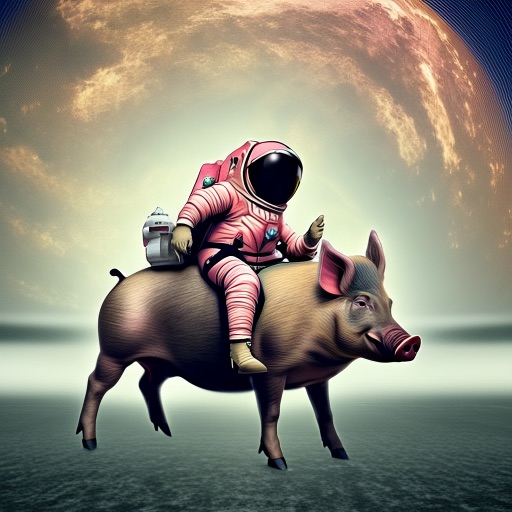} &  
        \includegraphics[width=0.16\textwidth]{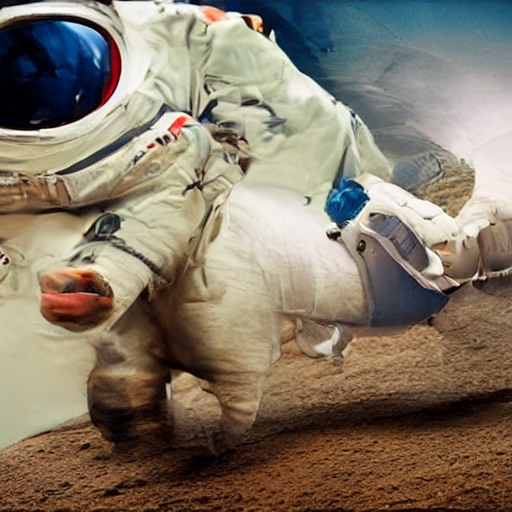} &
        \includegraphics[width=0.16\textwidth]{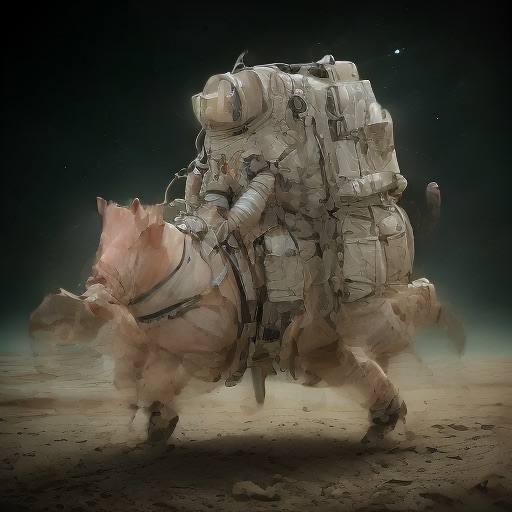} &
        \includegraphics[width=0.16\textwidth]{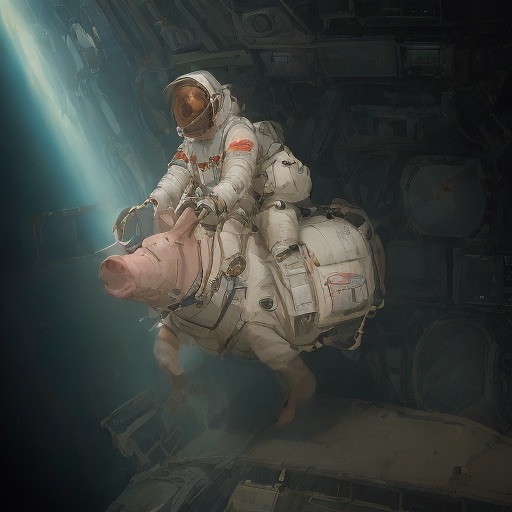} &  
        \includegraphics[width=0.16\textwidth]{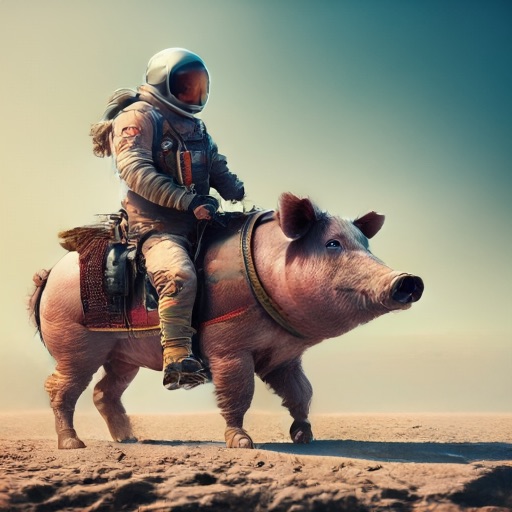} \\  
        \multicolumn{5}{c}{\textit{An astronaut riding a pig, highly realistic dslr photo, cinematic shot.}} \\
    \end{tabular}
    }
    \caption{Qualitative comparisons of UFOGen against competing methods and SD baseline. Zoom-in for better viewing.}
    \label{tab:main_comparison}
\end{table*}

%% file: sec/figure_tables/main_i2i.tex
\setlength{\tabcolsep}{2pt} 
\renewcommand{\arraystretch}{1.0} 
\begin{table}[!t]
    \centering
    \scriptsize
    \scalebox{1.0}{
\begin{tabular}{llp{5cm}}
        \includegraphics[width=0.15\textwidth]{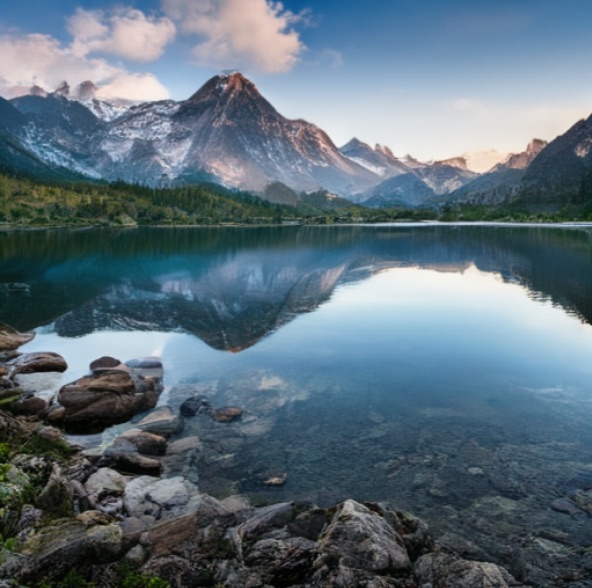} &  
        \includegraphics[width=0.15\textwidth]{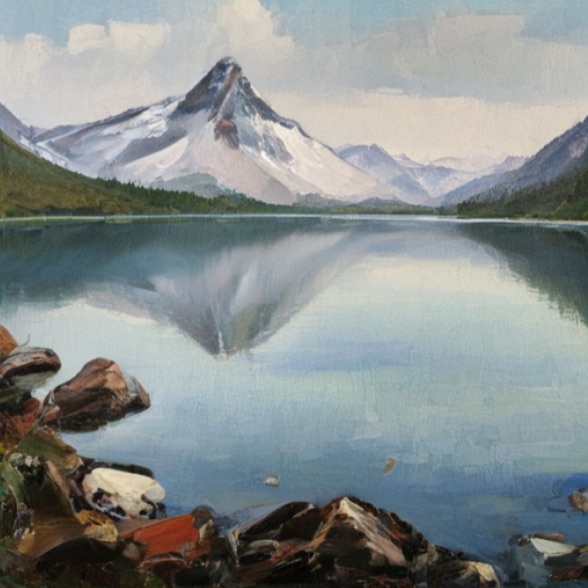} &
        \includegraphics[width=0.15\textwidth]{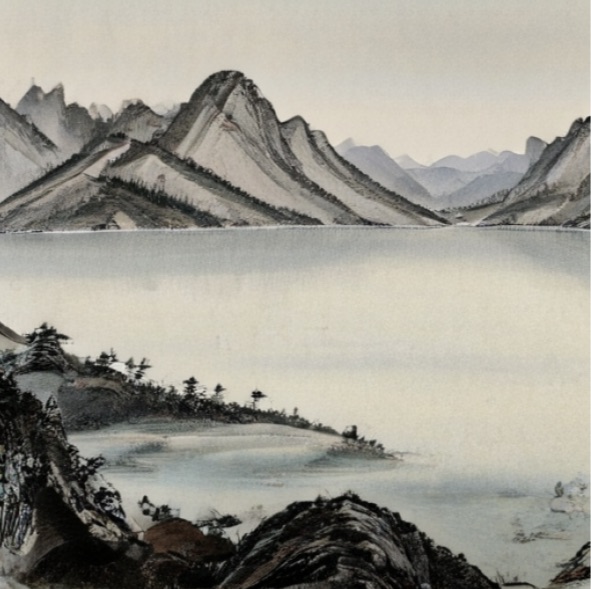} \\  
        \multicolumn{1}{c}{Input} & \parbox{2.5cm}{\textit{Oil painting of mountain and lake.}} & \parbox{2.5cm}{\textit{Chinese landscape painting.}}\\
        \includegraphics[width=0.15\textwidth]{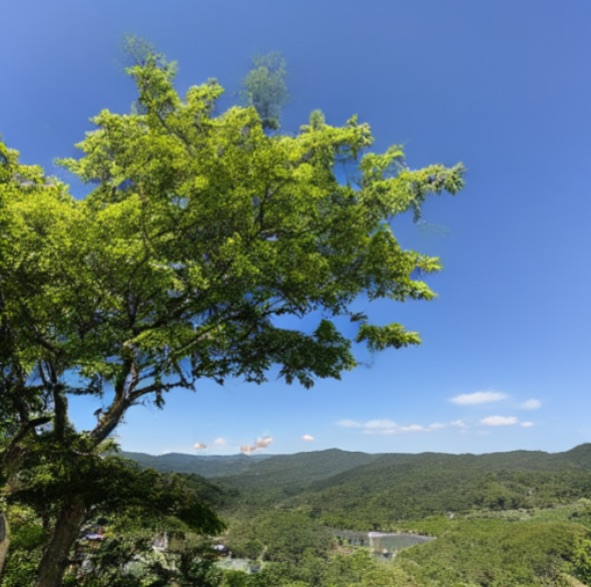} &  
        \includegraphics[width=0.15\textwidth]{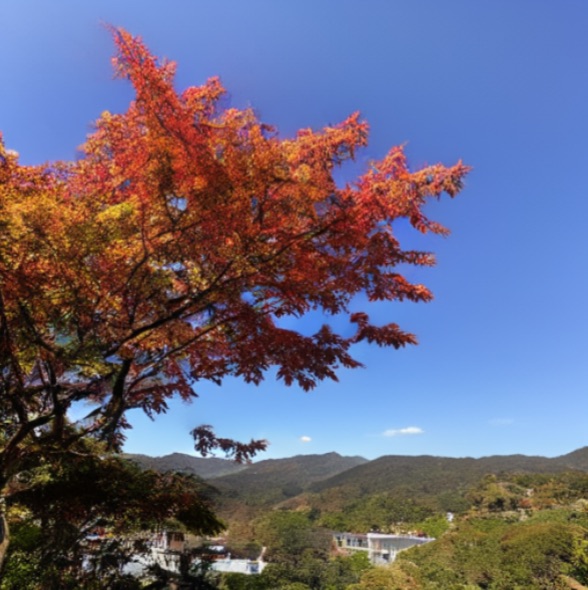} &
        \includegraphics[width=0.15\textwidth]{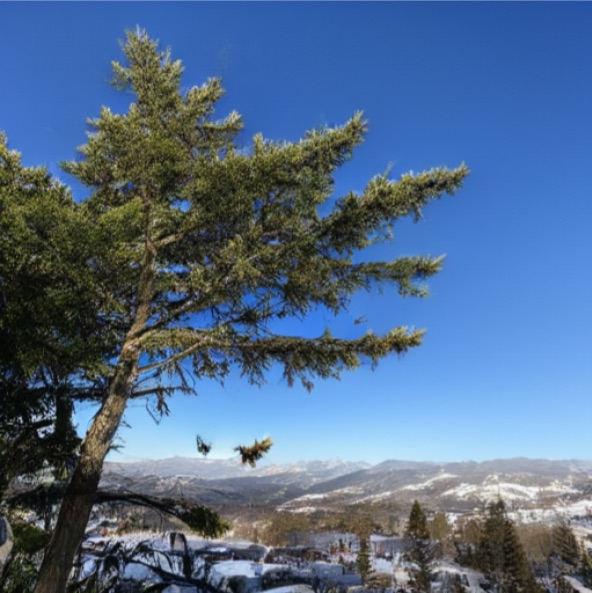} \\  
        \multicolumn{1}{c}{Input} & \parbox{2.5cm}{\textit{Tree with autumn leaves.}}&\textit{A winter scene.} \\

    \end{tabular}
    }
    \caption{Results of single-step image-to-image generation by UFOGen. Zoom in to view the details.}\label{tab:main_i2i}
\end{table}

\begin{table}[!t]
    \centering
    \footnotesize
    \scalebox{1.0}{
\begin{tabular}{llp{5cm}}
        \includegraphics[width=0.15\textwidth]{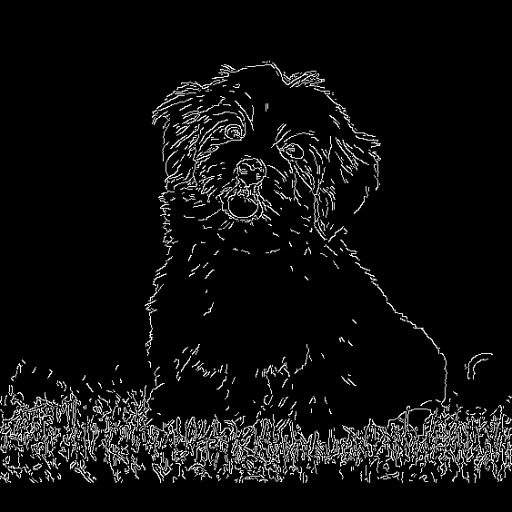} &  
        \includegraphics[width=0.15\textwidth]{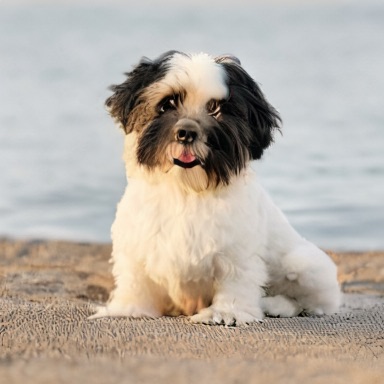} &
        \includegraphics[width=0.15\textwidth]{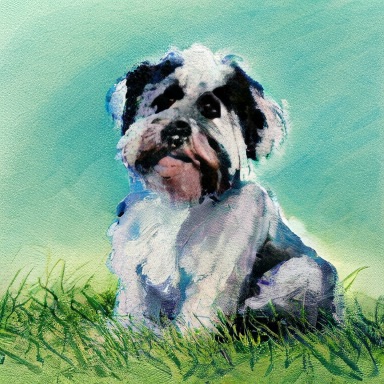} \\  
        \multicolumn{1}{c}{\scriptsize{Canny edge}} & \parbox{2.5cm}{\textit{\scriptsize{A cute black and white dog, sitting on the beach.}}} & \parbox{2.5cm}{\textit{\scriptsize{A cute dog, sitting on the grass, watercolor painting.}}}\\
        \includegraphics[width=0.15\textwidth]{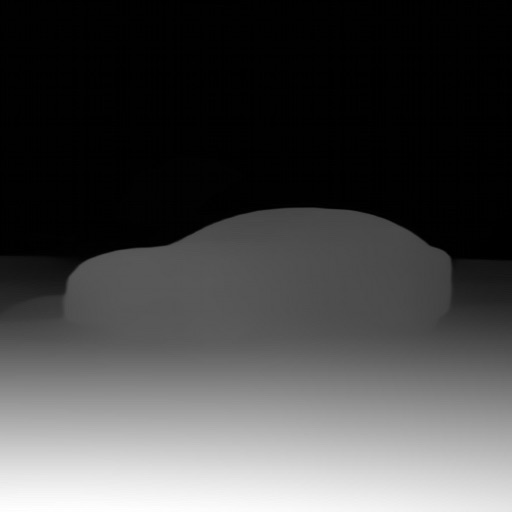} &  
        \includegraphics[width=0.15\textwidth]{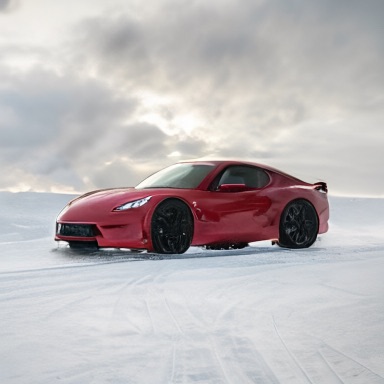} &
        \includegraphics[width=0.15\textwidth]{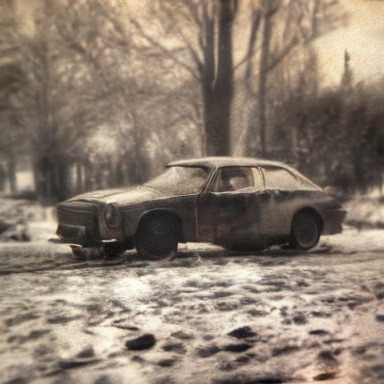} \\  
        \multicolumn{1}{c}{\scriptsize{Depth map}} & \parbox{2.5cm}{\textit{\scriptsize{a red sport car on snowfield.}}}&\textit{\scriptsize{Vintage photo of a rusty car.}} \\

    \end{tabular}
    }
    \caption{Results of controllable generation by UFOGen.}\label{tab:main_control}
\end{table}

%% file: sec/conclusion.tex
\section{Conclusions}
In this paper, we present UFOGen, a groundbreaking advancement in text-to-image synthesis that effectively addresses the enduring challenge of inference efficiency. Our innovative hybrid approach, combining diffusion models with a GAN objective, propels UFOGen to achieve ultra-fast, one-step generation of high-quality images conditioned on textual descriptions.
The comprehensive evaluations consistently affirm UFOGen's superiority over existing accelerated diffusion-based methods. Its distinct capability for one-step text-to-image synthesis and proficiency in downstream tasks underscore its versatility and mark it as a standout in the field. As a pioneer in enabling ultra-fast text-to-image synthesis, UFOGen paves the way for a transformative shift in the generative models landscape. The potential impact of UFOGen extends beyond academic discourse, promising to revolutionize the practical landscape of rapid and high-quality image generation.

%% file: sec/appendix.tex
\section{Appendices} \label{sec appendix}
\subsection{Deriving the KL objective in Equation \ref{eq:siddms_match}}\label{app:kl derivation}
In this section, we provide a derivation of obtaining a reconstruction objective from the KL term in Equation \ref{eq:siddms_match}:
\begin{align} \label{eq: kl term}
    \text{KL}(p_\theta(x_t|x_{t-1}')||q(x_t|x_{t-1})). 
\end{align}
Note that $q(x_t|x_{t-1}) = \mathcal{N}(\sqrt{1-\beta_t}x_{t-1}, \beta_t\mathbf{I})$ is a Gaussian distribution defined by the forward diffusion. For $p_\theta(x_t|x_{t-1}')$, although the distribution on $p_{\theta}(x_{t-1}')$ is quite complicated (because this depends on the generator model), given a specific $x_{t-1}'$, it follows the same distribution of forward diffusion: $p_\theta(x_t|x_{t-1}') = \mathcal{N}(\sqrt{1-\beta_t}x_{t-1}', \beta_t\mathbf{I})$. Therefore, Equation \ref{eq: kl term} is the KL divergence between two Gaussian distributions, which we can computed in closed form. For two multivariate Gaussian distributions with means $\mu_1, \mu_2$ and covariance $\Sigma_1, \Sigma_2$, the KL divergence can be expressed as 
\begin{align}
    \frac{1}{2}\left[\log \frac{\left|\Sigma_2\right|}{\left|\Sigma_1\right|}-d+\operatorname{tr}\left\{\Sigma_2^{-1} \Sigma_1\right\}+\left(\mu_2-\mu_1\right)^T \Sigma_2^{-1}\left(\mu_2-\mu_1\right)\right]. \nonumber
\end{align}
We can easily plug-in the means and variances for $q(x_t|x_{t-1})$ and $p_\theta(x_t'|x_{t-1}')$ into the expression. Note that $\Sigma_1 = \Sigma_2 = \beta_t\mathbf{I}$, so the expression can be simplified to
\begin{align}\label{eq: recon obj}
    \frac{(1-\beta_t)||x_{t-1}' - x_{t-1}||^2}{2\beta_t} + C,
\end{align}
where $C$ is a constant. Therefore, with the outer expectation over $q(x_t)$ in Equation \ref{eq:siddms_match}, minimizing the KL objective is equivalent to minimizing a weighted reconstruction loss between $x_{t-1}'$ and $x_{t-1}$, where $x_{t-1}$ is obtained by sampling $x_0 \sim q(x_0)$ and $x_{t-1} \sim q(x_{t-1}|x_0)$; $x_{t-1}'$ is obtained from generating an $x_0'$ from the generator followed by sampling $x_{t-1}' \sim q(x_{t-1}'|x_0')$.

Note that in ~\cite{ssidms}, the authors did not leverage the Gaussian distribution's KL-divergence property. Instead, they decomposed the KL-divergence into an entropy component and a cross-entropy component, subsequently simplifying each aspect by empirically estimating the expectation. This simplification effectively converges to the same objective as expressed in Equation \ref{eq: recon obj}, albeit with an appended term associated with entropy. The authors of ~\cite{ssidms} introduced an auxiliary parametric distribution for entropy estimation, which led to an adversarial training objective. Nevertheless, our analysis suggests that this additional term is dispensable, and we have not encountered any practical challenges when omitting it.

\subsection{Analysis of the distribution matching at $x_0$}\label{app_param_compare}
In this section, we offer a detailed explanation of why training the model with the objective presented in Equation \ref{eq:siddms_match} effectively results in matching $x_0$ and $x_0'$. The rationale is intuitive: $x_{t-1}$ and $x_{t-1}'$ are both derived from their respective base images, $x_0$ and $x_0'$, through independent Gaussian noise corruptions. As a result, when we enforce the alignment of distributions between $x_{t-1}$ and $x_{t-1}'$, this implicitly encourages a matching of the distributions between $x_0$ and $x_0'$ as well. To provide a more rigorous and formal analysis, we proceed as follows.

\subsubsection{Adversarial term} \label{app adv equivalent}
We provide an explanation of why the adversarial objective in Equation \ref{eq:siddms_match} corresponds to matching the distributions $q(x_0)$ and $p_{\theta}(x_0')$. Firstly, note that since $q(x_{t-1})=\mathbb{E}_{q(x_0)}\left[q(x_{t-1}|x_0)\right]$, where $q(x_{t-1}|x_0)$ is the Gaussian distribution defined by the forward diffusion process. Therefore, $q(x_{t-1})$ can be expressed as a convolution between $q(x_0)$ and a Gaussian kernel: 
\begin{align}
    q(x_{t-1}) = q(x_0)*k(x), \quad k(x) = \mathcal{N}(0, \left(1-\bar{\alpha}_{t-1}\right)\mathbf{I}).
\end{align}
Similarly, $p_(\theta)(x_{t-1}) =  p_{\theta}(x_0)*k(x)$, where $p_{\theta}(x_0)$ is the implicit distribution defined by the generator $G_{\theta}$.

In the following lemma, we show that for a probability divergence $D$, if $p(x)$ and $q(x)$ are convoluted with the same kernel $k(x)$, then minimizing $D$ on the distributions after the convolution is equivalent to matching the original distributions $p(x)$ and $q(x)$.

\begin{lemma}\label{theorem: onestep}
    Let $Y=X+K$, if $K$ is absolutely continuous with density $k(x)>0, x\in \mathbb{R}$. And a divergence $\textbf{D}(\mathbb{Q}||\mathbb{P})$ is a measure of the difference between distribution $\mathbb{Q}$ and $\mathbb{P}$, where $\textbf{D}(\mathbb{Q}||\mathbb{P})\geq 0$ and $\textbf{D}(\mathbb{Q}||\mathbb{P})=0 \Longleftrightarrow \mathbb{Q}=\mathbb{P}$. Then $\textbf{D}(q(y)||p(y))=0 \Longleftrightarrow q(x)=p(x)$.

    Proof: The probability density of the summation between two variables is the convolution between their probability densities. Thus, we have:
    \begin{align}
        \textbf{D}&(q(y)||p(y)) = \textbf{D}(q(x)*k(x)||p(x)*k(x)), \nonumber\\
        \textbf{D}&(q(x)*k(x)||p(x)*k(x)) = 0 ~~ a.e., \nonumber\\
        &\Longleftrightarrow q(x)*k(x)= p(x)*k(x), \nonumber\\
        &\Longleftrightarrow \mathcal{F}(q(x)*k(x))= \mathcal{F}(p(x)*k(x)), \nonumber\\
        &\Longleftrightarrow \mathcal{F}(q(x))\mathcal{F}(k(x))= \mathcal{F}(p(x))\mathcal{F}(k(x)), \nonumber\\
        &\Longleftrightarrow q(x)=p(x) \quad a.e., \nonumber
    \end{align}
    where $\mathcal{F}$ denotes the Fourier Transform, and we utilize the invertibility of the Fourier Transform for the above derivation.
\end{lemma}

Thus, from \textbf{Lemma}~\ref{theorem: onestep}, we can get $q(x_0)=p_\theta(x_0)$ almost everywhere when $\text{JSD}(q(x_{t-1})||p_\theta(x_{t-1})) = 0$. Notably, while training with the adversarial objective on $x_{t-1}$ inherently aligns the distributions of  $q(x_0)$ and $p_{theta}(x_0')$,it is crucial to acknowledge that we cannot directly employ GAN training on $x_0$. This is because the additive Gaussian noise, which serves to smooth the distributions, rendering GAN training more stable. Indeed, training GANs on smooth distributions is one of the essential components of all diffusion-GAN hybrid models, as highlighted in ~\cite{ddgans}.

\subsubsection{KL term} \label{app recon equivalent}
Here we show that minimizing the reconstruction loss in Equation \ref{eq: recon obj} over the expectation of $q(x_t)$ as in Equation \ref{eq:siddms_match} is equivalent to minimizing the reconstruction loss between $x_0$ and $x_{0}'$. According to the sampling scheme of $x_{t-1}$ and $x_{t-1}'$, we have 
\begin{align}\label{eq: expectation recon}
    \mathbb{E}_{q(x_{t-1}), p_{\theta}(x_{t-1}')}\left[ \frac{(1-\beta_t)||x_{t-1}' - x_{t-1}||^2}{2\beta_t}\right] & = \mathbb{E}_{q(x_0)q(x_{t-1}|x_0),  p_{\theta}(x_0')p_{\theta}(x_{t-1}'|x_0')}\left[ \frac{(1-\beta_t)||x_{t-1}' - x_{t-1}||^2}{2\beta_t}\right].
\end{align}
Since the forward diffusion $q(x_{t-1}|x_0)$ has the Gaussian form ~\cite{ho2020denoising}
\begin{align}
    q(x_{t-1}|x_0) = \mathcal{N}\left(\sqrt{\bar{\alpha}_{t-1}} \mathbf{x}_0,\left(1-\bar{\alpha}_{t-1}\right) \mathbf{I}\right)
\end{align}
and similar form holds for $p_{\theta}(x_{t-1}'|x_0')$, we can rewrite the expectation in Equation \ref{eq: expectation recon} over the distribution of simple Gaussian distribution $p(\epsilon)=\mathcal{N}\left(\epsilon; 0, \mathbf{I}\right)$: 
\begin{align}\label{eq: expectation recon 2}
\mathbb{E}_{q(x_0)q(x_{t-1}|x_0),  p_{\theta}(x_0')p_{\theta}(x_{t-1}'|x_0')}\left[ \frac{(1-\beta_t)||x_{t-1}' - x_{t-1}||^2}{2\beta_t}\right] = \mathbb{E}_{q(x_0), p_{\theta}(x_0'), p(\epsilon)}\left[ \frac{(1-\beta_t)||x_{t-1}' - x_{t-1}||^2}{2\beta_t}\right],
\end{align}
where $x_{t-1}' = \sqrt{\bar{\alpha}_{t-1}} \mathbf{x}_0' + \left(1-\bar{\alpha}_{t-1}\right)\epsilon'$ and $x_{t-1} = \sqrt{\bar{\alpha}_{t-1}} \mathbf{x}_0 + \left(1-\bar{\alpha}_{t-1}\right)\epsilon$ are obtained by i.i.d. samples $\epsilon', \epsilon$ from $p(\epsilon)$. Plug in the expressions to Equation \ref{eq: expectation recon 2}, we obtain 
\begin{align}
    &\mathbb{E}_{q(x_0), p_{\theta}(x_0'), p(\epsilon)}\left[ \frac{(1-\beta_t)||x_{t-1}' - x_{t-1}||^2}{2\beta_t}\right] \nonumber \\
    &=\mathbb{E}_{q(x_0), p_{\theta}(x_0'), p(\epsilon)}\left[ \frac{(1-\beta_t)||\sqrt{\bar{\alpha}_{t-1}}(x_0'-x_0) + \left(1-\bar{\alpha}_{t-1}\right)(\epsilon'-\epsilon)||^2}{2\beta_t}\right] \nonumber\\
    &=\mathbb{E}_{q(x_0), p_{\theta}(x_0')}\left[ \frac{(1-\beta_t)\bar{\alpha}_{t-1}||x_0'-x_0||^2}{2\beta_t}\right] + C, \nonumber
\end{align}
where $C$ is a constant independent of the model. Therefore, we claim the equivalence of the reconstruction objective and the matching between $x_0$ and $x_0'$.

However, it's essential to emphasize that the matching between $x_0$ and $x_0'$ is performed with an expectation over Gaussian noises. In practical terms, this approach can introduce significant variance during the sampling of $x_{t-1}$ and $x_{t-1}'$. This variance, in turn, may result in a less robust learning signal when it comes to aligning the distributions at clean data.As detailed in Section \ref{sec improved rec}, we propose a refinement to address this issue. Specifically, we advocate for the direct enforcement of reconstruction between $x_0$ and $x_0'$. This modification introduces explicit distribution matching at the level of clean data, enhancing the model's robustness and effectiveness.

\subsection{Experimental Details}\label{app exp details}
For all the experiments, we initialize the parameters of both the generator and discriminator with the pre-trained Stable Diffusion 1.5 checkpoint. In consequence, we follow SD 1.5 to use the same VAE for image encoding/decoding and the frozen text encoder of CLIP ViT-L/14 for text conditioning. Note that both the generator and discriminator operates on latent space. In other words, the generator generates the latent variables and the discriminator distinguishes the fake and true (noisy) latent variables. 

\paragraph{Important Hyper-parameters} One important hyper-parameter is the \textit{denoising step size} during training, which is the gap between $t-1$ and $t$. Note that in Section \ref{sec_method one step}, we mentioned that the model is trained with multiple denoising steps, while it enables one-step inference. Throughout the experiments, we train the models using denoising step size 250, given the $1000$-step discrete time scheduler of SD. Specifically, during training, we sample $t$ randomly from $1$ to $1000$, and the time step for $t-1$ is $max(0, t-250)$. We conduct ablation studies on this hyper-parameter in Section \ref{app extra ablation}.

Another important hyper-parameter is $\lambda_{KL}$, the weighting coefficient for reconstruction term in the objective in Equation \ref{eq:ufo obj}. We set $\lambda_{KL} = 1.0$ throughout the experiments. We found the results insensitive to slight variations of this coefficient.
\paragraph{Common Hyper-parameters}
We train our models on the LAION Aesthetic 6+ dataset. For the generator, we use AdamW optimizer \cite{loshchilov2018decoupled} with $\beta_1=0.9$ and $\beta_2=0.999$; for the discriminator, we use AdamW optimizer with $\beta_1=0.0$ and $\beta_2=0.999$. We adopt learning rate warm-up in the first 1000 steps, with peak learning rate $1e-4$ for both the discriminator and the generator. For training the generator, we apply gradient norm clipping with value $1.0$ for generator only. We use batch size 1024. For the generator, we apply EMA with coefficient 0.999. We observe quick convergence, typically in $<50$k steps. 

\subsection{Additional Results of Ablation Studies}\label{app extra ablation}
In  this section, we provide additional results for ablation studies, which are briefly covered in the main text due to the constraints of space. In Appendix \ref{app qualitative ablation}, we provide qualitative results corresponds to the ablation study conducted in Section \ref{sec exp ablation}. In Appendix \ref{app stepsize ablation}, we conduct an additional ablation experiment on the denoising step size during training. 

\subsubsection{Qualitative Results for Table \ref{tab:main_ablation}}\label{app qualitative ablation}
We provide qualitative examples to contrast between the single-step sample generated by SIDDM \cite{ssidms} and our proposed UFOGen. Results are shown in Table \ref{tab: app qualitative ablation 1} and \ref{tab: app qualitative ablation 2}. We observe that when sampling from SIDDM in only one-step, the samples are blurry and over-smoothed, while UFOGen can produce sharp samples in single step. The observation strongly supports the effectiveness of our introduced modifications to the training objective. 
\input{sec/figure_tables/app_ablation_1}

\subsubsection{Ablation on Denoising Step-size}\label{app stepsize ablation}
One important hyper-parameter of training UFOGen is the denoising step size, which is the gap between $t$ and $t-1$ during training. Note that although UFOGen can produce samples in one step, the training requires a meaningful denoising step size to compute the adversarial loss on noisy observations. Our model is based on Stable Diffusion, which adopts a discrete time scheduler with 1000 steps. Previous diffusion GAN hybrid models \cite{ssidms,ddgans} divides the denoising process into 2 to 4 steps. We explore denoising step size 125, 250, 500 and 1000, which corresponds to divide the denoising process to 8, 4, 2, and 1 steps. Note that during training, we sample $t$ uniformly in $[1,1000)$, and when the sampled $t$ is smaller than the denoising step size, we set $t-1$ to be $0$. In other words, a denoising step size $1000$ corresponds to always setting $t-1=0$ and hence the adversarial loss is computed on clean data $x_0$. 

Quantitative results of the ablation study is presented in Table \ref{tab:steps_ablation}. We observe that a denoising step size 1000 fails, suggesting that training with the adversarial loss on noisy data is critical for stabilizing the diffusion-GAN training. This observation was made on earlier work \cite{ssidms,ddgans} as well. We also observe that denoising step size 250 is the sweet spot, which is also aligned with the empirical observations of \cite{ddgans,ssidms}. We conjecture that the reason for the performance degrade when reducing the denoising step size is that the discriminator does not have enough capacity to discriminate on many distinct noise levels.

\begin{table*}[t]
    \centering
    \begin{tabular}{ccc}
    \toprule
    Denoising Step-size  & FID-5k & CLIP \\ \midrule
1000 & 32.92  & 0.288 \\
500 & 23.2  & 0.314 \\
250 & 22.5  & 0.311 \\
125 & 24.7  & 0.305 \\
    \bottomrule
    \end{tabular}
    \caption{Ablation study comparing the denoising step size during training. For all training denoising step sizes, we generate the samples in one step.
    }
    \label{tab:steps_ablation}
\end{table*}

\subsection{Additional Results for Qualitative Comparisons}
\subsubsection{Failure of Single-step LCM} \label{app failed lcm-1}
Consistency models try to learn the consistency mapping that maps every point on the PF-ODE trajectory to its boundary value, i.e., $x_0$ \cite{song2023consistency}, and therefore ideally consistency models should generate samples in one single step. However, in practice, due to the complexity of the ODE trajectory, one-step generation for consistency models is not feasible, and some iterative refinements are necessary. Notably, Latent consistency models (LCM) ~\cite{luo2023latent} distilled the Stable Diffusion model into a consistency model, and we observe that single-step sampling fail to generate reasonable textures. We demonstrate the single-step samples from LCM in figure \ref{fig:lcm1 fail}. Due to LCM's ineffectiveness of single-step sampling, we only qualitatively compare our model to 2-step and 4-step LCM.
\begin{figure*}[!htp]
  \centering
    \includegraphics[width=0.9\textwidth]{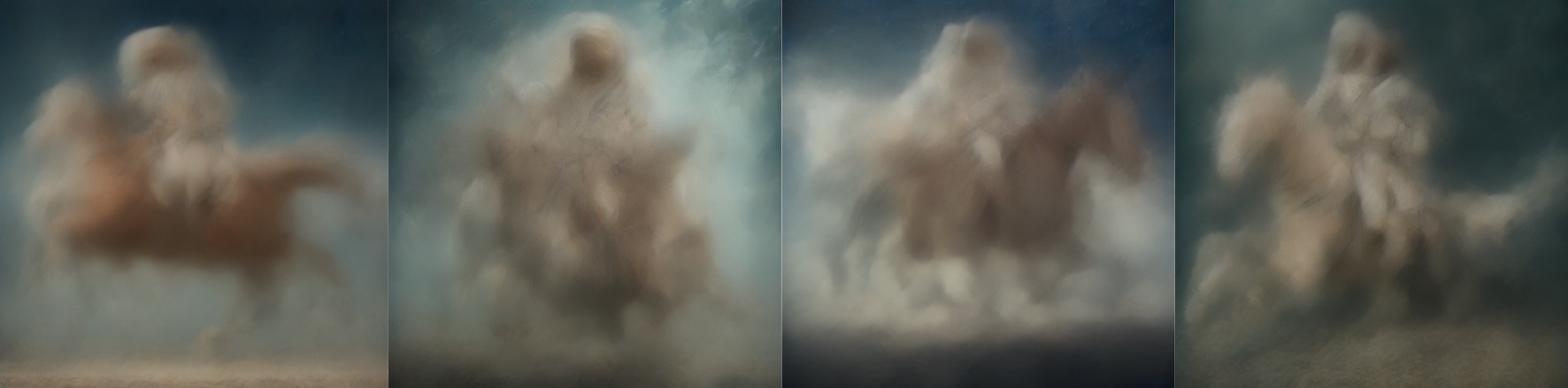}
  \caption{Single-step samples from LCM ~\cite{luo2023latent} with prompt ``Photo of an astronaut riding a horse".}
\end{figure*}\label{fig:lcm1 fail}

\subsubsection{Extended Results of Table \ref{tab:main_comparison}}\label{app extra comparison}
\input{sec/figure_tables/app_compare}
In consideration of space constraints in the main text, our initial qualitative comparison of UFOGen with competing methods for few-step generation in Table \ref{tab:main_comparison} employs a single image per prompt. It is essential to note that this approach introduces some variability due to the inherent randomness in image generation. To provide a more comprehensive and objective evaluation, we extend our comparison in this section by presenting four images generated by each method for every prompt. This expanded set of prompts includes those featured in Table \ref{tab:main_comparison}, along with additional prompts. The results of this in-depth comparison are illustrated across Table \ref{tab:app_comparison_1} to \ref{tab:app_comparison_8}, consistently highlighting UFOGen's advantageous performance in generating sharp and visually appealing images within an ultra-low number of steps when compared to competing methods.

Concurrent to our paper submission, the authors of LCM \cite{luo2023latent} released updated LCM models trained with more resources. The models are claimed to be stronger than the initially released LCM model, which is used in our qualitative evaluation. For fairness in the comparison, we obtain some qualitative samples of the updated LCM model that shares the SD 1.5 backbone with us\footnote{\url{https://huggingface.co/latent-consistency/lcm-lora-sdv1-5}}, and show them in Table \ref{tab: app new lcm 1} and \ref{tab: app new lcm 2}. We observe that while the new LCM model generates better samples than initial LCM model does, our single-step UFOGen is still highly competitive against 4-step LCM and significantly better than 2-step LCM. 
\input{sec/figure_tables/new_lcm_results}

\subsection{Additional Qualitative Samples from UFOGen}\label{app extra t2i results}
In this section, we present supplementary samples generated by UFOGen models, showcasing the diversity of results in Table \ref{tab:app_extra_1}, \ref{tab:app_extra_2} and \ref{tab:app_extra_3}. Through an examination of these additional samples, we deduce that UFOGen exhibits the ability to generate high-quality and diverse outputs that align coherently with prompts spanning various styles (such as painting, photo-realistic, anime) and contents (including objects, landscapes, animals, humans, etc.). Notably, our model demonstrates a promising capability to produce visually compelling images with remarkable quality within just a single sampling step.

In Table \ref{tab:app_fail}, we present some failure cases of UFOGen. We observe that UFOGen suffers from missing objects, attribute leakage and counting, which are common issues of SD based models, as discussed in \cite{chefer2023attend,feng2022training}.
\input{sec/figure_tables/app_extra_results}

\subsection{Additional Results of UFOGen's Applications}\label{app extra application}
In this section, we provide extended results of UFOGen's applications, including the image-to-image generation in Figure~\ref{fig:i2i_ext} and controllable generation in Figure~\ref{fig:plugin_ext}. 

%% file: sec/figure_tables/app_ablation_1.tex
\setlength{\tabcolsep}{2pt} 
\renewcommand{\arraystretch}{1.0} 
\begin{table*}[!t]
    \centering
    \scalebox{1.05}{
    \begin{tabular}{cc}
        SIDDM (1 step) & UFOen (1 step) \\
        \includegraphics[width=0.4\textwidth]{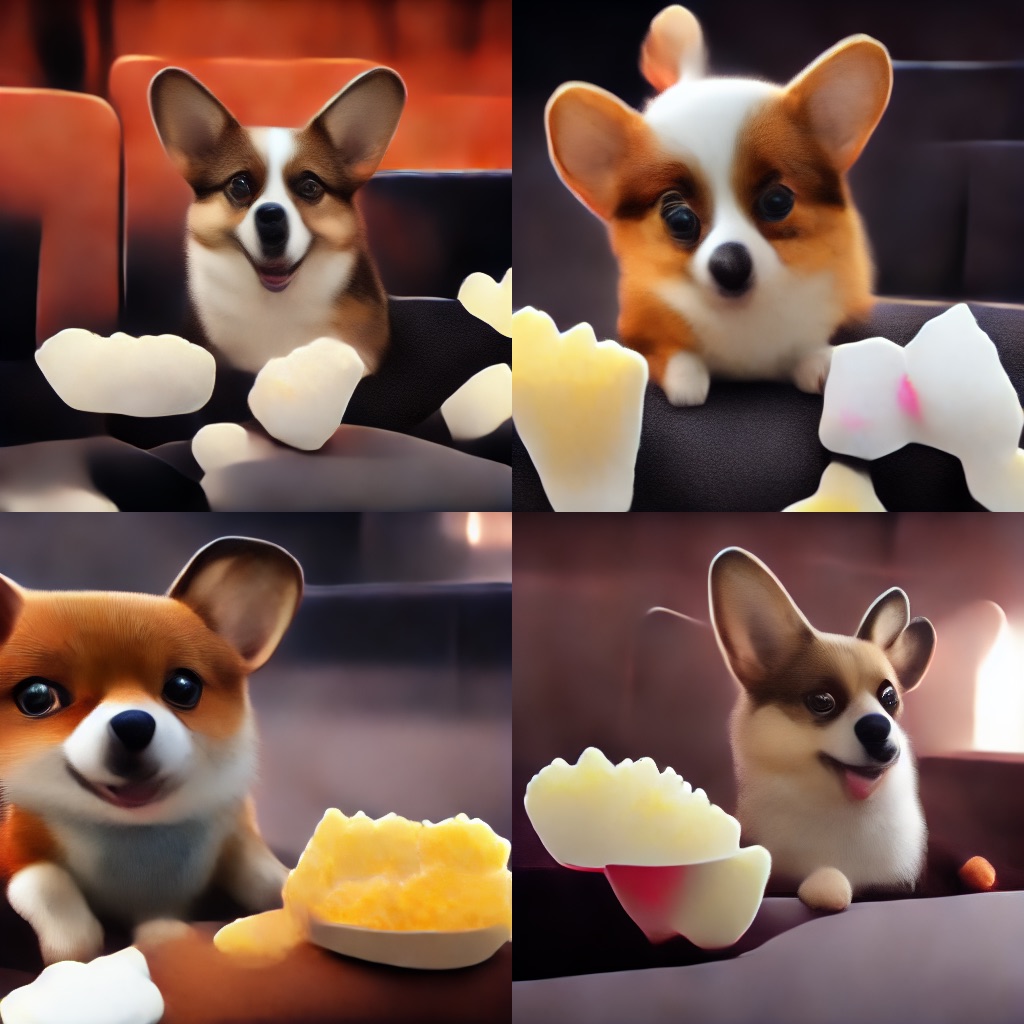} &
        \includegraphics[width=0.4\textwidth]{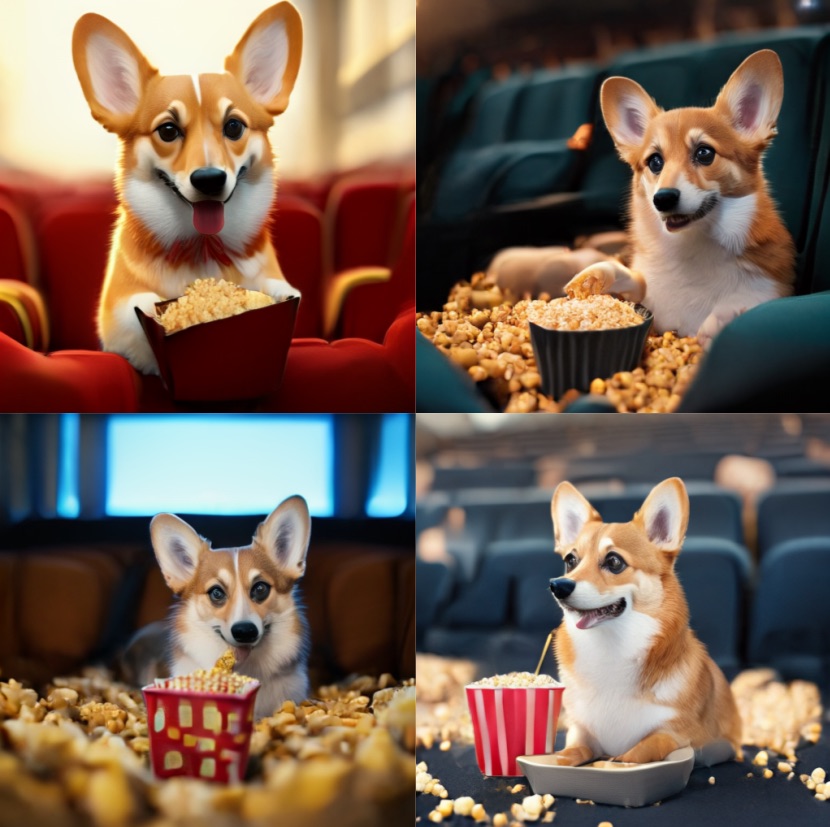} \\
        \multicolumn{2}{c}{\textit{Cute small corgi sitting in a movie theater eating popcorn, unreal engine.}} \\

        \includegraphics[width=0.4\textwidth]{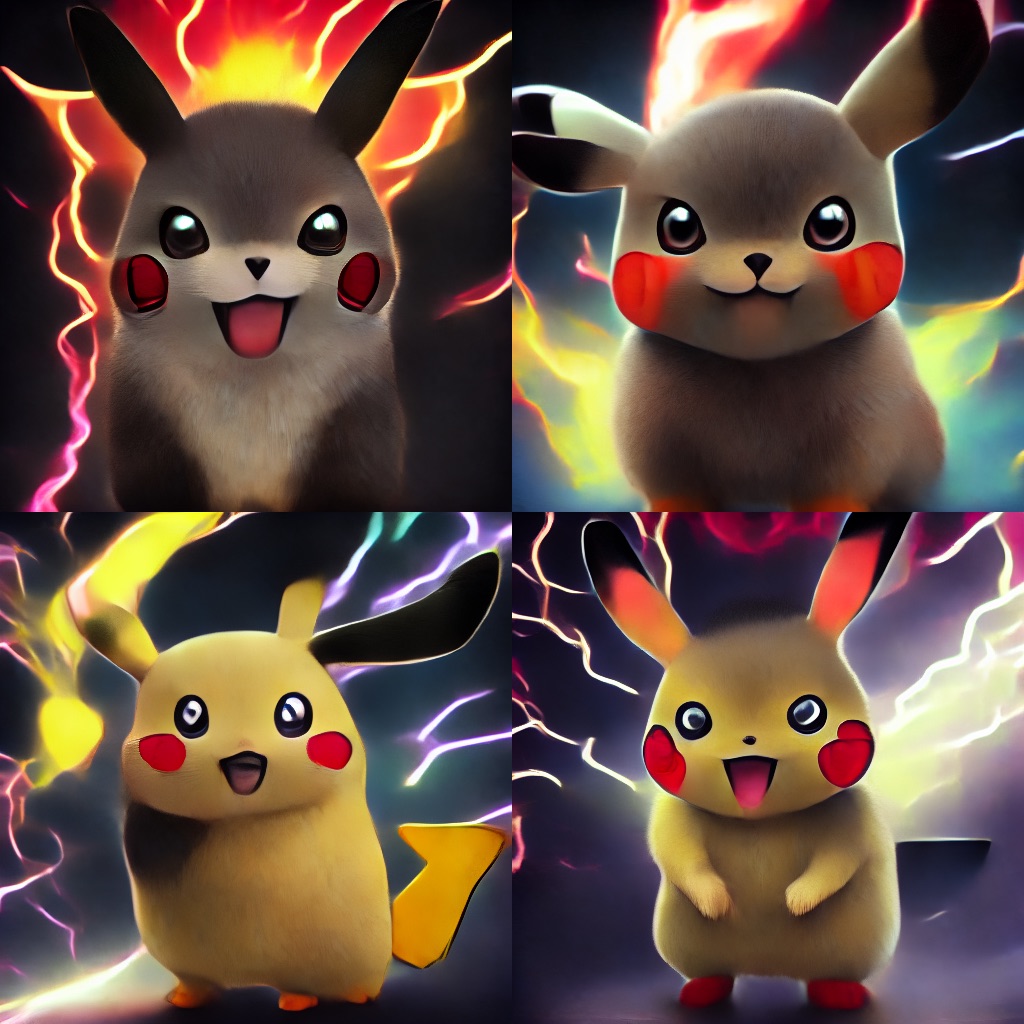} &
        \includegraphics[width=0.4\textwidth]{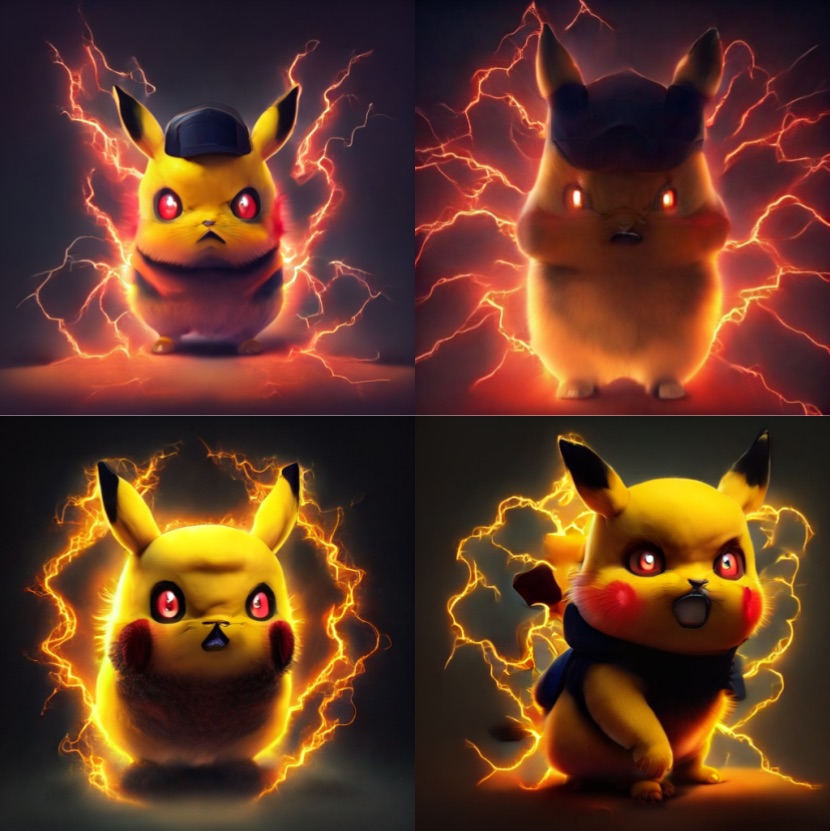} \\
        \multicolumn{2}{c}{\textit{A Pikachu with an angry expression and red eyes, with lightning around it, hyper realistic style.}} 
    \end{tabular}
    }
    \caption{Qualitative results for the ablation study that compares one-step samples from SIDDM and UFOGen.}
    \label{tab: app qualitative ablation 1}
\end{table*}

\begin{table*}[!t]
    \centering
    \scalebox{1.05}{
    \begin{tabular}{cc}
        SIDDM (1 step) & UFOen (1 step) \\
        \includegraphics[width=0.4\textwidth]{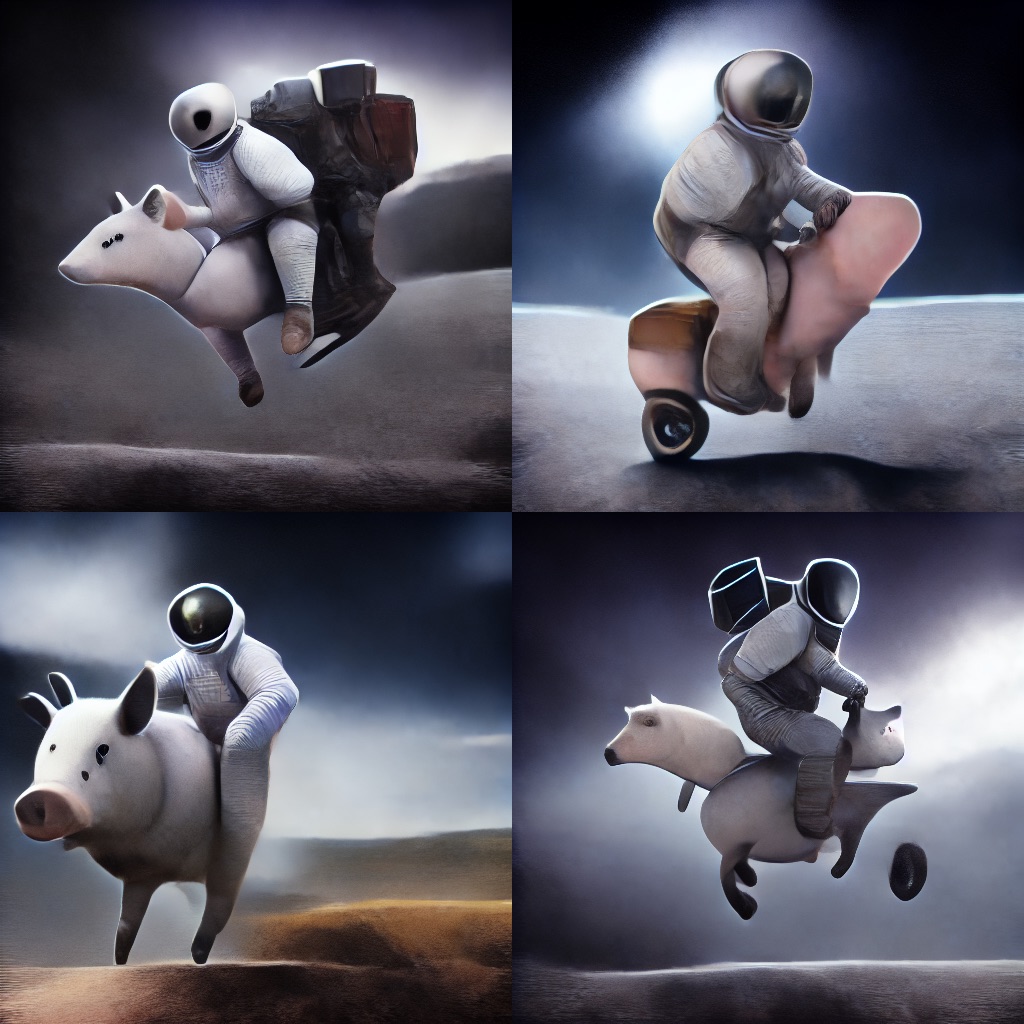} &
        \includegraphics[width=0.4\textwidth]{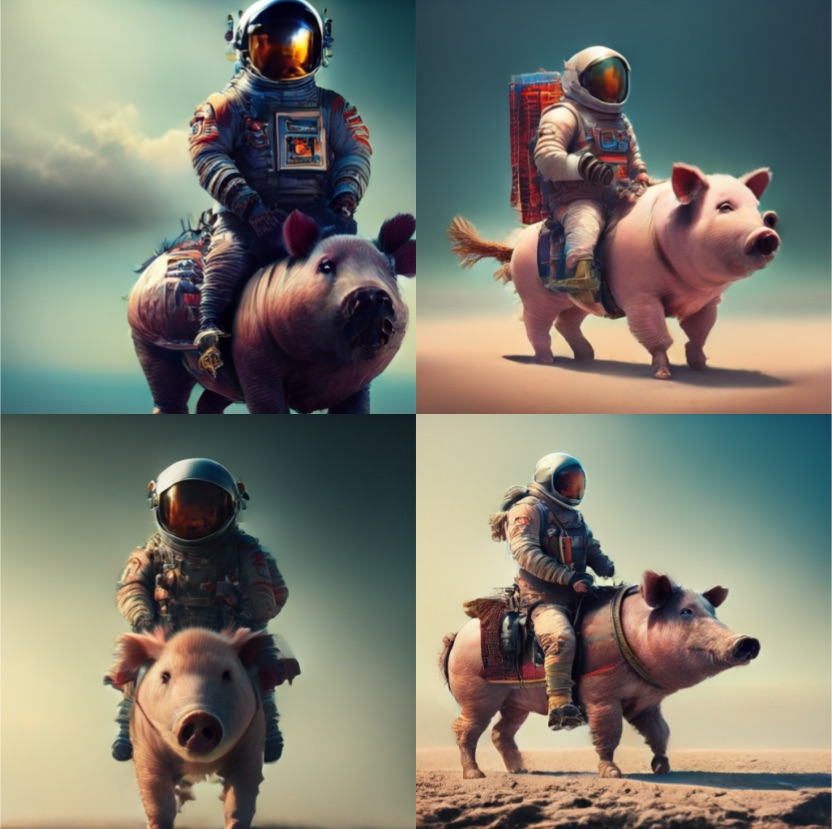} \\
        \multicolumn{2}{c}{\textit{An astronaut riding a pig, highly realistic dslr photo, cinematic shot.}} \\

        \includegraphics[width=0.4\textwidth]{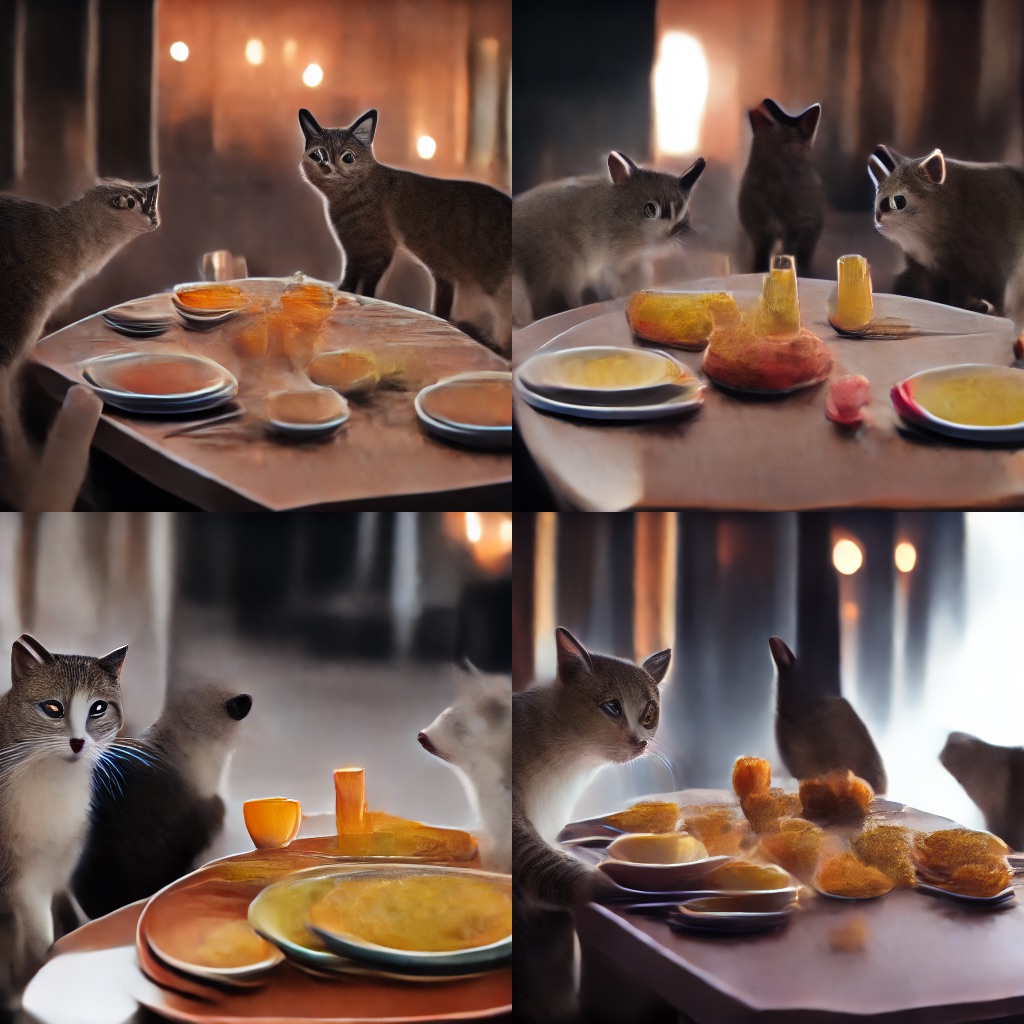} &
        \includegraphics[width=0.4\textwidth]{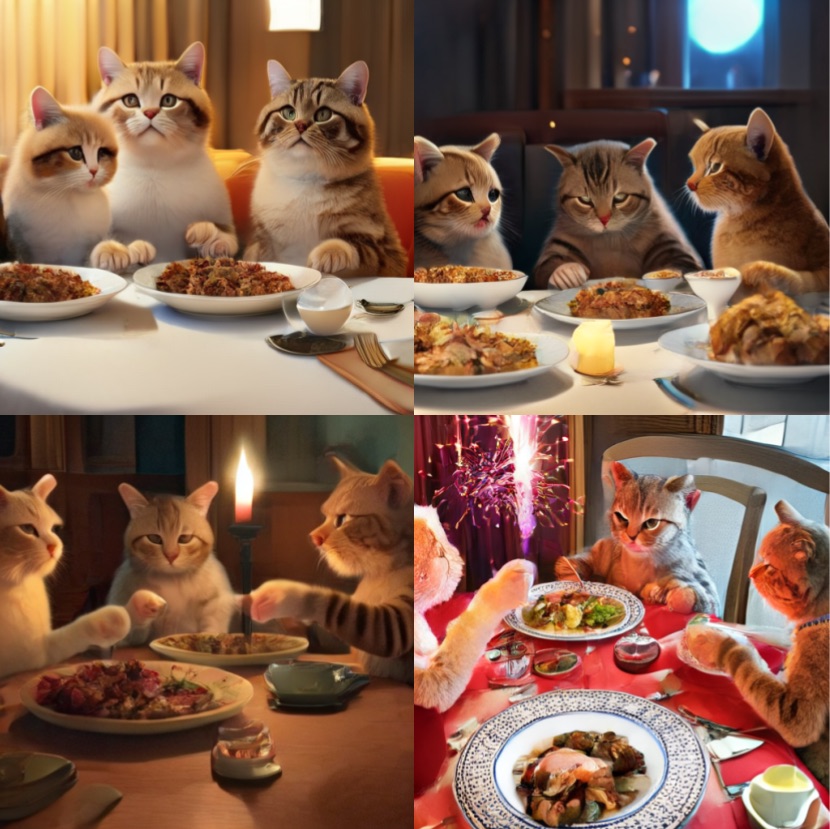} \\
        \multicolumn{2}{c}{\textit{Three cats having dinner at a table at new years eve, cinematic shot, 8k.}} 
    \end{tabular}
    }
    \caption{Qualitative results for the ablation study that compares one-step samples from SIDDM and UFOGen.}
    \label{tab: app qualitative ablation 2}
\end{table*}

%% file: sec/figure_tables/app_compare.tex
\setlength{\tabcolsep}{2pt} 
\renewcommand{\arraystretch}{1.0} 
\begin{table*}[!t]
    \centering
    \scalebox{1.05}{
    \begin{tabular}{cc}
        \includegraphics[width=0.4\textwidth]{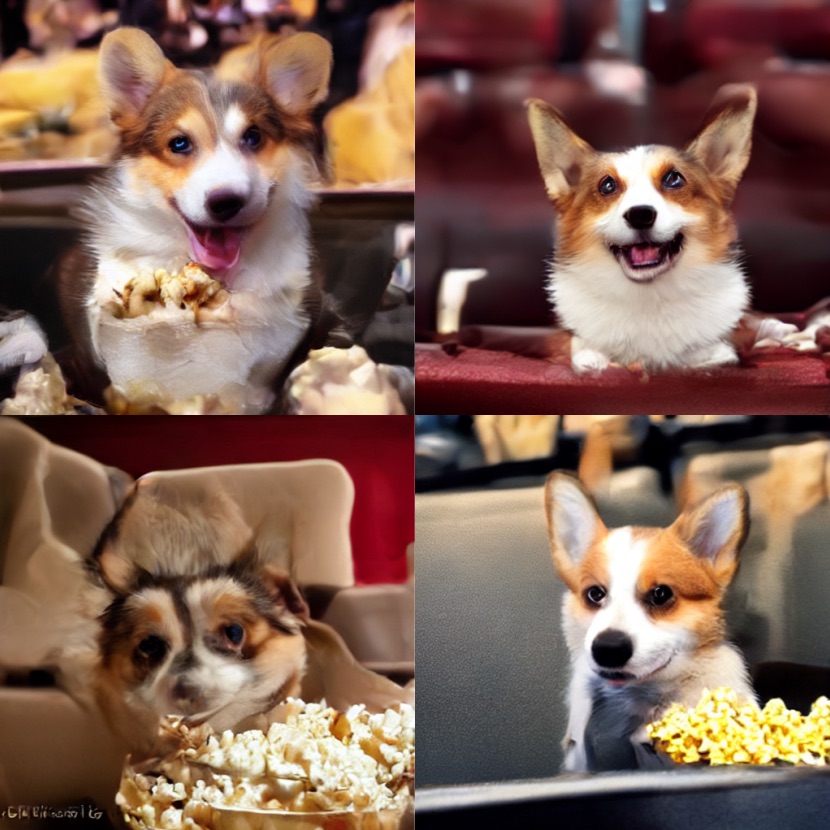} &
        \includegraphics[width=0.4\textwidth]{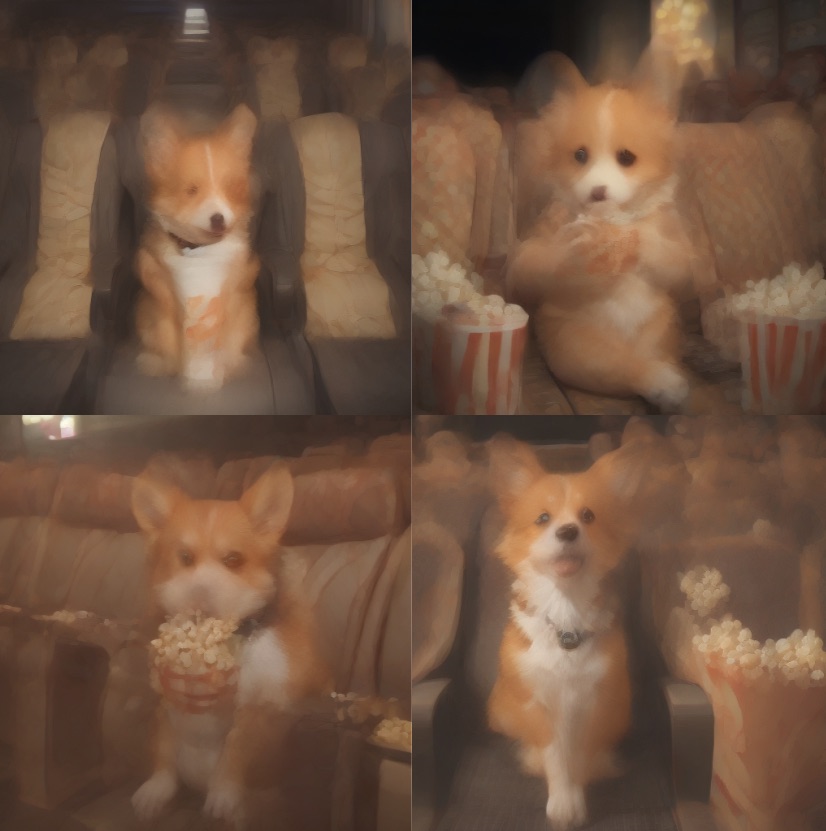} \\

        InstaFlow (1 step) & LCM (2 steps) \\
                &  \\
        \includegraphics[width=0.4\textwidth]{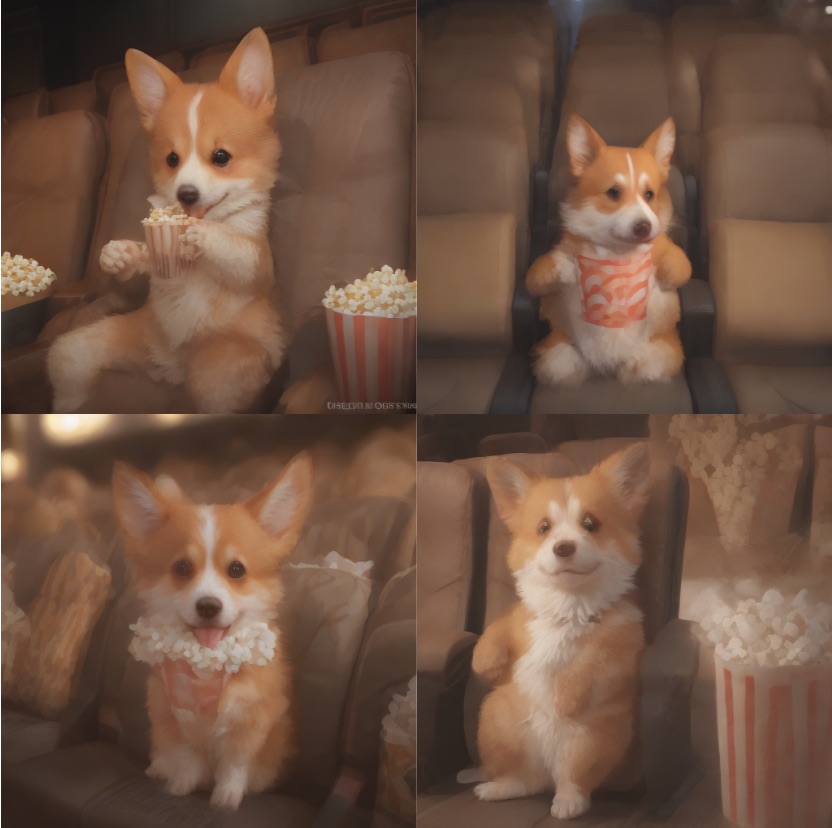} &
        \includegraphics[width=0.4\textwidth]{fig/app_compare/corgi/ufo.jpeg} \\
        LCM (4 steps) & UFOGen (1 step) \\
    \end{tabular}
    }
    \caption{Prompt: \textit{Cute small corgi sitting in a movie theater eating popcorn, unreal engine.}}
    \label{tab:app_comparison_1}
\end{table*}

\begin{table*}[!t]
    \centering
    \scalebox{1.05}{
    \begin{tabular}{cc}
        \includegraphics[width=0.4\textwidth]{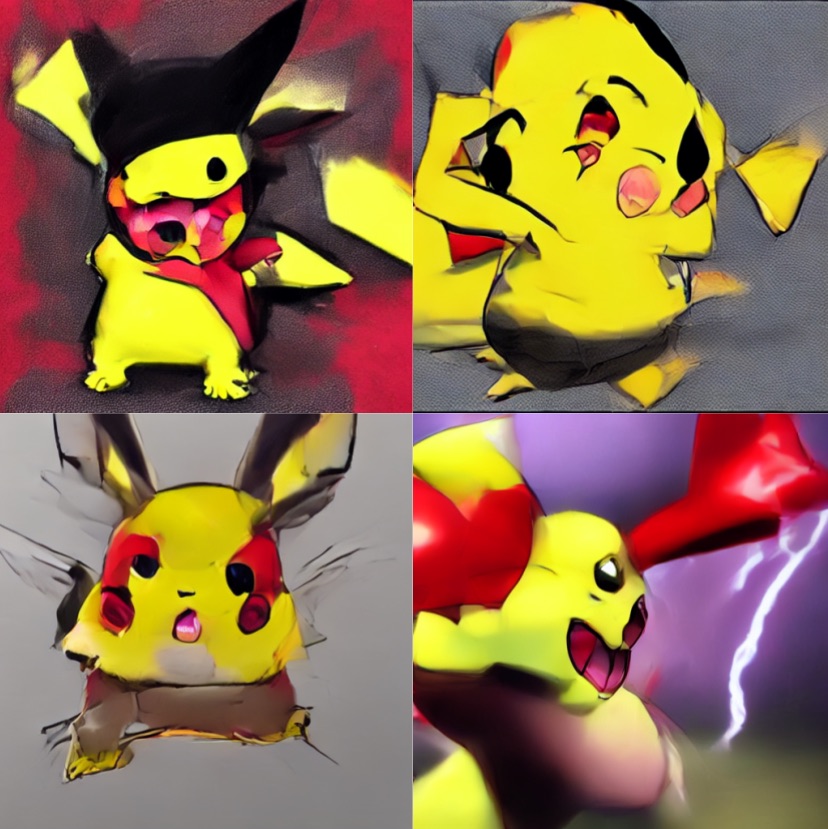} &
        \includegraphics[width=0.4\textwidth]{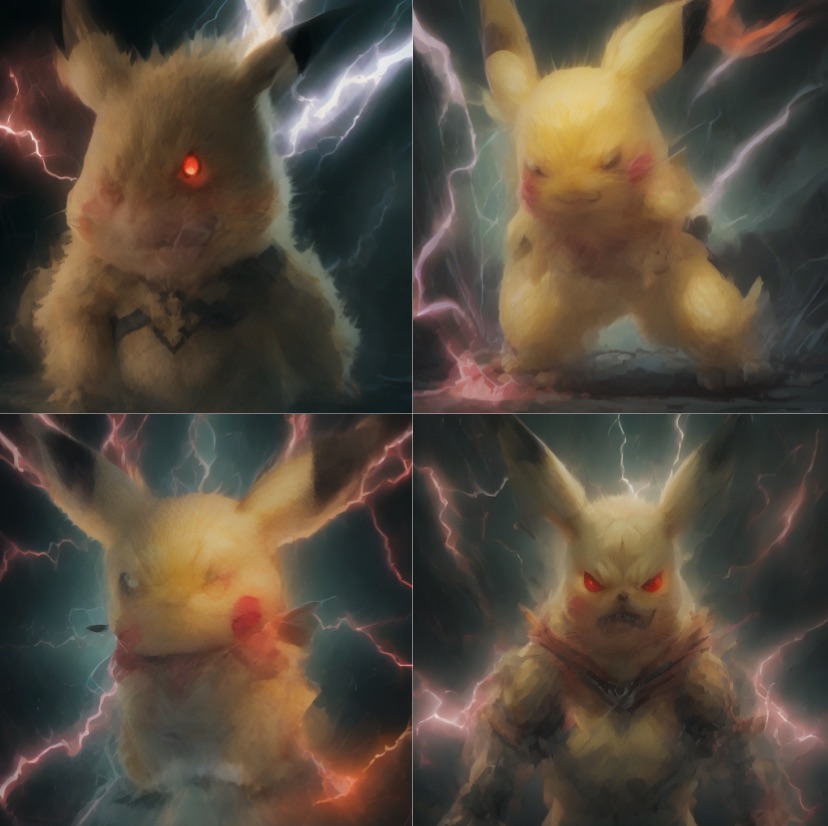} \\

        InstaFlow (1 step) & LCM (2 steps) \\
                &  \\
        \includegraphics[width=0.4\textwidth]{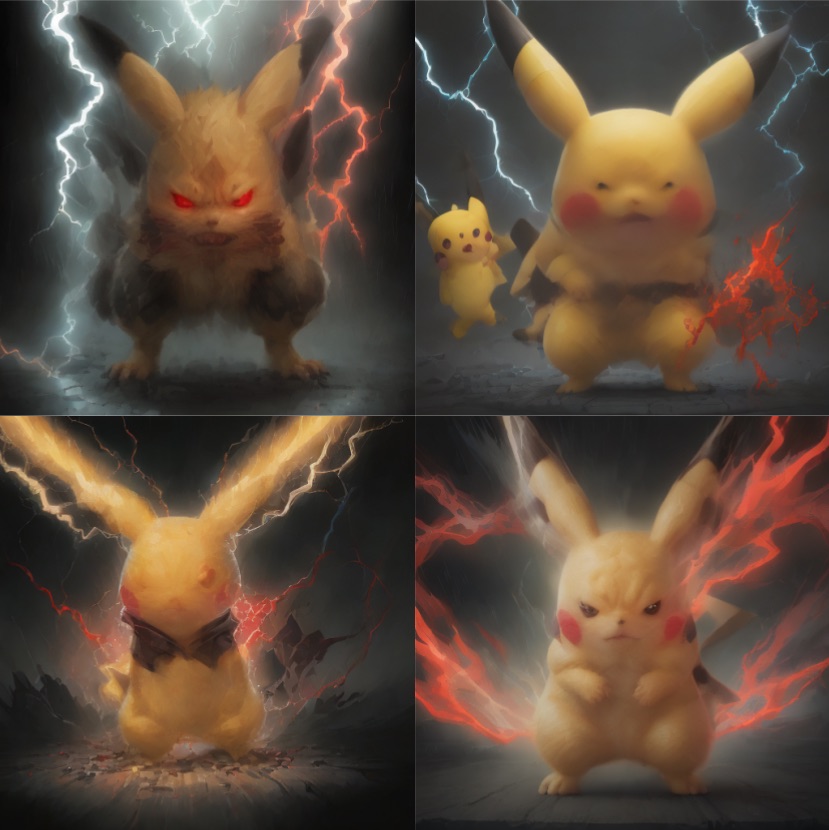} &
        \includegraphics[width=0.4\textwidth]{fig/app_compare/pika/ufo.jpeg} \\
        LCM (4 steps) & UFOGen (1 step) \\
    \end{tabular}
    }
    \caption{Prompt: \textit{A Pikachu with an angry expression and red eyes, with lightning around it, hyper realistic style.}}
    \label{tab:app_comparison_2}
\end{table*}

\begin{table*}[!t]
    \centering
    \scalebox{1.05}{
    \begin{tabular}{cc}
        \includegraphics[width=0.4\textwidth]{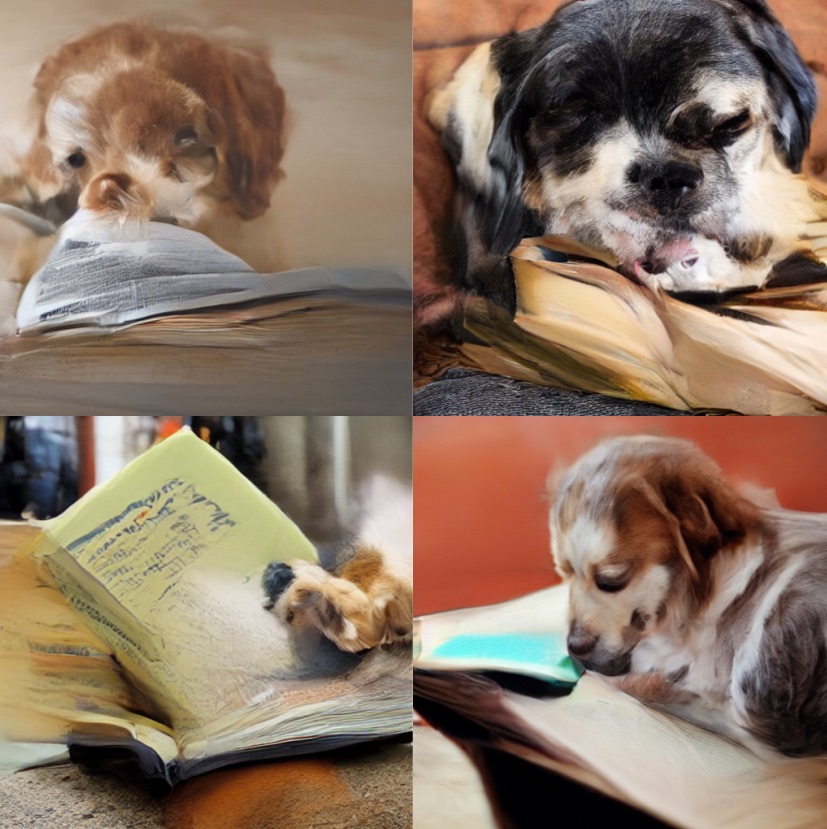} &
        \includegraphics[width=0.4\textwidth]{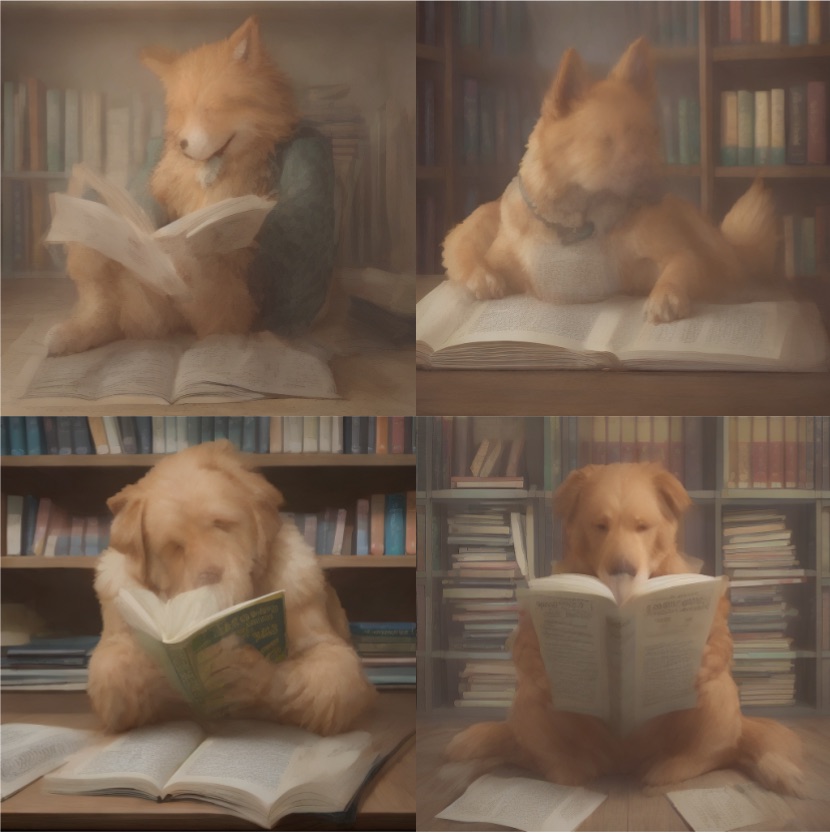} \\

        InstaFlow (1 step) & LCM (2 steps) \\
                &  \\
        \includegraphics[width=0.4\textwidth]{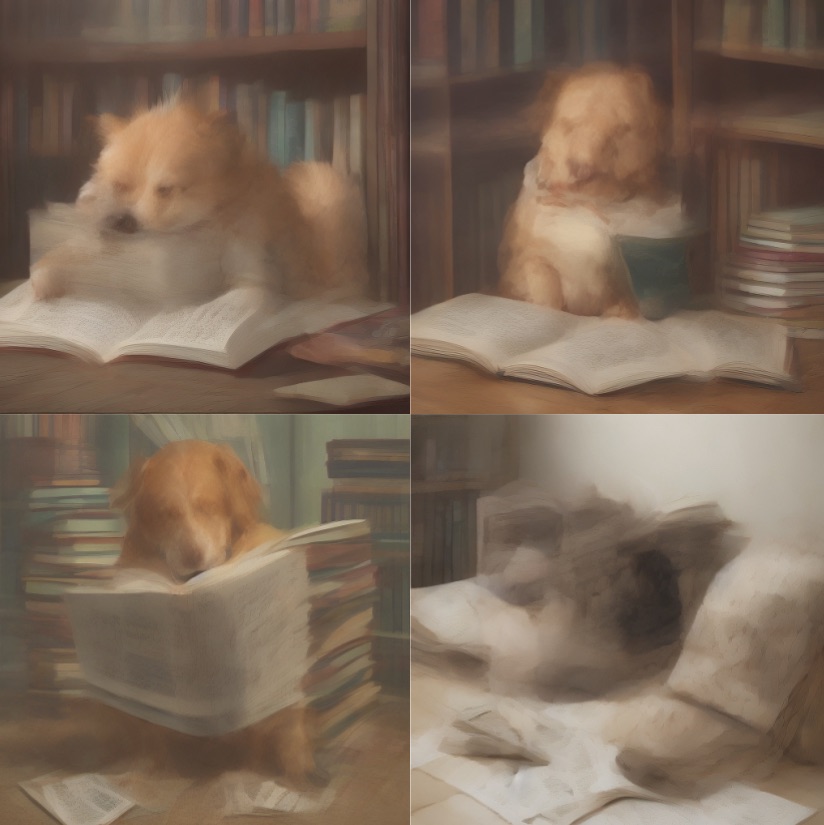} &
        \includegraphics[width=0.4\textwidth]{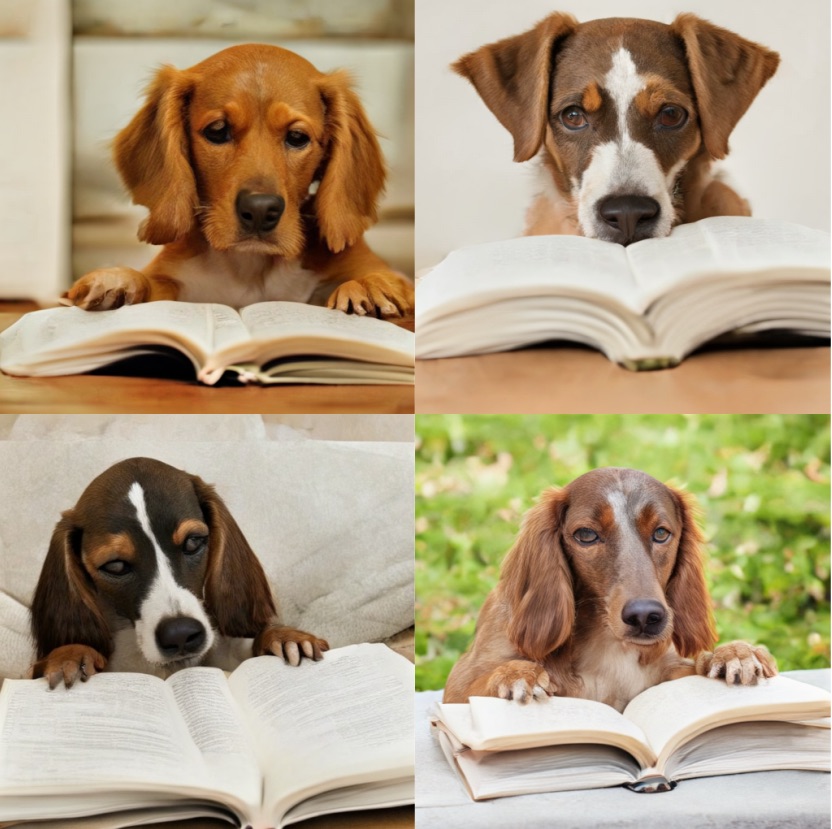} \\
        LCM (4 steps) & UFOGen (1 step) \\
    \end{tabular}
    }
    \caption{Prompt: \textit{A dog is reading a thick book.}}
    \label{tab:app_comparison_3}
\end{table*}

\begin{table*}[!t]
    \centering
    \scalebox{1.05}{
    \begin{tabular}{cc}
        \includegraphics[width=0.4\textwidth]{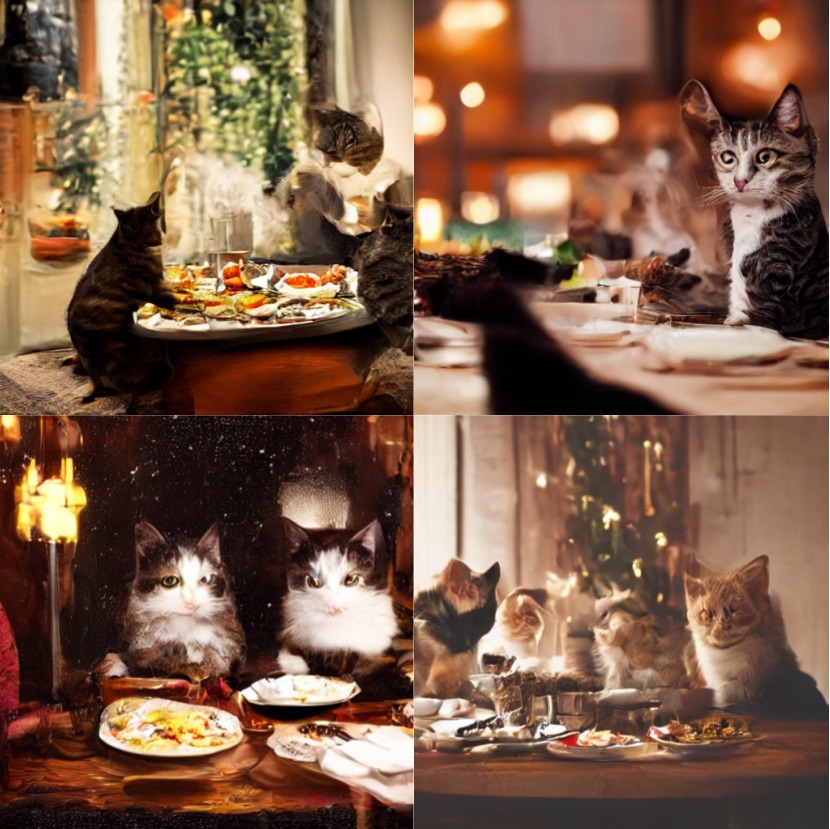} &
        \includegraphics[width=0.4\textwidth]{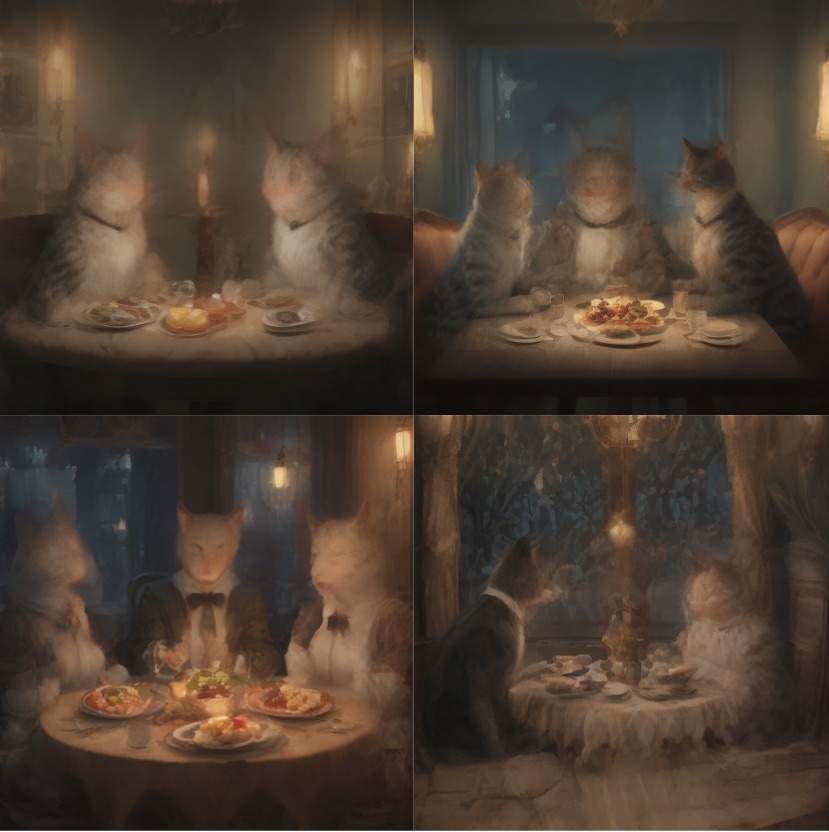} \\

        InstaFlow (1 step) & LCM (2 steps) \\
                &  \\
        \includegraphics[width=0.4\textwidth]{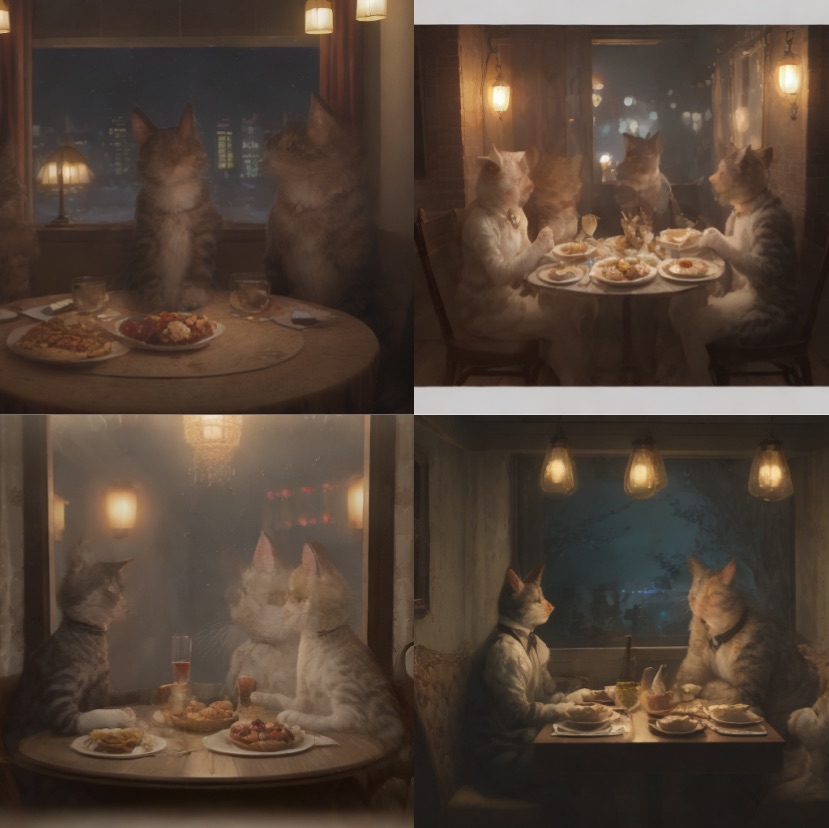} &
        \includegraphics[width=0.4\textwidth]{fig/app_compare/cat/ufo.jpeg} \\
        LCM (4 steps) & UFOGen (1 step) \\
    \end{tabular}
    }
    \caption{Prompt: \textit{Three cats having dinner at a table at new years eve, cinematic shot, 8k.}}
    \label{tab:app_comparison_4}
\end{table*}

\begin{table*}[!t]
    \centering
    \scalebox{1.05}{
    \begin{tabular}{cc}
        \includegraphics[width=0.4\textwidth]{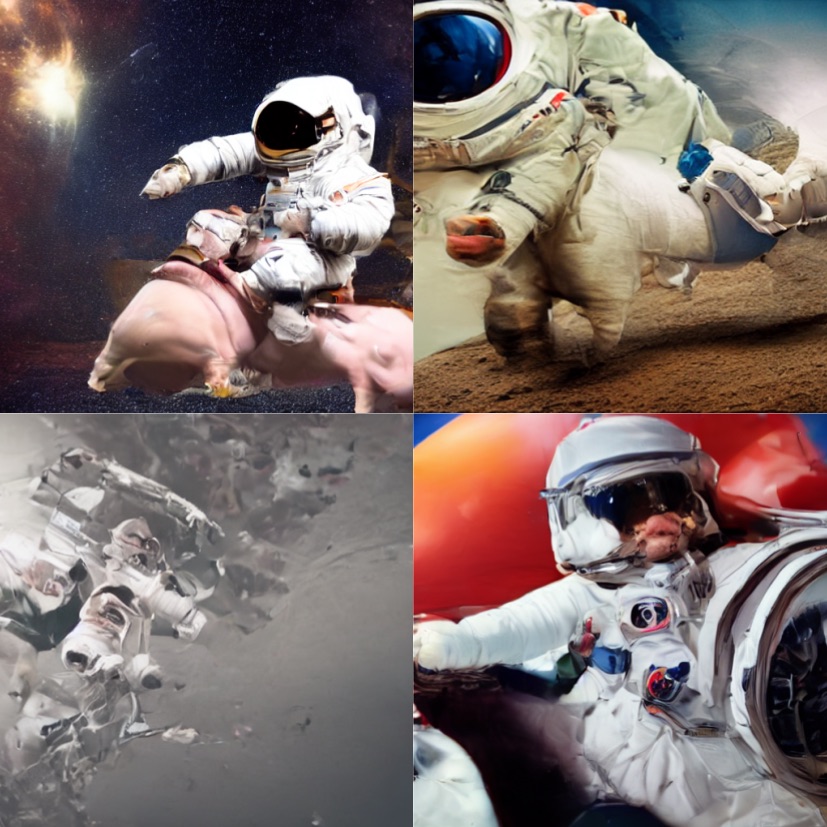} &
        \includegraphics[width=0.4\textwidth]{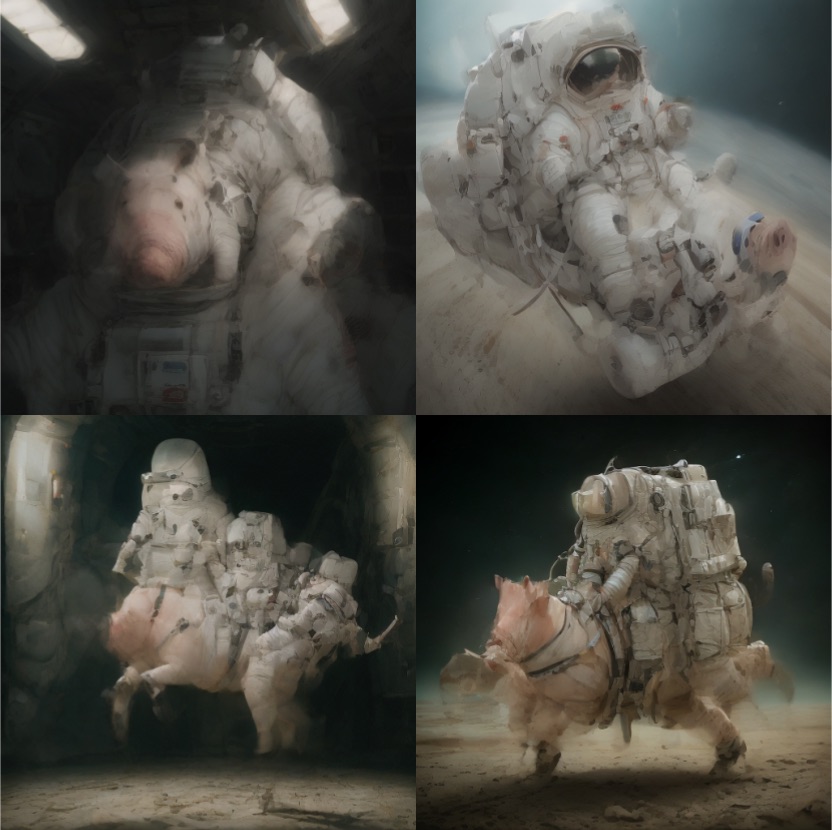} \\

        InstaFlow (1 step) & LCM (2 steps) \\
                &  \\
        \includegraphics[width=0.4\textwidth]{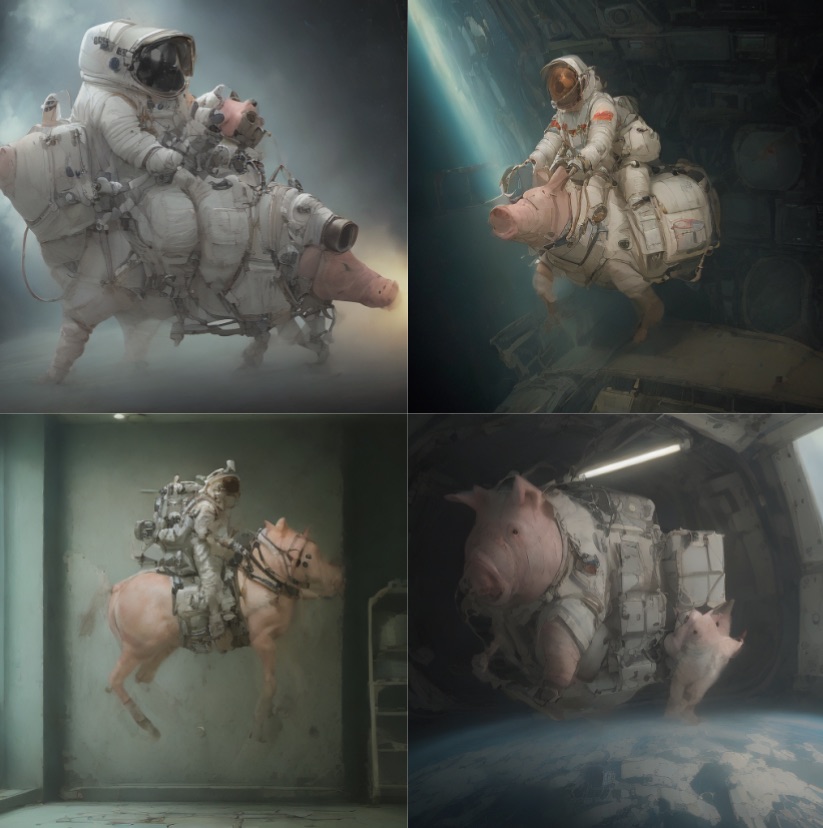} &
        \includegraphics[width=0.405\textwidth]{fig/app_compare/pig/ufo.jpeg} \\
        LCM (4 steps) & UFOGen (1 step) \\
    \end{tabular}
    }
    \caption{Prompt: \textit{An astronaut riding a pig, highly realistic dslr photo, cinematic shot.}}
    \label{tab:app_comparison_5}
\end{table*}

\begin{table*}[!t]
    \centering
    \scalebox{1.05}{
    \begin{tabular}{cc}
        \includegraphics[width=0.4\textwidth]{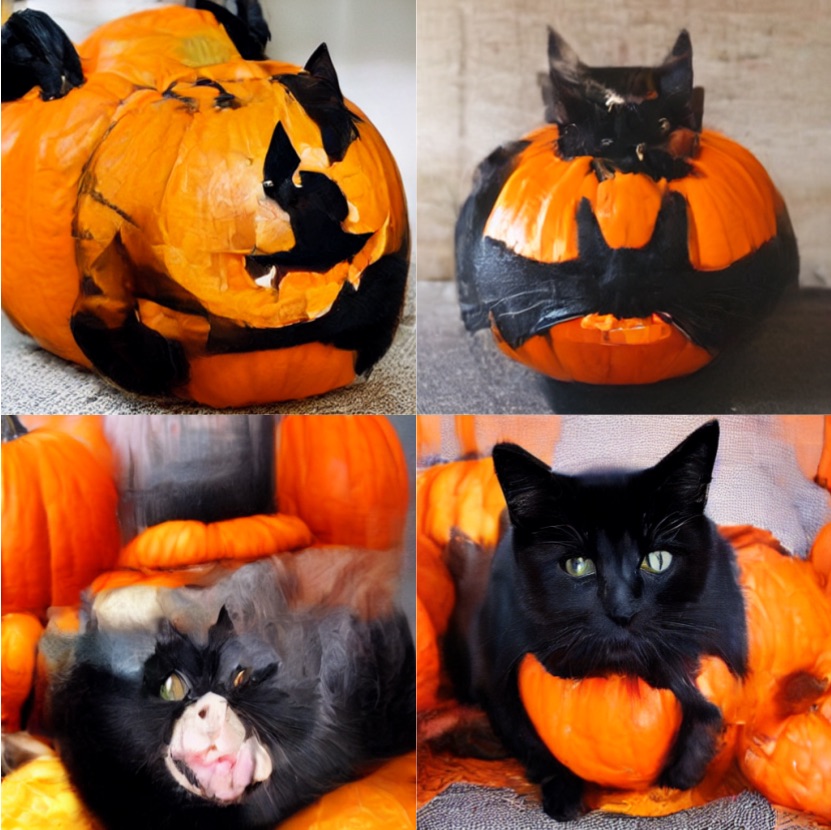} &
        \includegraphics[width=0.4\textwidth]{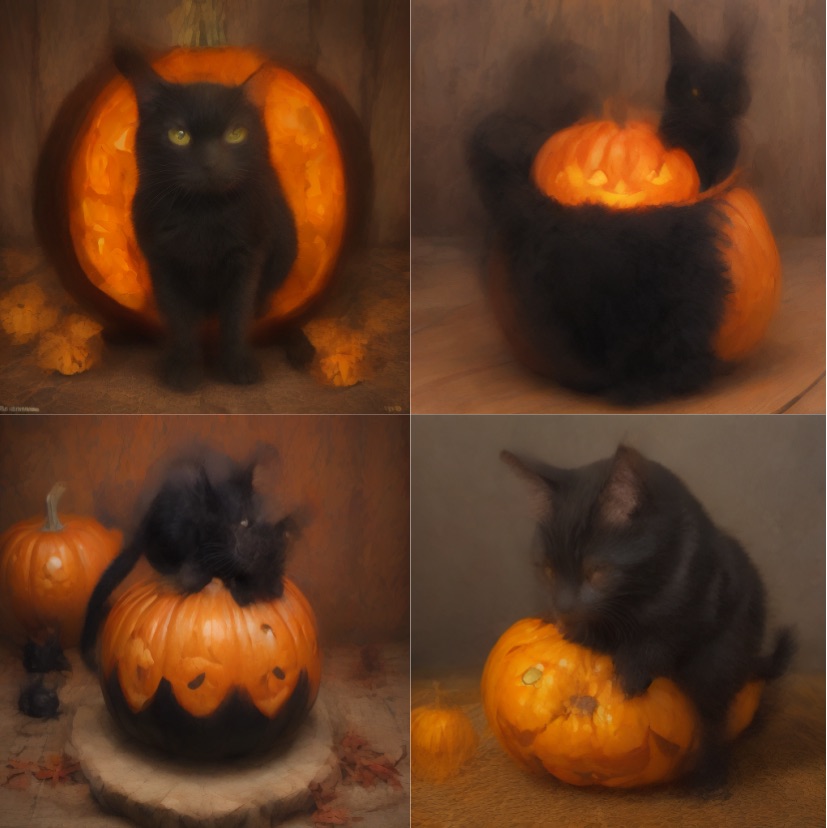} \\

        InstaFlow (1 step) & LCM (2 steps) \\
                &  \\
        \includegraphics[width=0.4\textwidth]{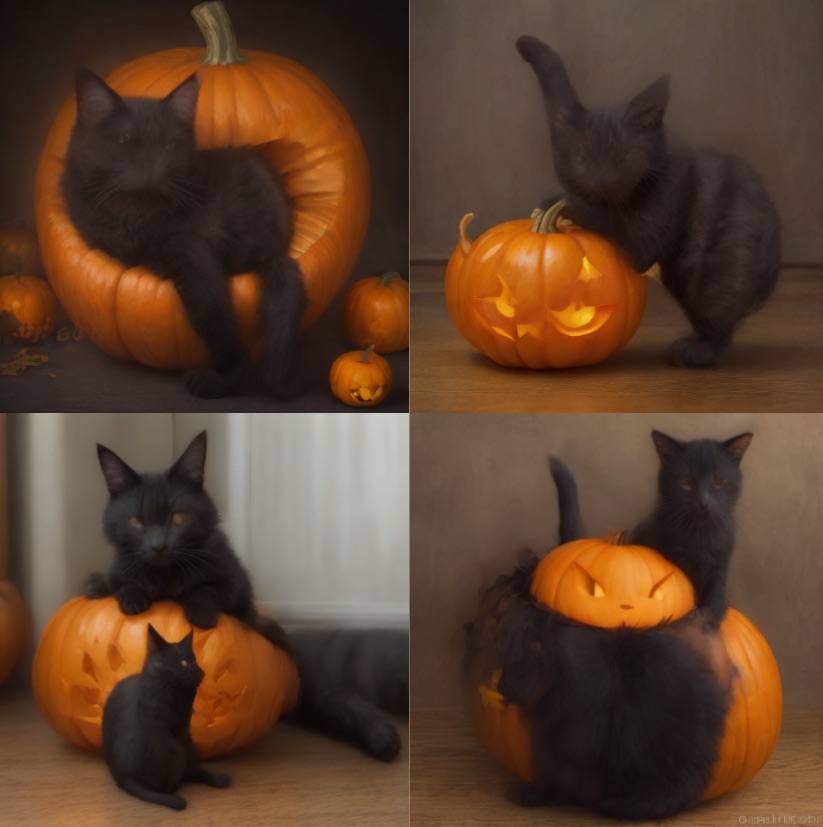} &
        \includegraphics[width=0.405\textwidth]{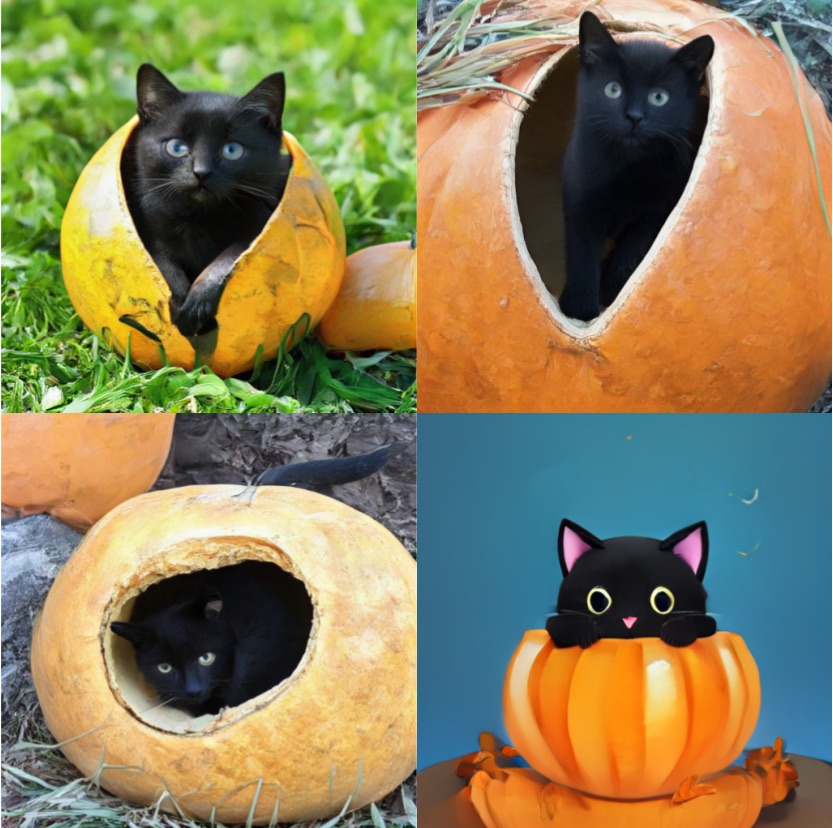} \\
        LCM (4 steps) & UFOGen (1 step) \\
    \end{tabular}
    }
    \caption{Prompt: \textit{A cute black cat inside of a pumpkin.}}
    \label{tab:app_comparison_6}
\end{table*}

\begin{table*}[!t]
    \centering
    \scalebox{1.05}{
    \begin{tabular}{cc}
        \includegraphics[width=0.4\textwidth]{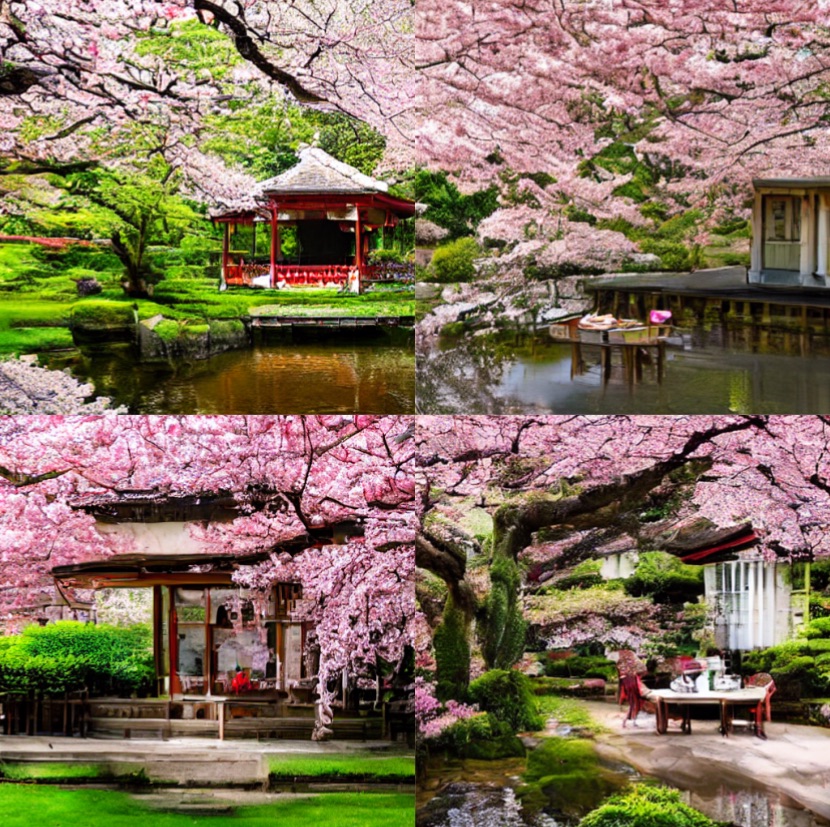} &
        \includegraphics[width=0.4\textwidth]{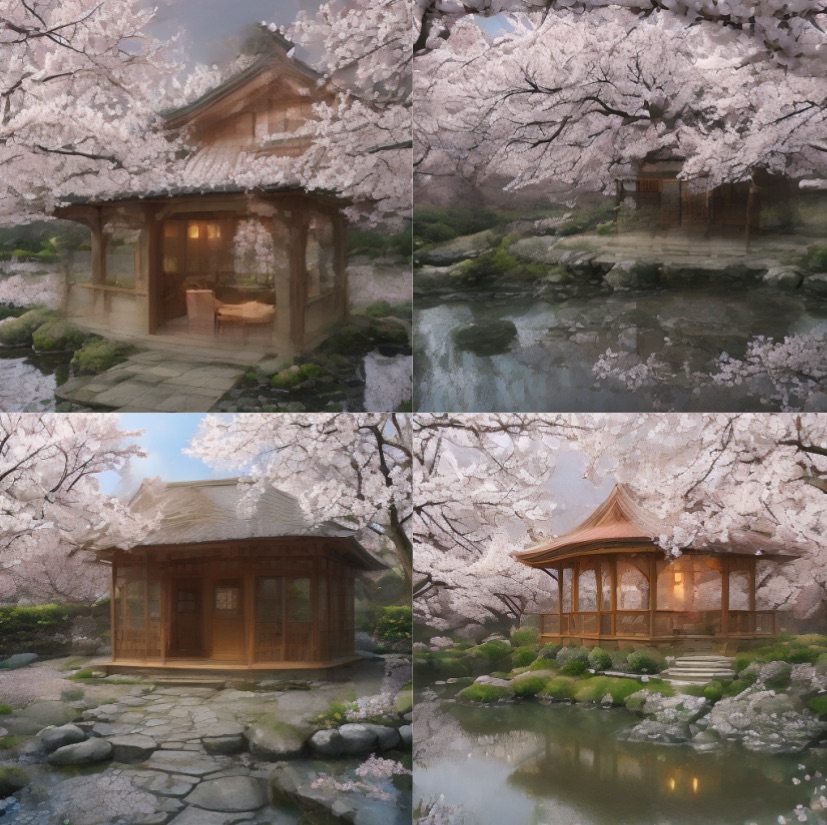} \\

        InstaFlow (1 step) & LCM (2 steps) \\
                &  \\
        \includegraphics[width=0.4\textwidth]{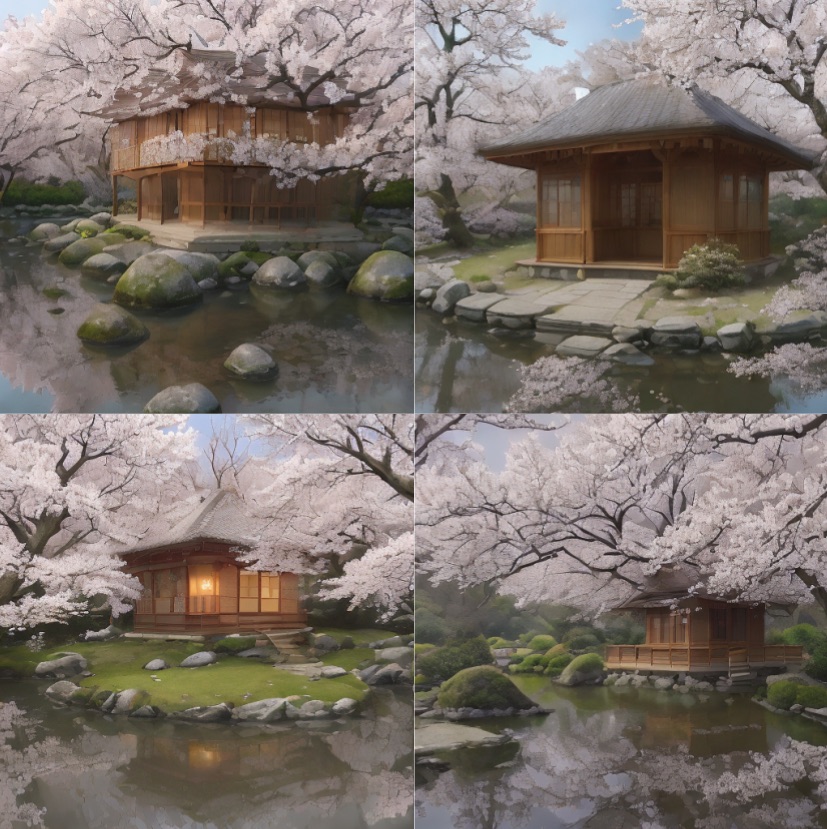} &
        \includegraphics[width=0.4\textwidth]{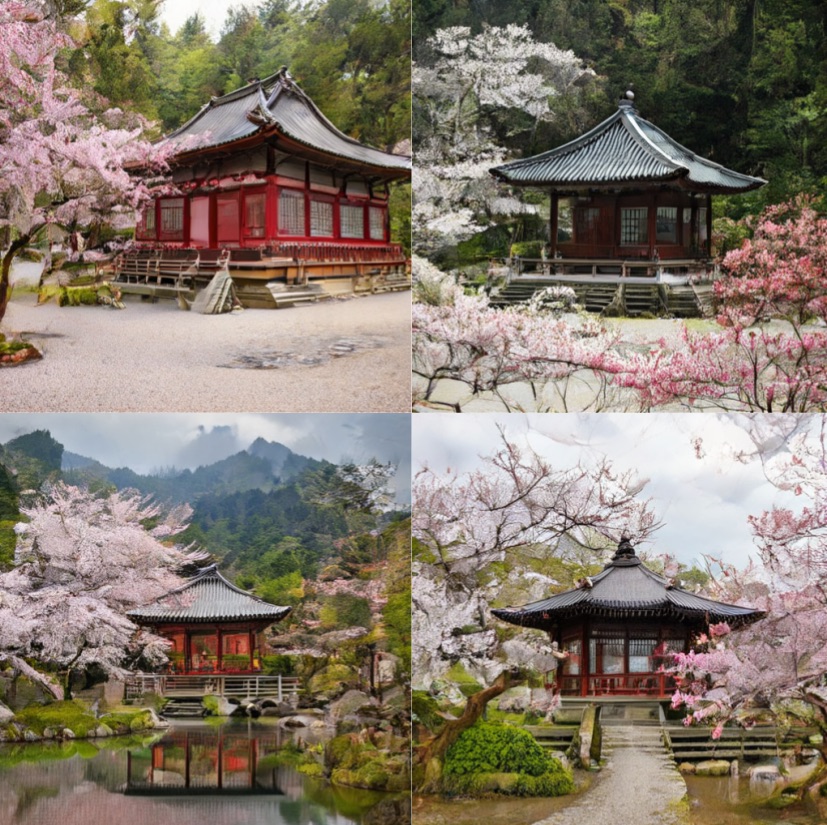} \\
        LCM (4 steps) & UFOGen (1 step) \\
    \end{tabular}
    }
    \caption{Prompt: \textit{A traditional tea house in a tranquil garden with blooming cherry blossom trees.}}
    \label{tab:app_comparison_7}
\end{table*}

\begin{table*}[!t]
    \centering
    \scalebox{1.05}{
    \begin{tabular}{cc}
        \includegraphics[width=0.4\textwidth]{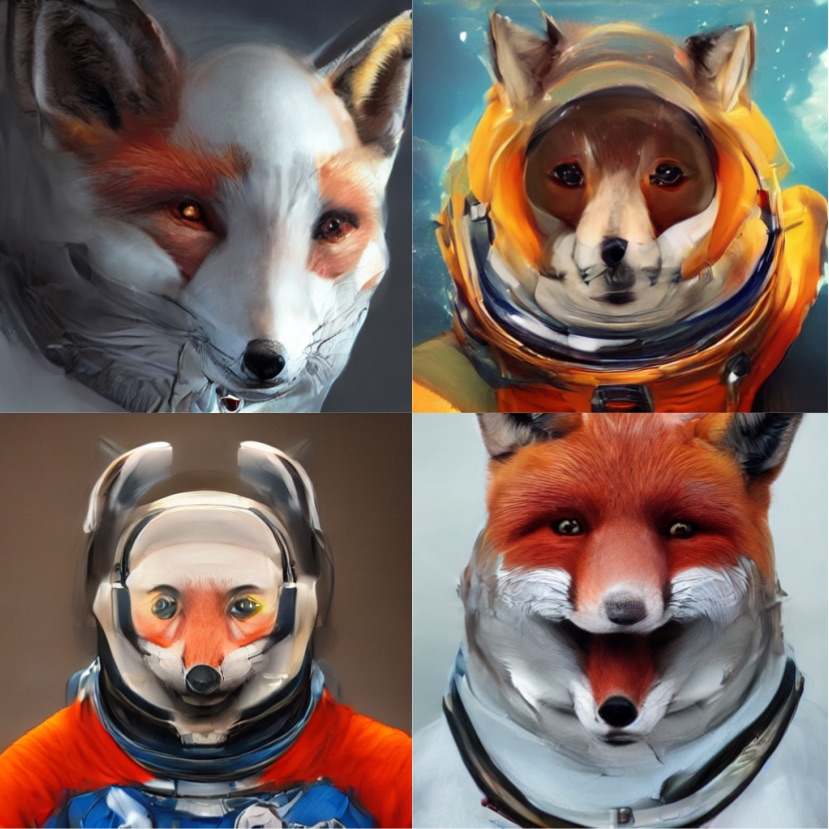} &
        \includegraphics[width=0.4\textwidth]{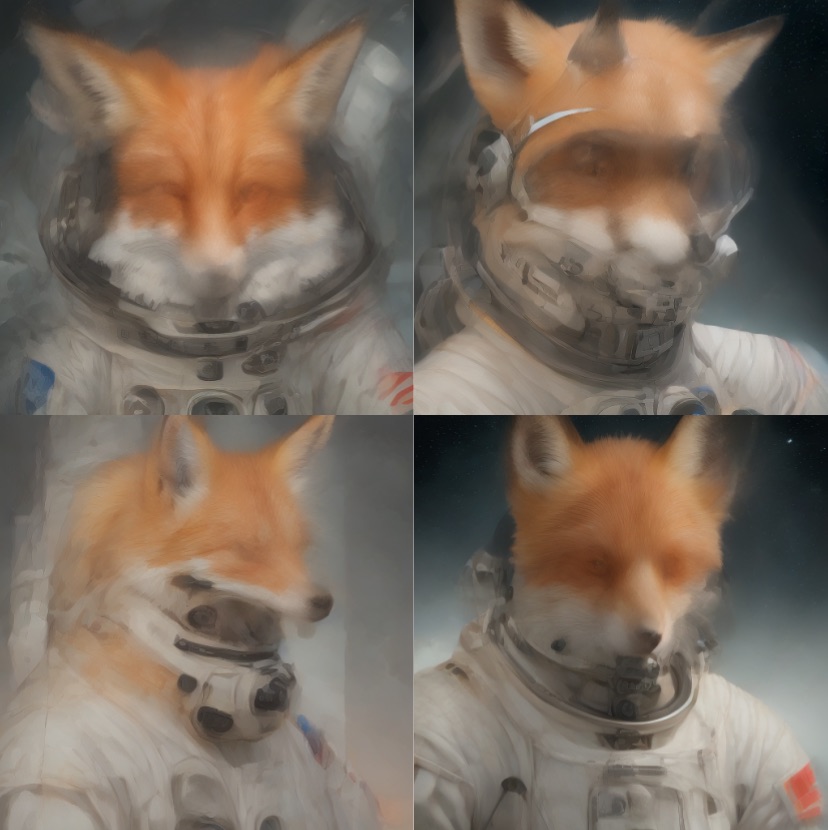} \\

        InstaFlow (1 step) & LCM (2 steps) \\
                &  \\
        \includegraphics[width=0.4\textwidth]{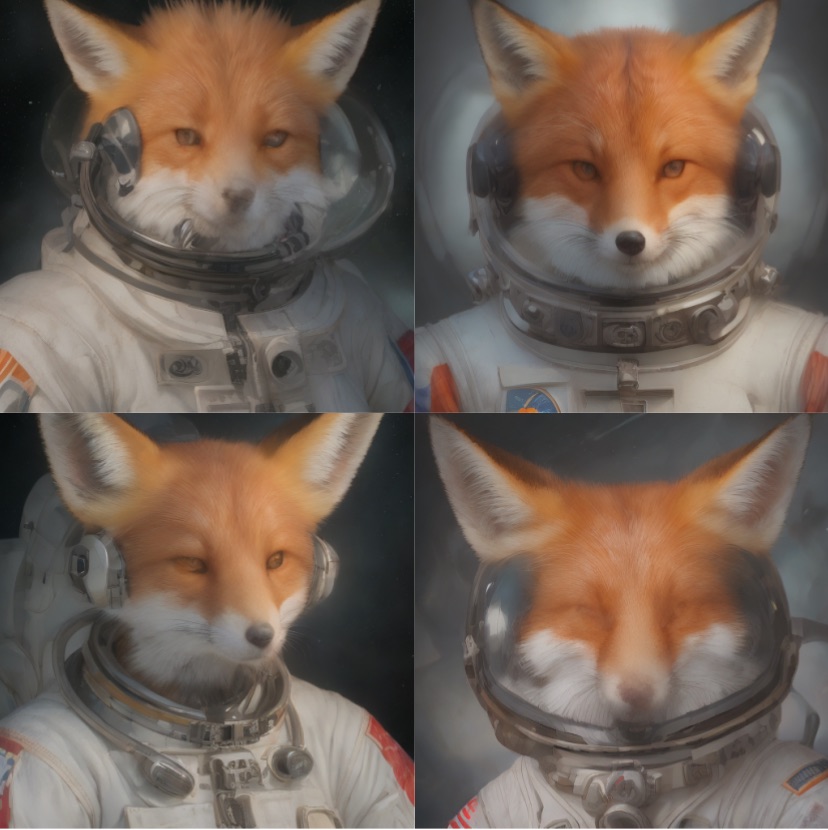} &
        \includegraphics[width=0.4\textwidth]{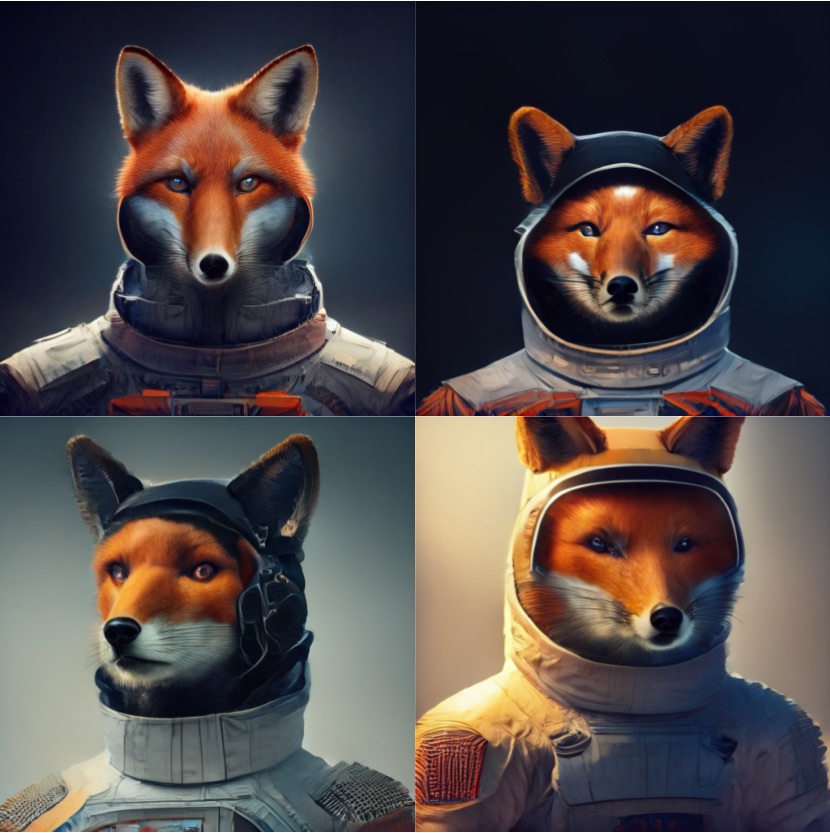} \\
        LCM (4 steps) & UFOGen (1 step) \\
    \end{tabular}
    }
    \caption{Prompt: \textit{Hyperrealistic photo of a fox astronaut, perfect face, artstation.}}
    \label{tab:app_comparison_8}
\end{table*}

%% file: sec/figure_tables/new_lcm_results.tex
\setlength{\tabcolsep}{2pt} 
\renewcommand{\arraystretch}{1.0} 
\begin{table*}[!t]
    \centering
    \scalebox{1.05}{
    \begin{tabular}{cc}
        Updated LCM (2 steps) & Updated LCM (4 steps) \\
        \includegraphics[width=0.4\textwidth]{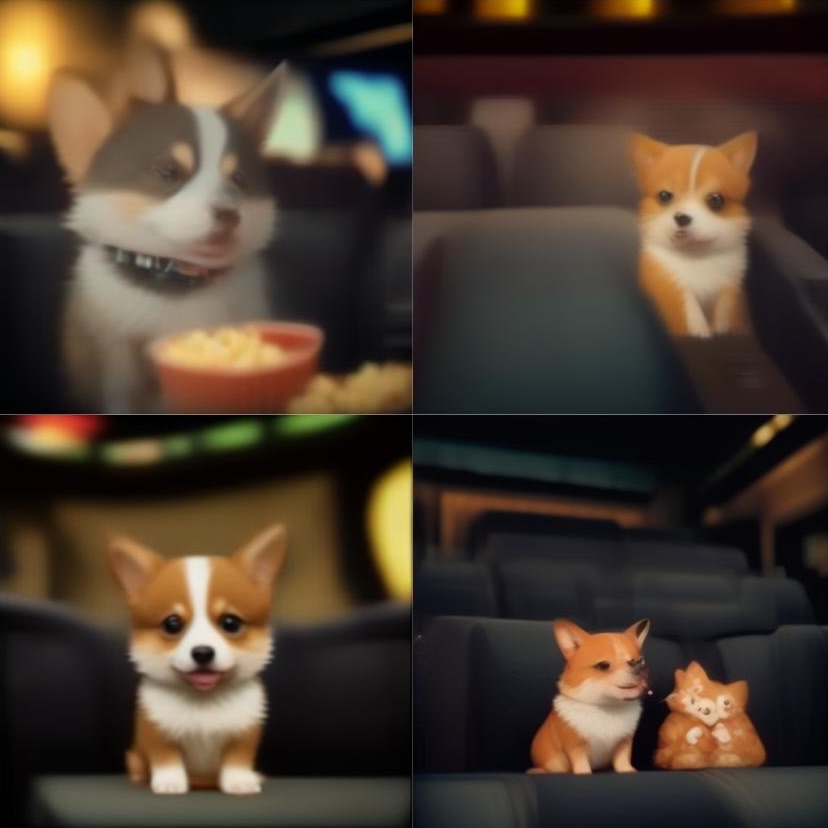} &
        \includegraphics[width=0.4\textwidth]{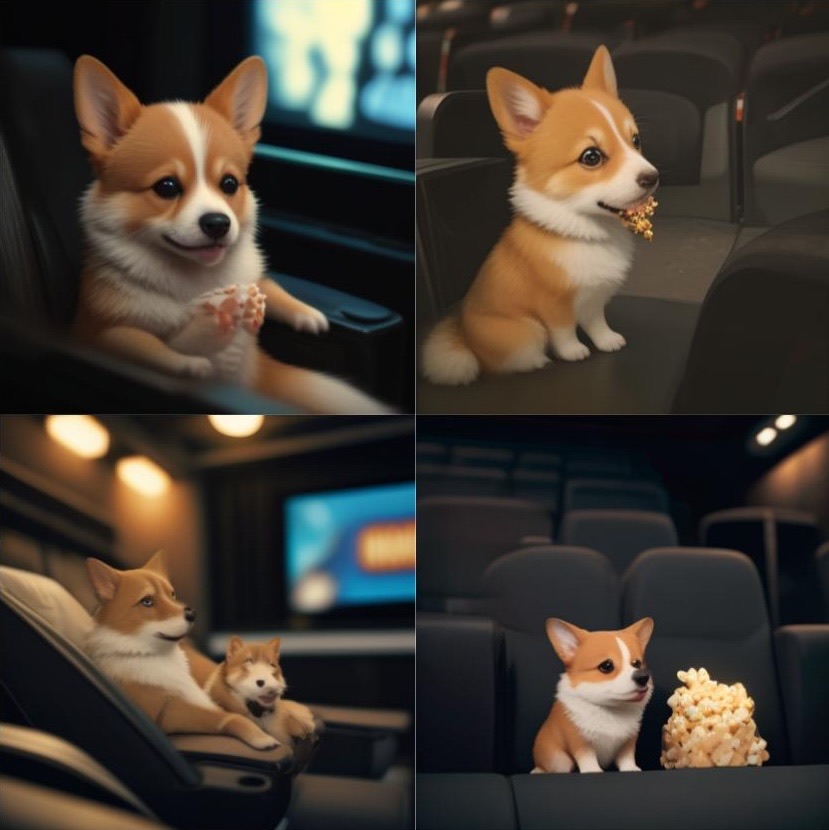} \\
        \multicolumn{2}{c}{\textit{Cute small corgi sitting in a movie theater eating popcorn, unreal engine.}} \\

        \includegraphics[width=0.4\textwidth]{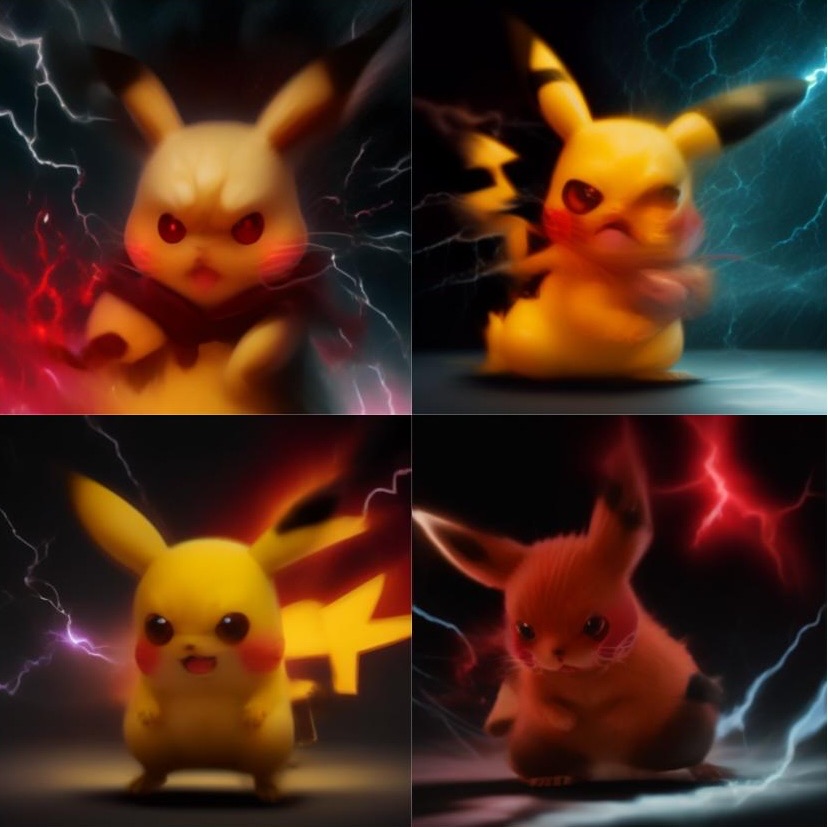} &
        \includegraphics[width=0.4\textwidth]{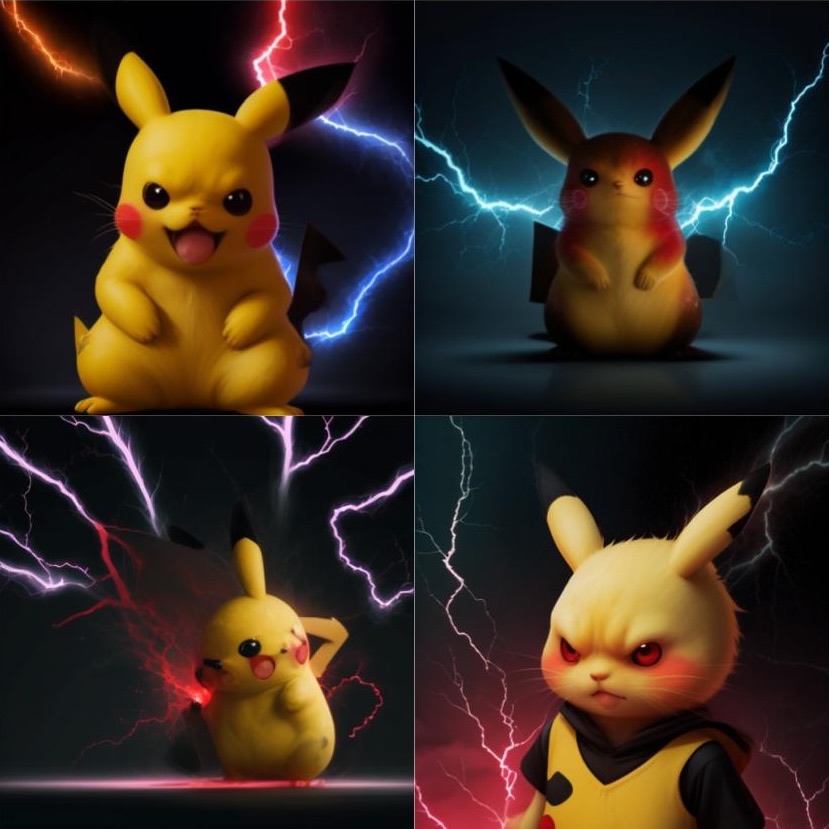} \\
        \multicolumn{2}{c}{\textit{A Pikachu with an angry expression and red eyes, with lightning around it, hyper realistic style.}} 
    \end{tabular}
    }
    \caption{Qualitative results of updated LCM model.}
    \label{tab: app new lcm 1}
\end{table*}

\begin{table*}[!t]
    \centering
    \scalebox{1.05}{
    \begin{tabular}{cc}
        Updated LCM (2 steps) & Updated LCM (4 steps) \\
        \includegraphics[width=0.4\textwidth]{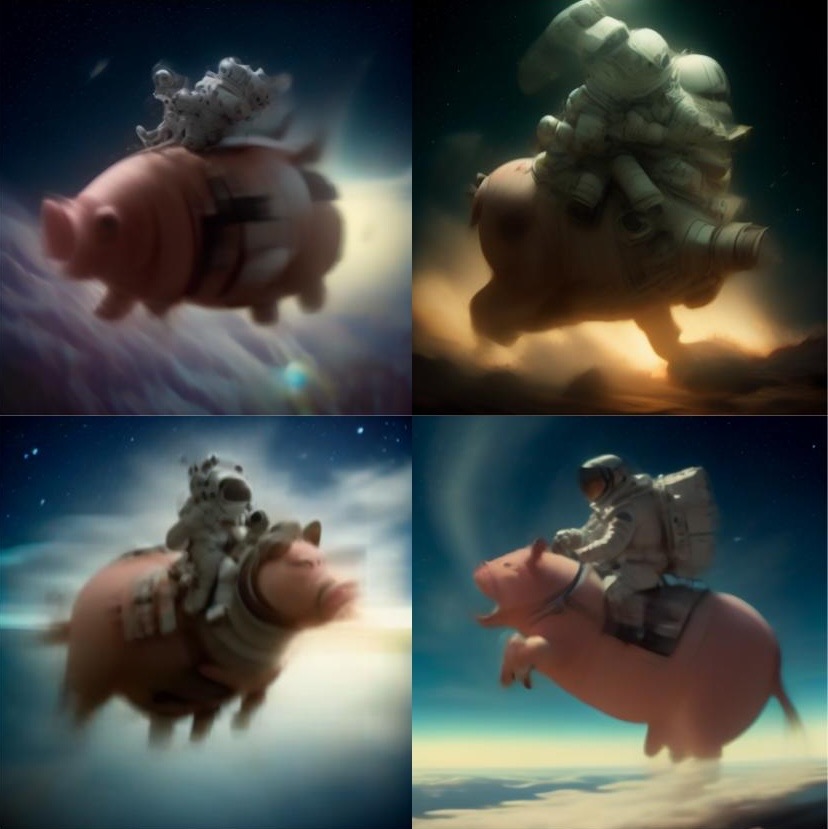} &
        \includegraphics[width=0.4\textwidth]{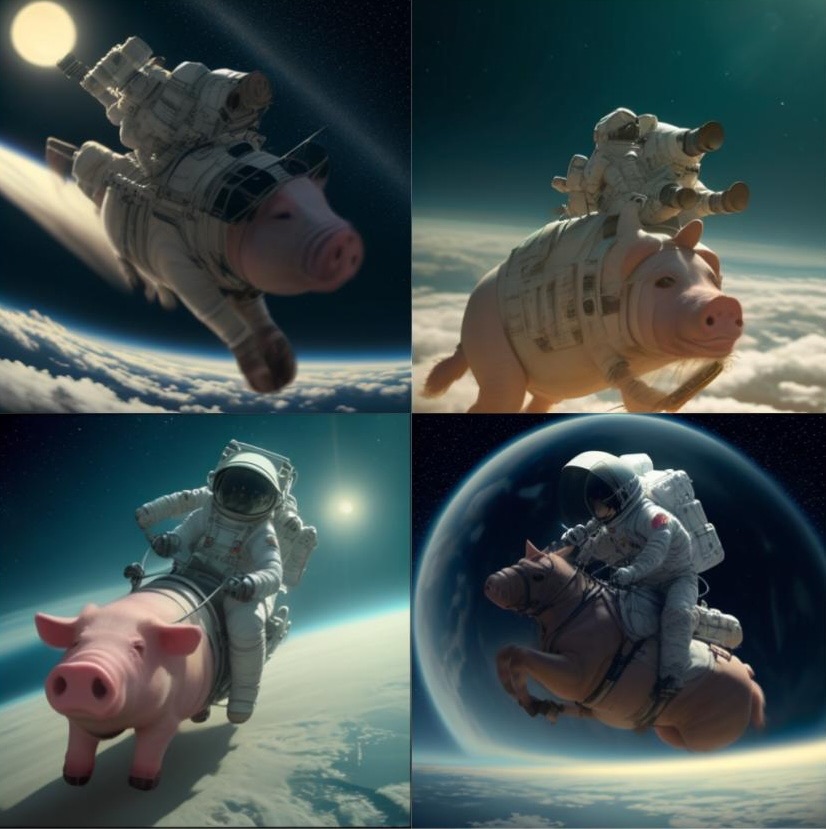} \\
        \multicolumn{2}{c}{\textit{An astronaut riding a pig, highly realistic dslr photo, cinematic shot.}} \\

        \includegraphics[width=0.4\textwidth]{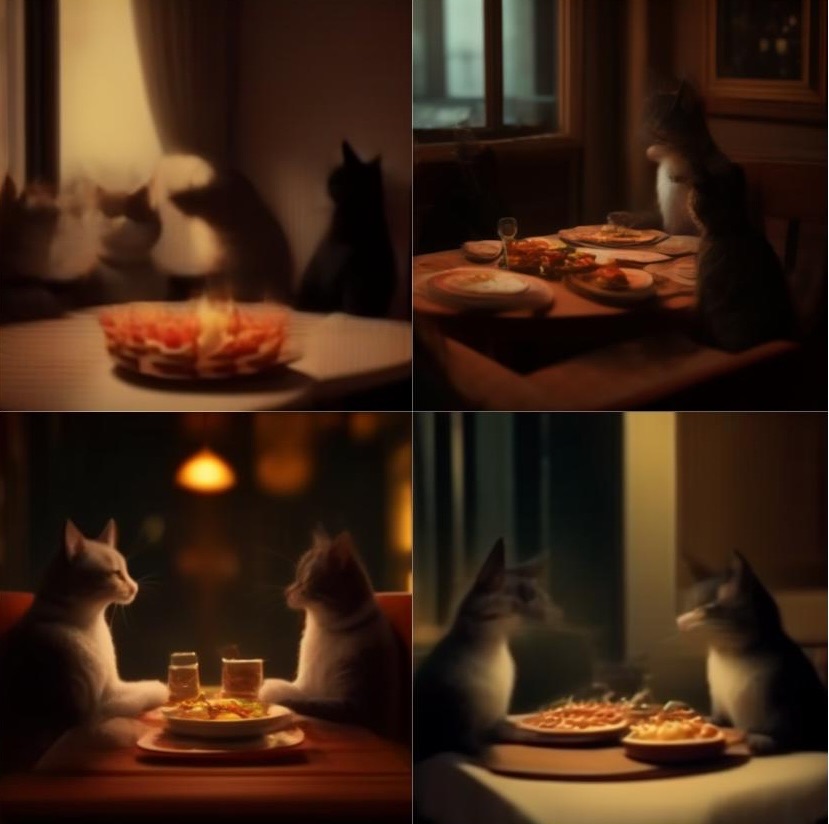} &
        \includegraphics[width=0.4\textwidth]{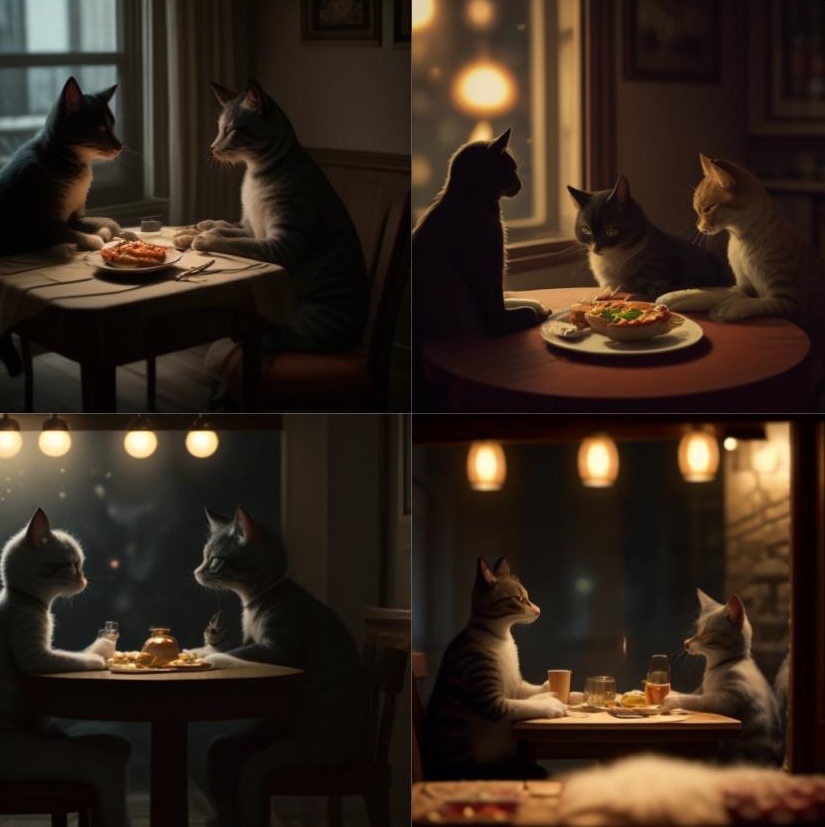} \\
        \multicolumn{2}{c}{\textit{Three cats having dinner at a table at new years eve, cinematic shot, 8k.}} 
    \end{tabular}
    }
    \caption{Qualitative results of updated LCM model.}
    \label{tab: app new lcm 2}
\end{table*}

%% file: sec/figure_tables/app_extra_results.tex
\setlength{\tabcolsep}{2pt} 
\renewcommand{\arraystretch}{1.0} 
\begin{table*}[!t]
    \centering
    \begin{tabular}{cccc}
        \includegraphics[width=0.21\textwidth]{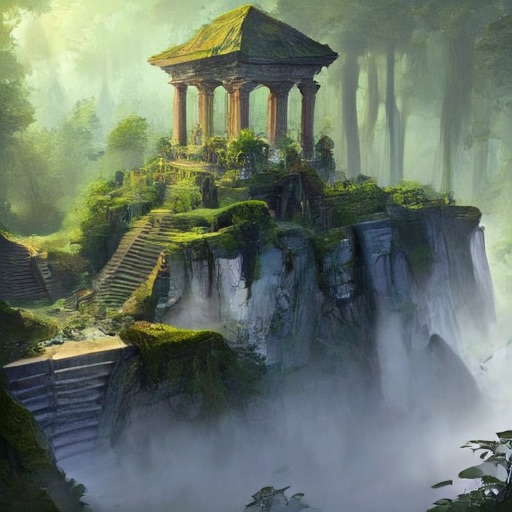} &  
        \includegraphics[width=0.21\textwidth]{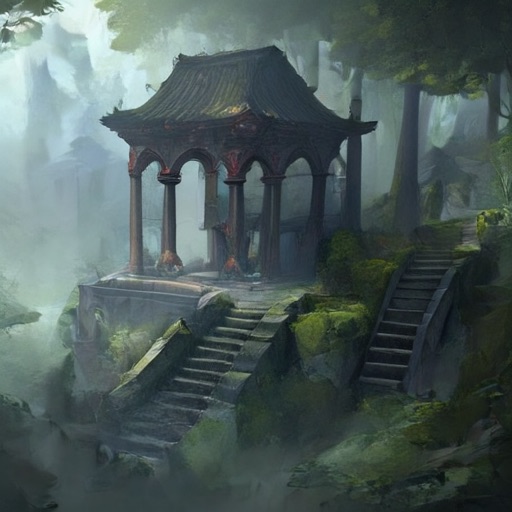} &
        \includegraphics[width=0.21\textwidth]{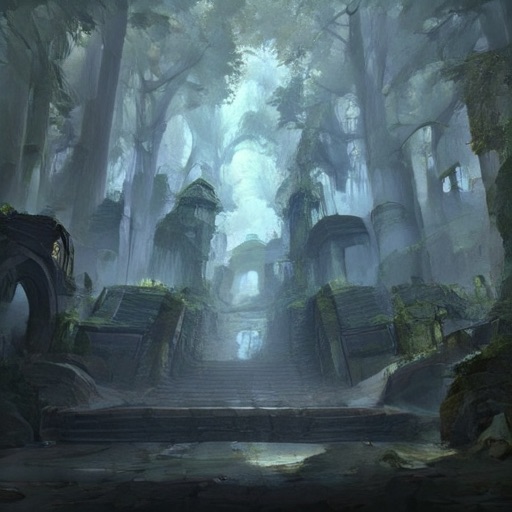} &
        \includegraphics[width=0.21\textwidth]{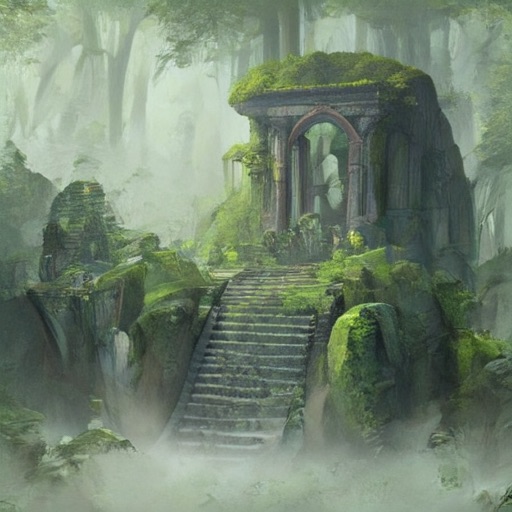} \\
        \multicolumn{4}{c}{\textit{Temple ruins in forest, stairs, mist, concept art.}} \\

        \includegraphics[width=0.21\textwidth]{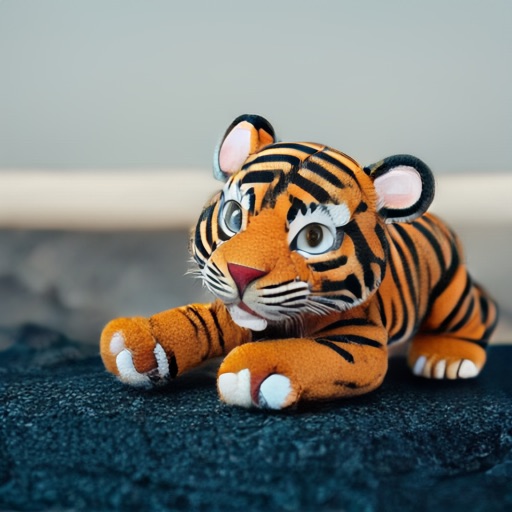} &  
        \includegraphics[width=0.21\textwidth]{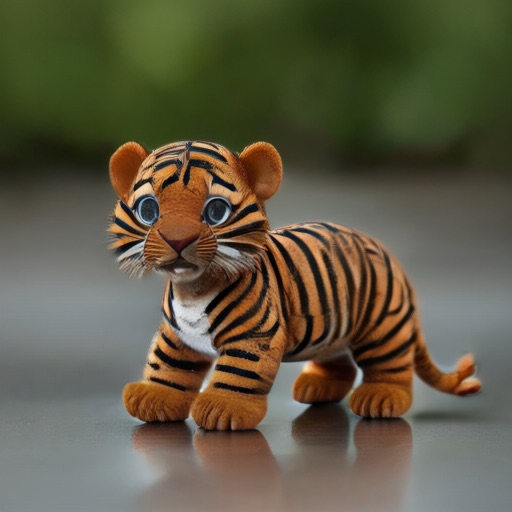} &
        \includegraphics[width=0.21\textwidth]{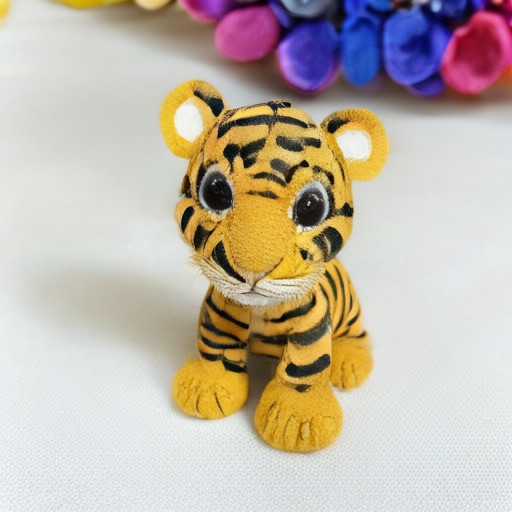} &
        \includegraphics[width=0.21\textwidth]{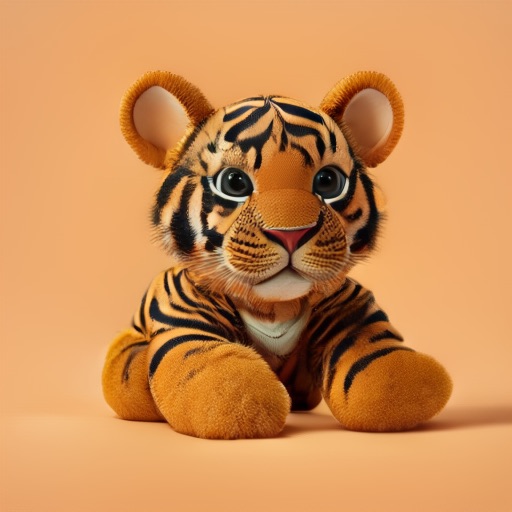} \\
        \multicolumn{4}{c}{\textit{Cute toy tiger made of suede, geometric accurate, intricate details, cinematic.}} \\

        \includegraphics[width=0.21\textwidth]{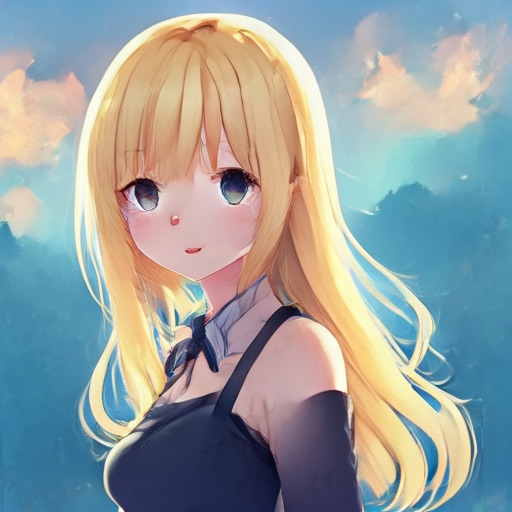} &  
        \includegraphics[width=0.21\textwidth]{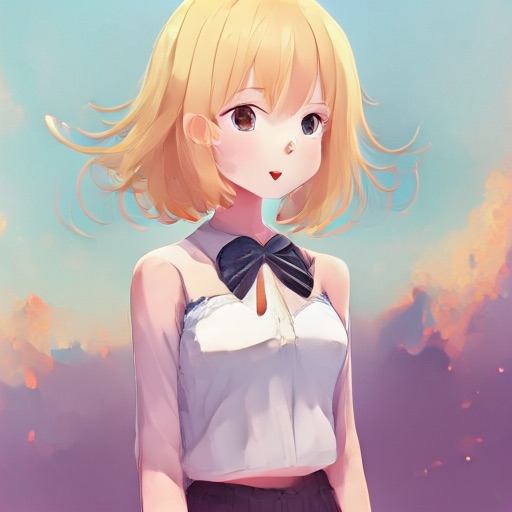} &
        \includegraphics[width=0.21\textwidth]{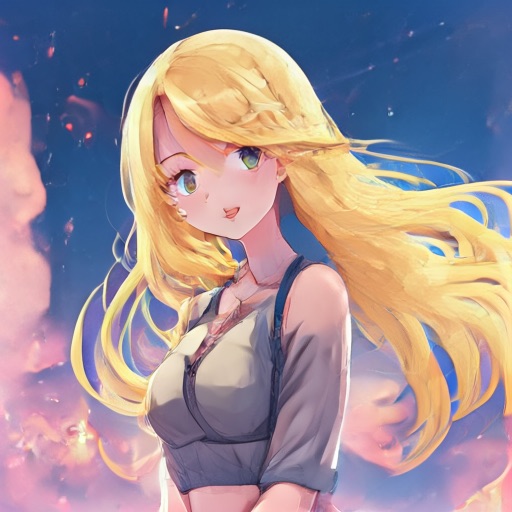} &
        \includegraphics[width=0.21\textwidth]{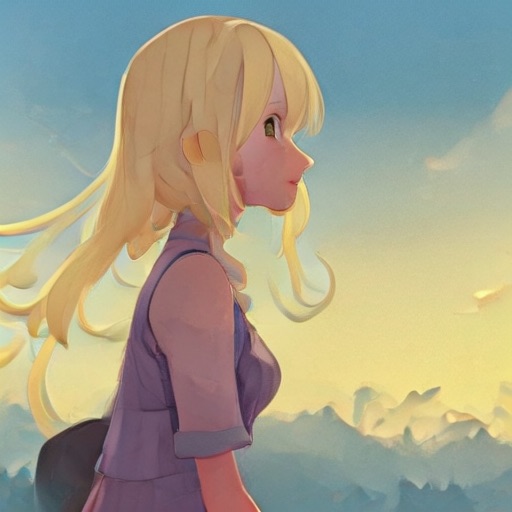} \\
        \multicolumn{4}{c}{\textit{Cute girl, crop-top, blond hair, animation key art feminine mid shot.}} \\

        \includegraphics[width=0.21\textwidth]{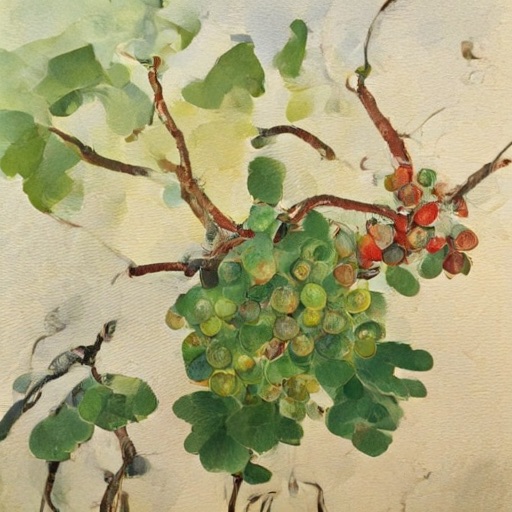} &  
        \includegraphics[width=0.21\textwidth]{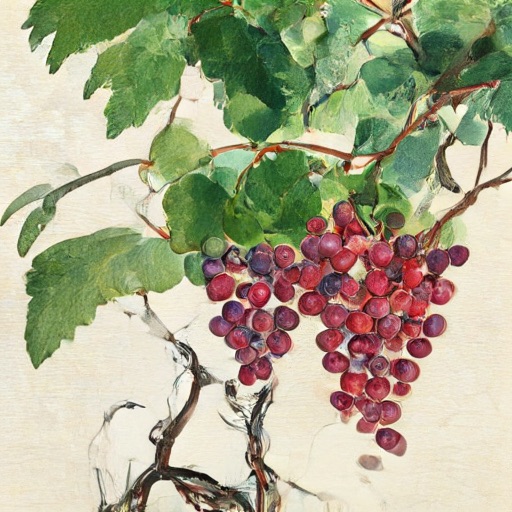} &
        \includegraphics[width=0.21\textwidth]{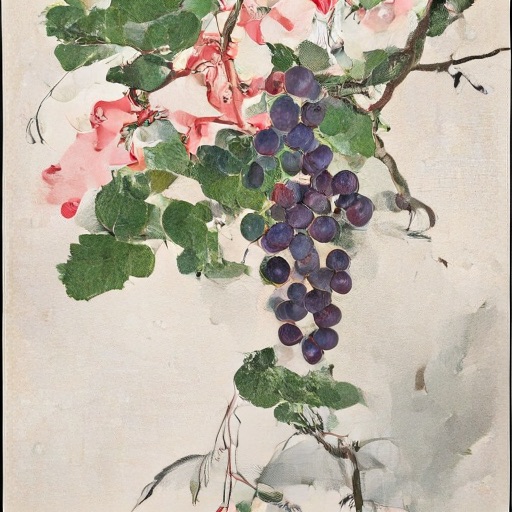} &
        \includegraphics[width=0.21\textwidth]{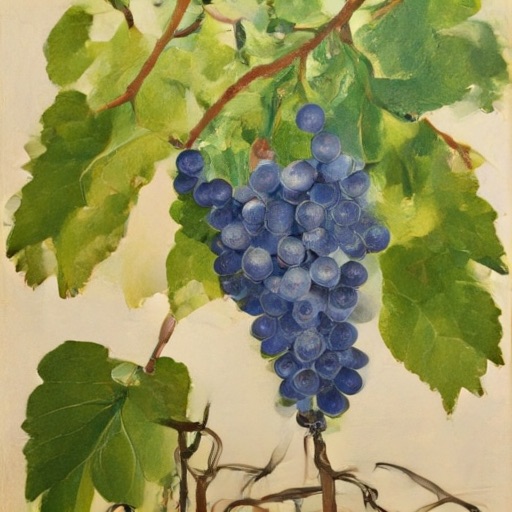} \\
        \multicolumn{4}{c}{\textit{Chinese painting of grapes.}} \\

        \includegraphics[width=0.21\textwidth]{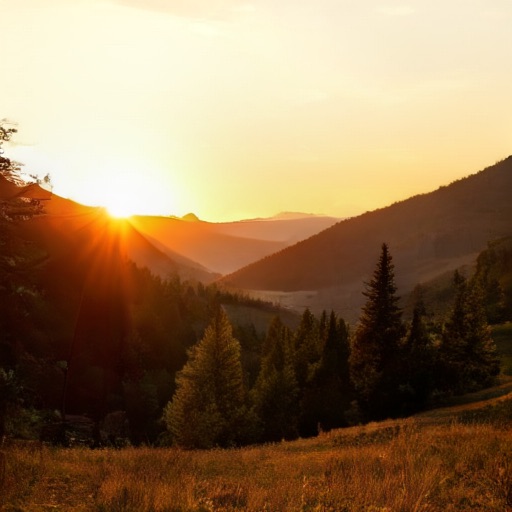} &  
        \includegraphics[width=0.21\textwidth]{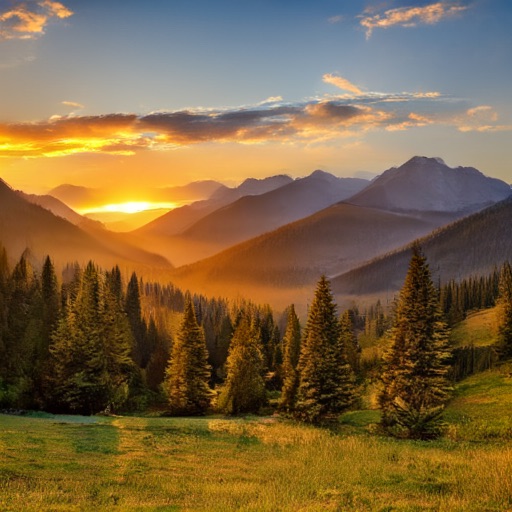} &
        \includegraphics[width=0.21\textwidth]{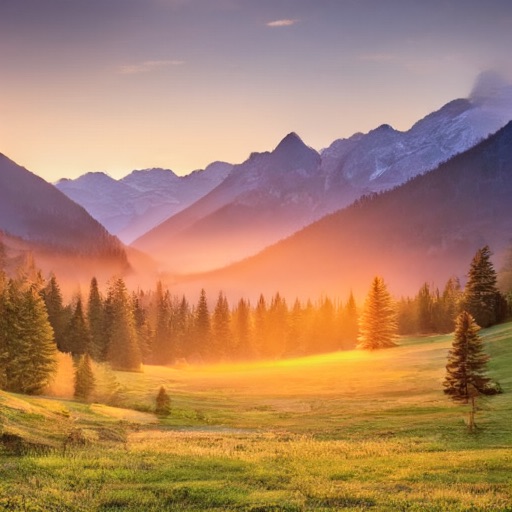} &
        \includegraphics[width=0.21\textwidth]{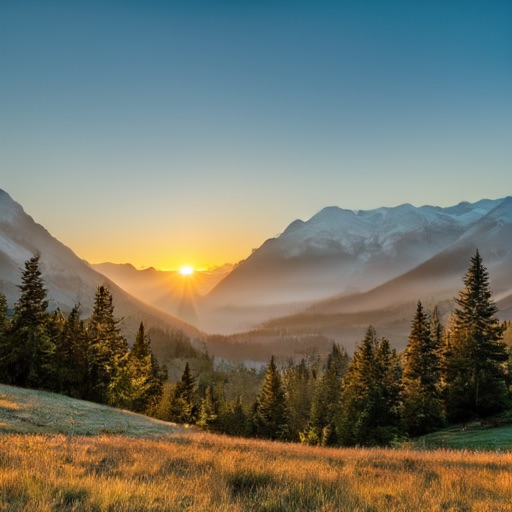} \\ 
        \multicolumn{4}{c}{\textit{Sunset in a valley with trees and mountains.}} \\
    \end{tabular}
    \caption{Additional qualitative results of UFOGen. Zoom-in for better viewing.}
    \label{tab:app_extra_1}
\end{table*}

\begin{table*}[!t]
    \centering
    \begin{tabular}{cccc}
        \includegraphics[width=0.21\textwidth]{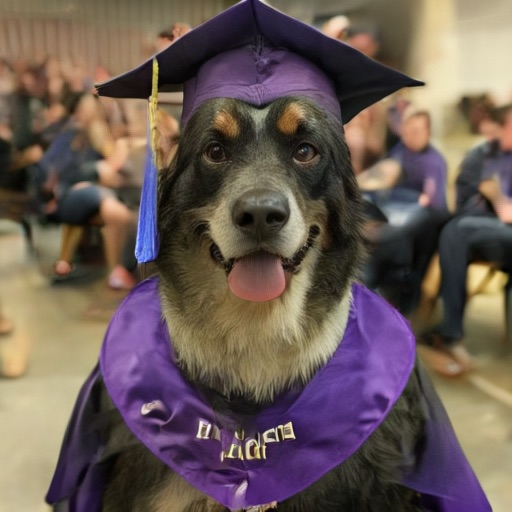} &  
        \includegraphics[width=0.21\textwidth]{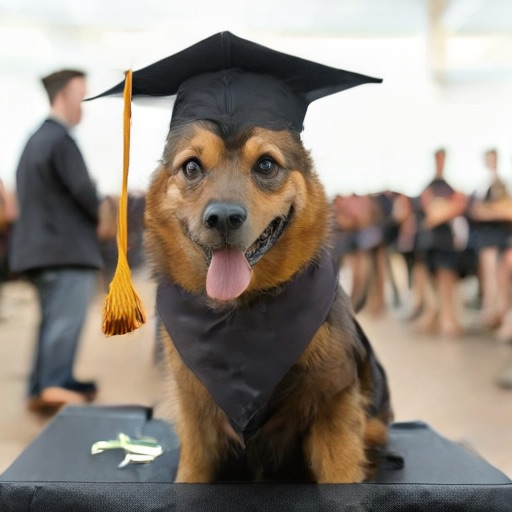} &
        \includegraphics[width=0.21\textwidth]{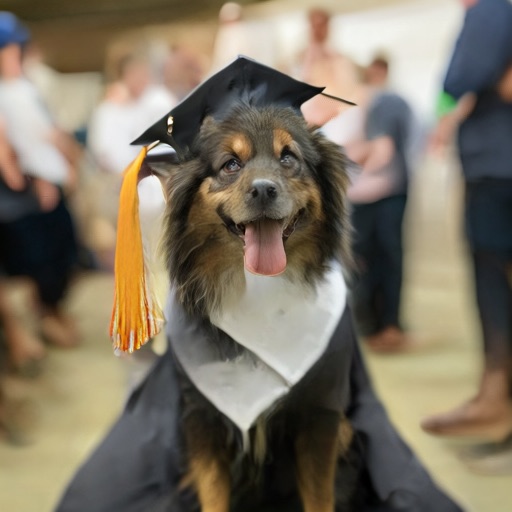} &
        \includegraphics[width=0.21\textwidth]{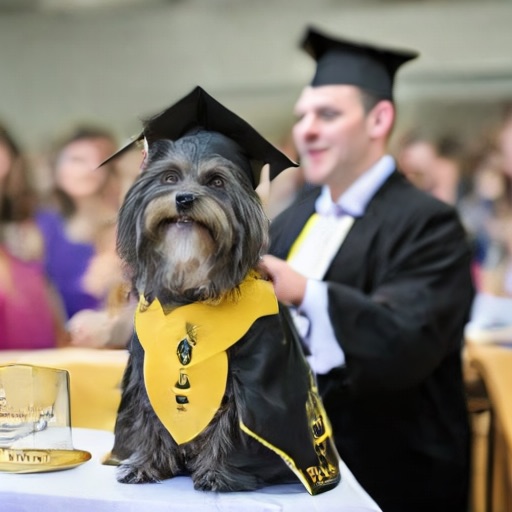} \\
        \multicolumn{4}{c}{\textit{Dog graduation at university.}} \\

        \includegraphics[width=0.21\textwidth]{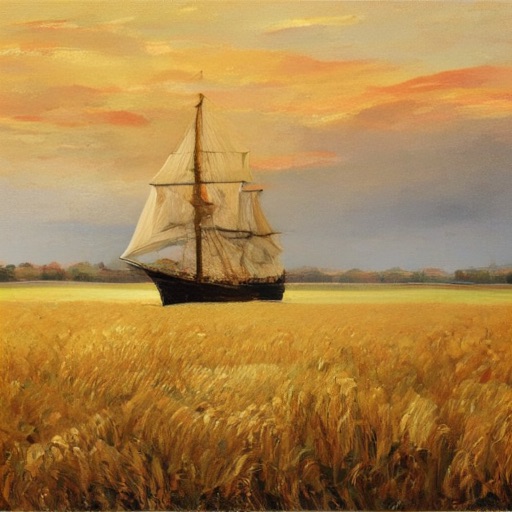} &  
        \includegraphics[width=0.21\textwidth]{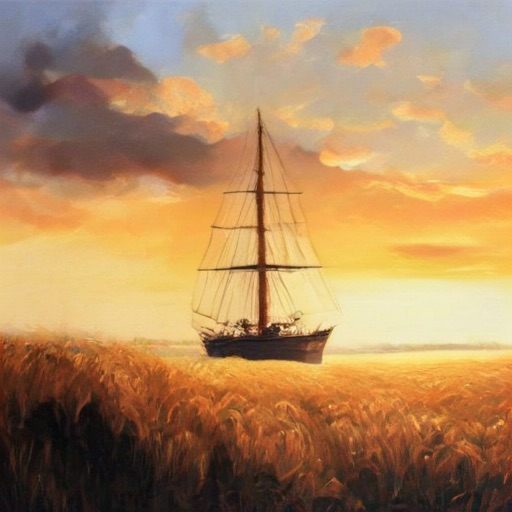} &
        \includegraphics[width=0.21\textwidth]{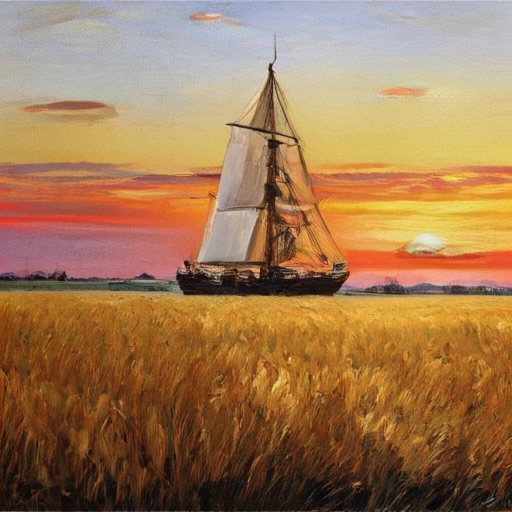} &
        \includegraphics[width=0.21\textwidth]{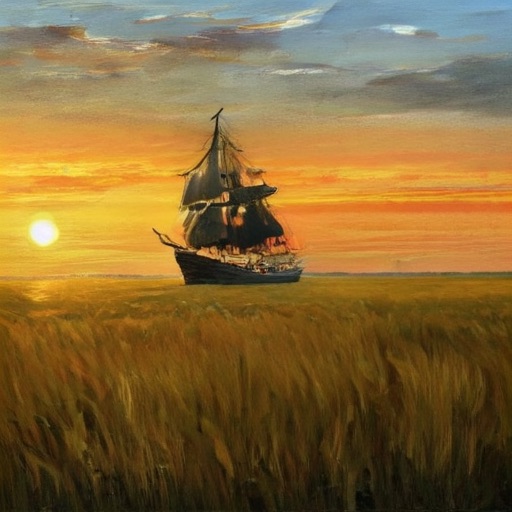} \\
        \multicolumn{4}{c}{\textit{An oil painting of a tall ship sailing through a field of wheat at sunset.}} \\

        \includegraphics[width=0.21\textwidth]{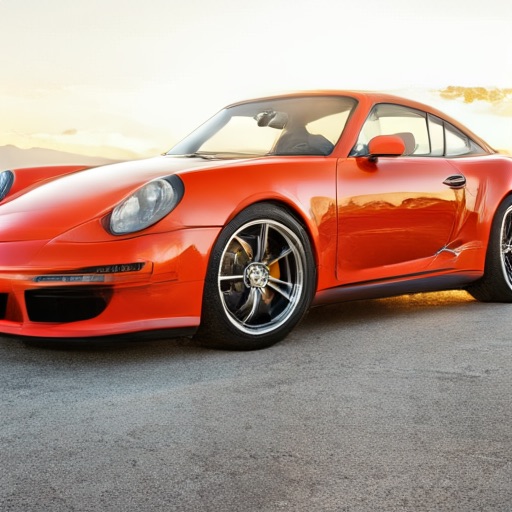} &  
        \includegraphics[width=0.21\textwidth]{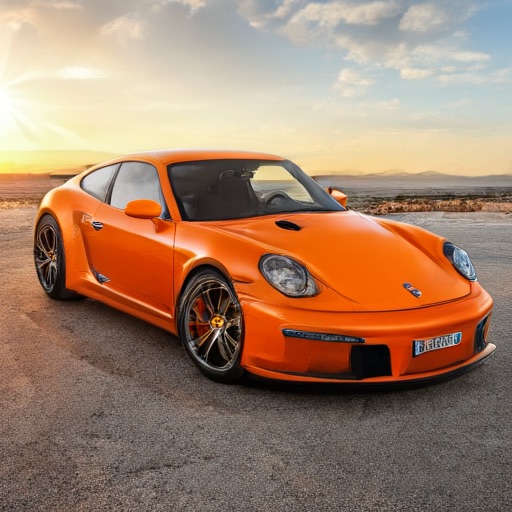} &
        \includegraphics[width=0.21\textwidth]{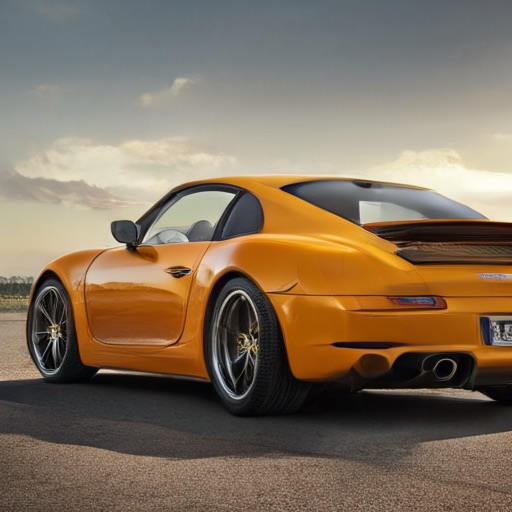} &
        \includegraphics[width=0.21\textwidth]{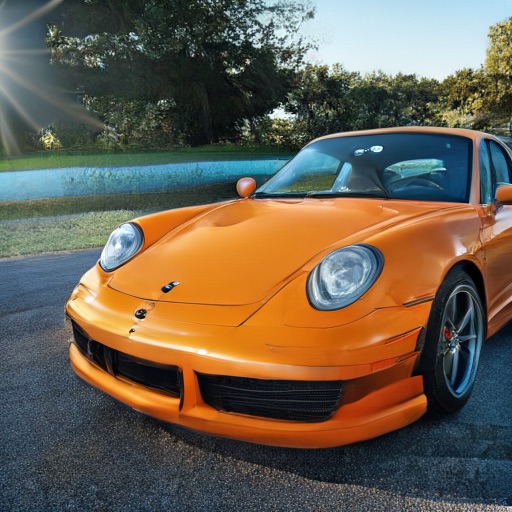} \\
        \multicolumn{4}{c}{\textit{An high-resolution photo of a orange Porsche under sunshine.}} \\

        \includegraphics[width=0.21\textwidth]{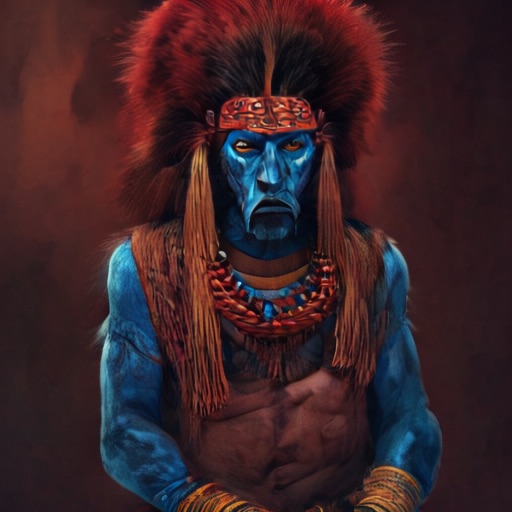} &  
        \includegraphics[width=0.21\textwidth]{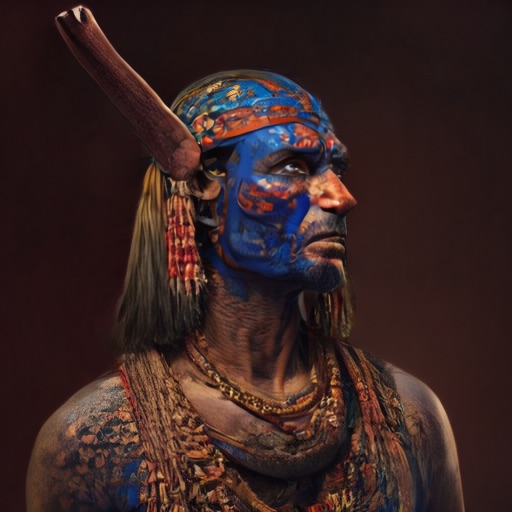} &
        \includegraphics[width=0.21\textwidth]{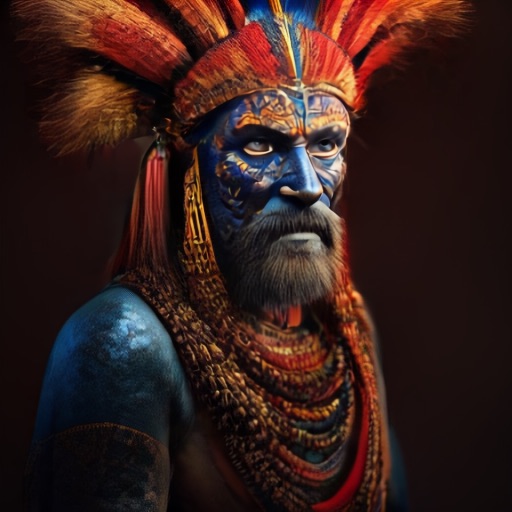} &
        \includegraphics[width=0.21\textwidth]{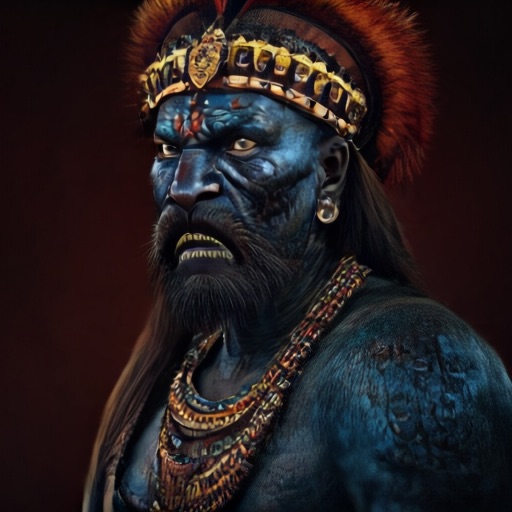} \\
        \multicolumn{4}{c}{\textit{Portrait photo of a Asian old warrior chief, tribal panther make up, blue on red.}} \\

        \includegraphics[width=0.21\textwidth]{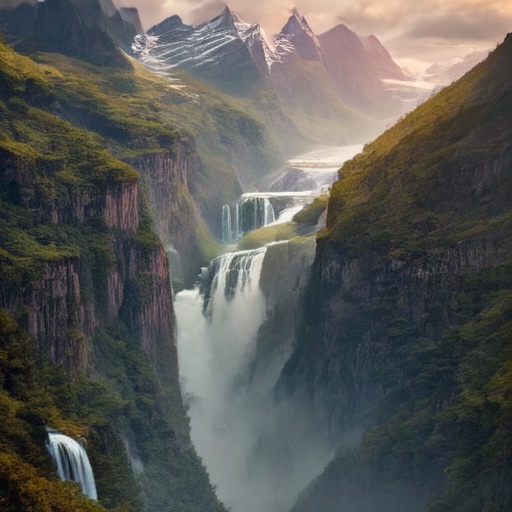} &  
        \includegraphics[width=0.21\textwidth]{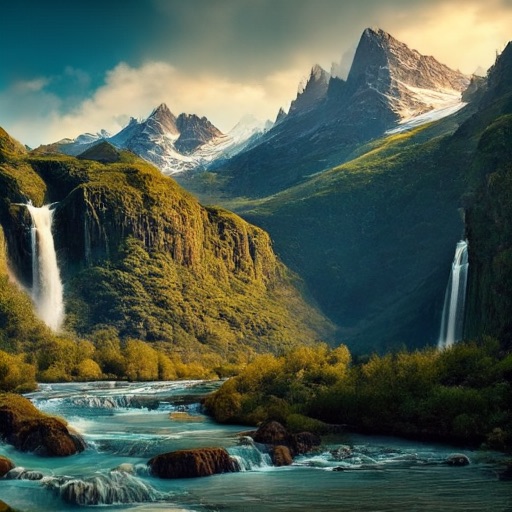} &
        \includegraphics[width=0.21\textwidth]{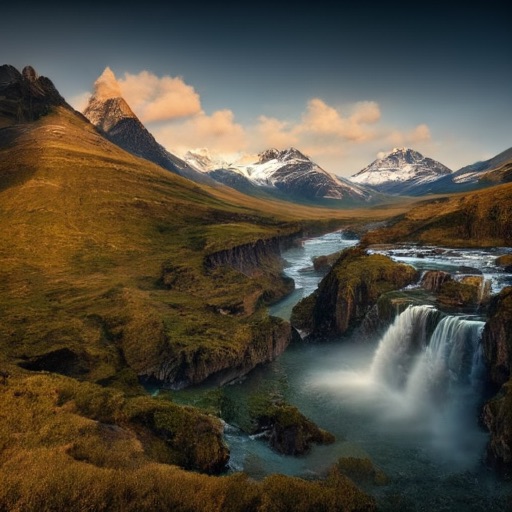} &
        \includegraphics[width=0.21\textwidth]{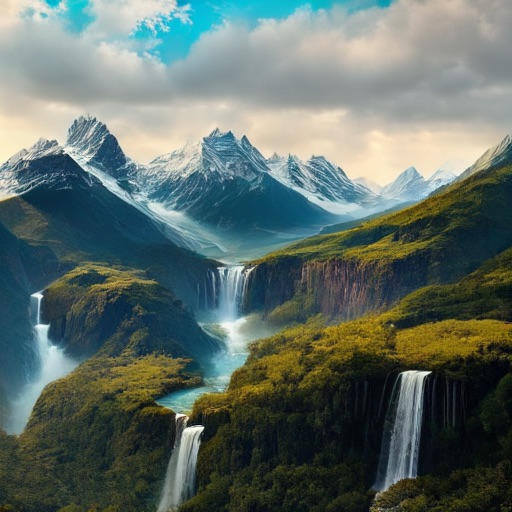} \\ 
        \multicolumn{4}{c}{\textit{A close-up photo of a intricate beautiful natural landscape of mountains and waterfalls.}} \\
    \end{tabular}
    \caption{Additional qualitative results of UFOGen. Zoom-in for better viewing.}
    \label{tab:app_extra_2}
\end{table*}

\begin{table*}[!t]
    \centering
    \begin{tabular}{cccc}
        \includegraphics[width=0.21\textwidth]{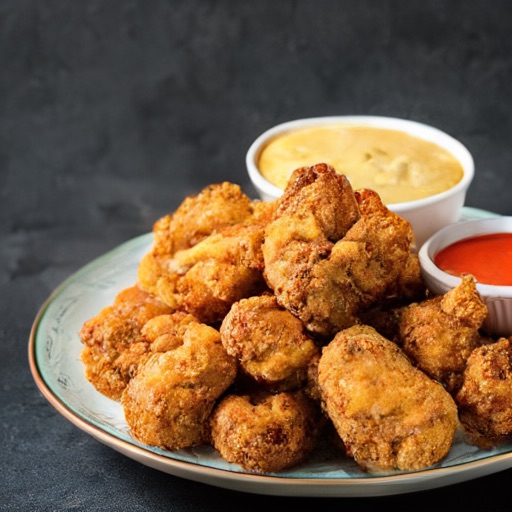} &  
        \includegraphics[width=0.21\textwidth]{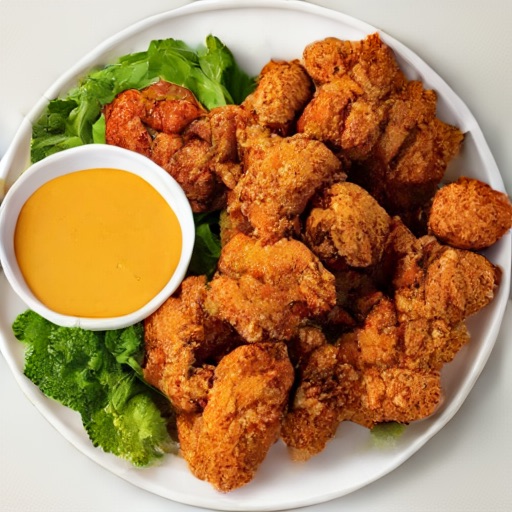} &
        \includegraphics[width=0.21\textwidth]{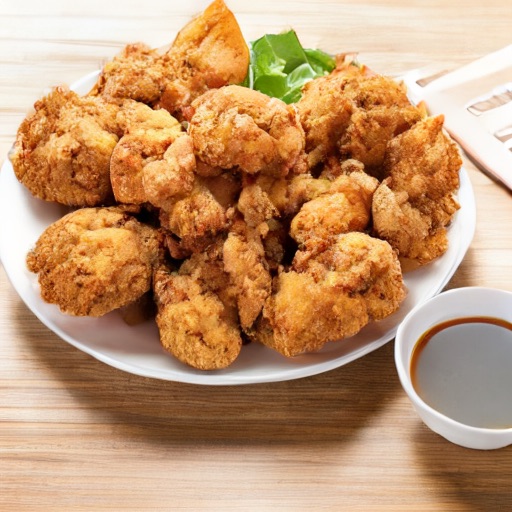} &
        \includegraphics[width=0.21\textwidth]{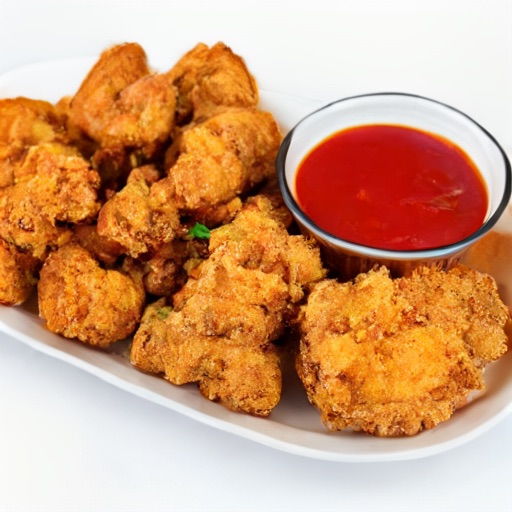} \\
        \multicolumn{4}{c}{\textit{Large plate of delicious fried chicken, with a side of dipping sauce, realistic advertising photo, 4k.}} \\

        \includegraphics[width=0.21\textwidth]{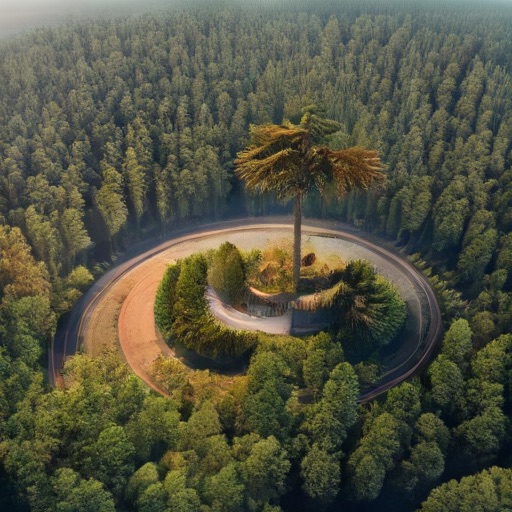} &  
        \includegraphics[width=0.21\textwidth]{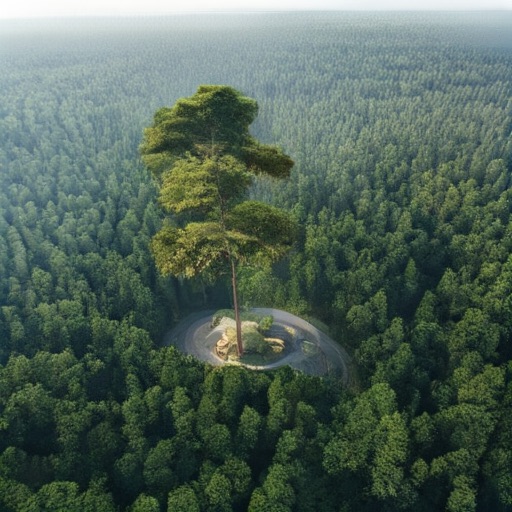} &
        \includegraphics[width=0.21\textwidth]{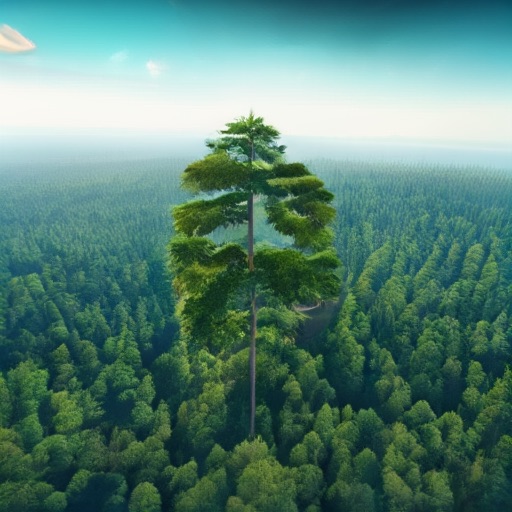} &
        \includegraphics[width=0.21\textwidth]{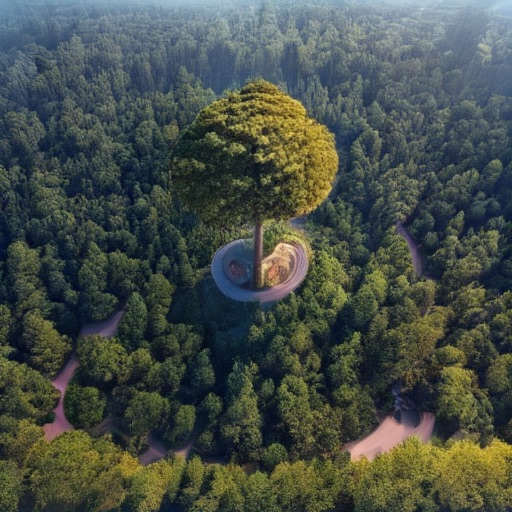} \\
        \multicolumn{4}{c}{\textit{An aerial view of a forest, with a giant tree in the center, realistic render, 4k.}} \\

        \includegraphics[width=0.21\textwidth]{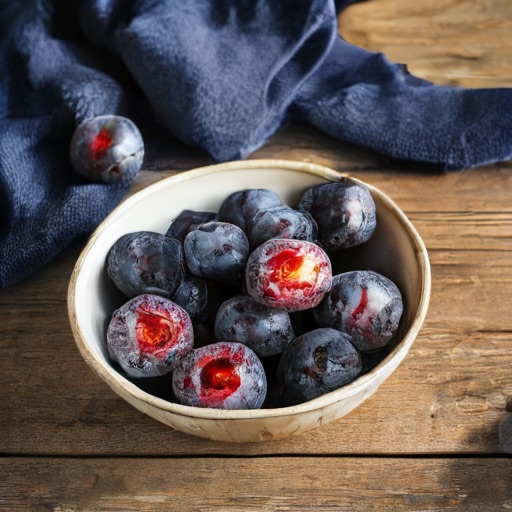} &  
        \includegraphics[width=0.21\textwidth]{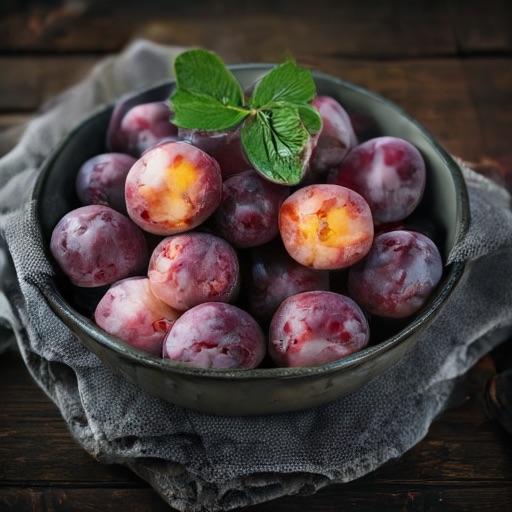} &
        \includegraphics[width=0.21\textwidth]{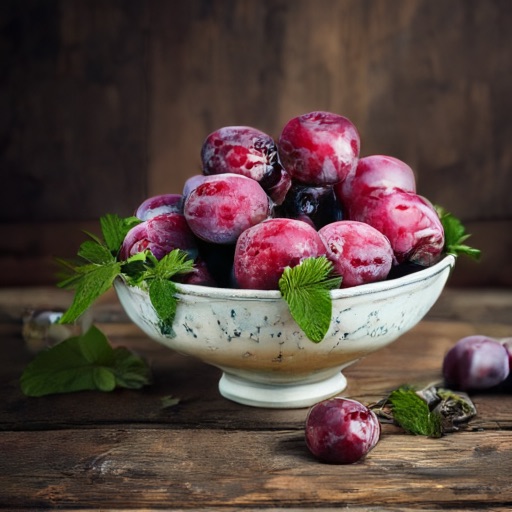} &
        \includegraphics[width=0.21\textwidth]{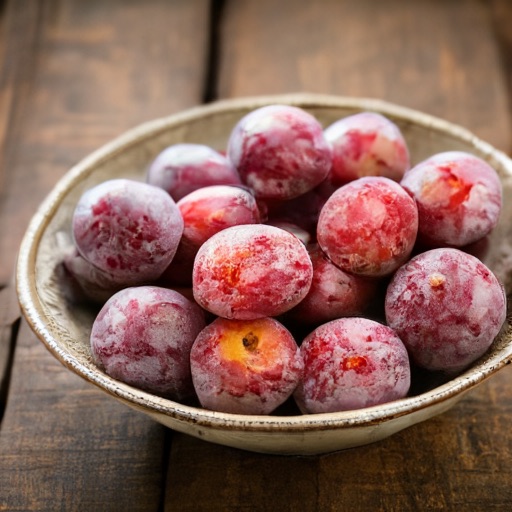} \\
        \multicolumn{4}{c}{\textit{Photo of a bowl filled with plums on a wooden table, volumetric lighting.}} \\

        \includegraphics[width=0.21\textwidth]{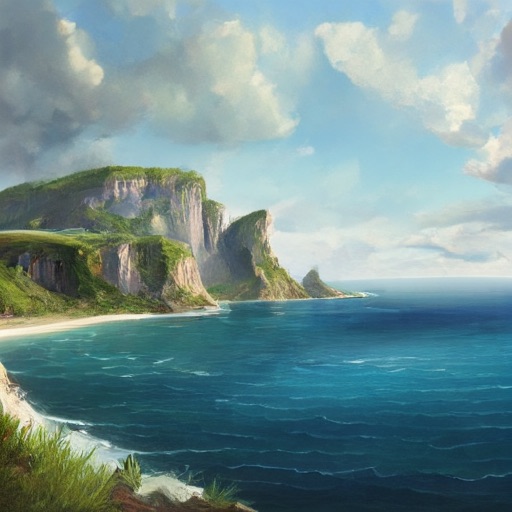} &  
        \includegraphics[width=0.21\textwidth]{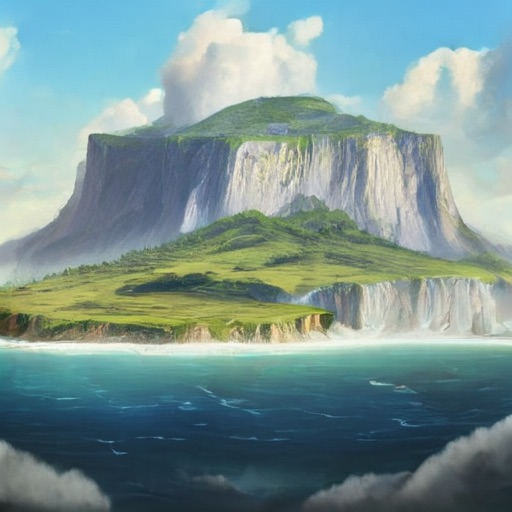} &
        \includegraphics[width=0.21\textwidth]{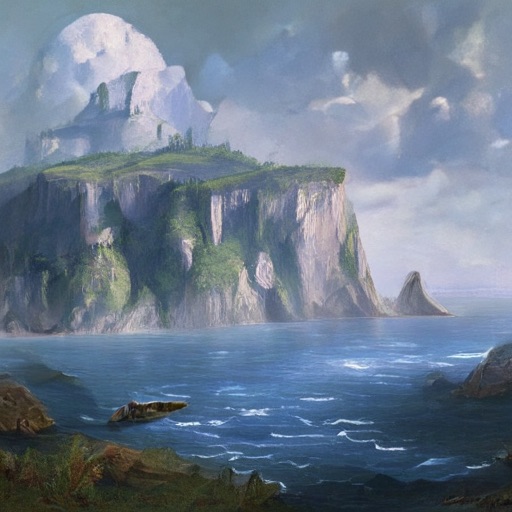} &
        \includegraphics[width=0.21\textwidth]{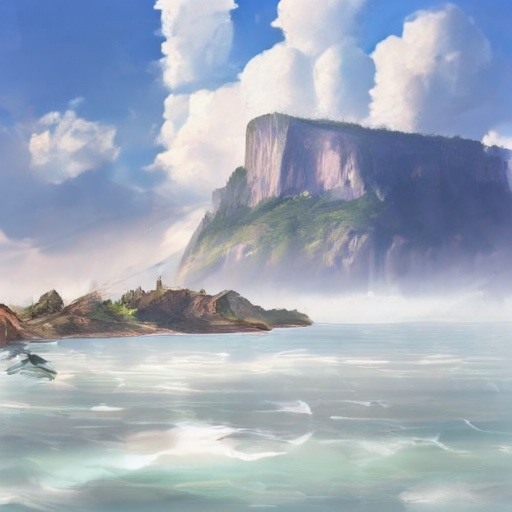} \\
        \multicolumn{4}{c}{\textit{Painting of island and cliff overseeing a vast ocean.}} \\

        \includegraphics[width=0.21\textwidth]{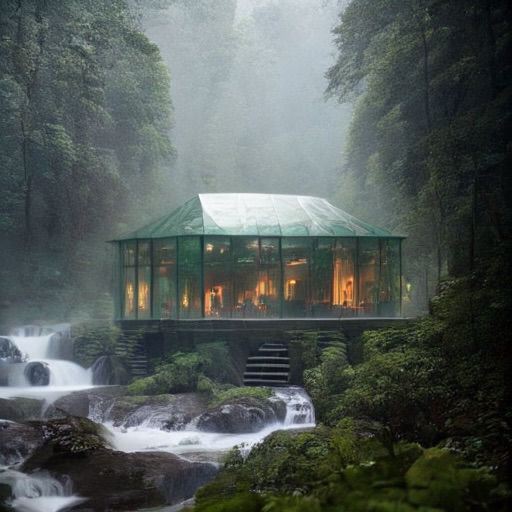} &  
        \includegraphics[width=0.21\textwidth]{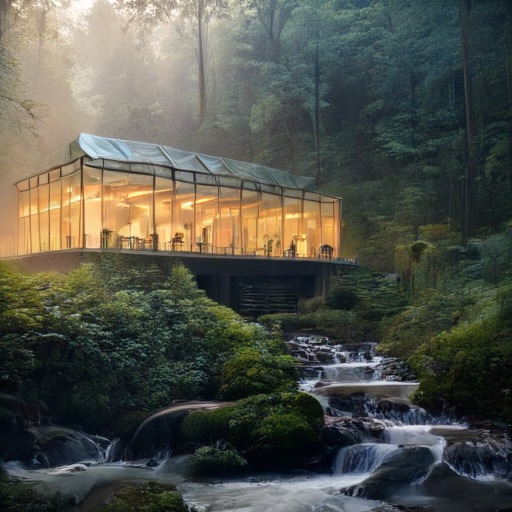} &
        \includegraphics[width=0.21\textwidth]{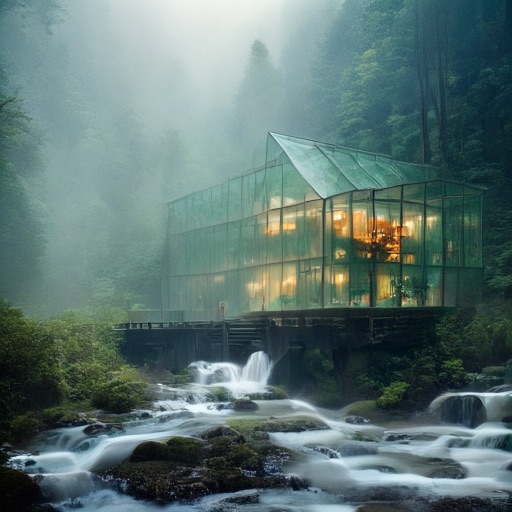} &
        \includegraphics[width=0.21\textwidth]{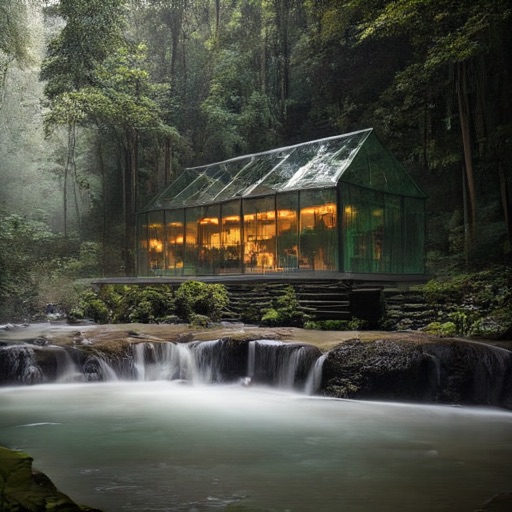} \\ 
        \multicolumn{4}{c}{\textit{Photo of a modern glass house in the jungle, small stream flowing, mist, atmospheric.}} \\
    \end{tabular}
    \caption{Additional qualitative results of UFOGen. Zoom-in for better viewing.}
    \label{tab:app_extra_3}
\end{table*}

\begin{table*}[!t]
    \centering
    \begin{tabular}{cccc}
        \includegraphics[width=0.21\textwidth]{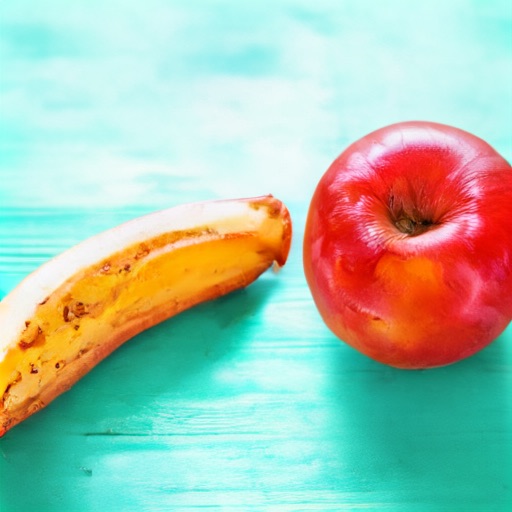} &  
        \includegraphics[width=0.21\textwidth]{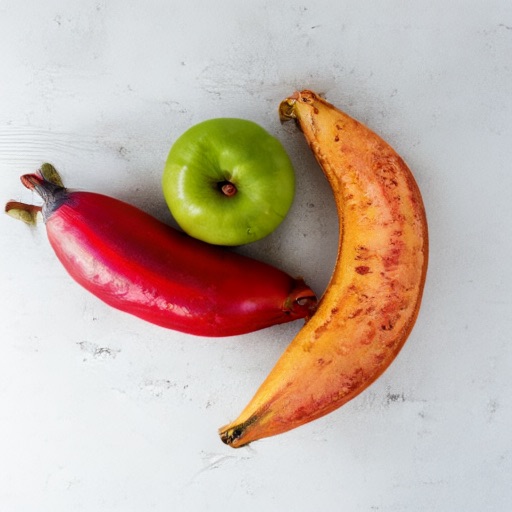} &
        \includegraphics[width=0.21\textwidth]{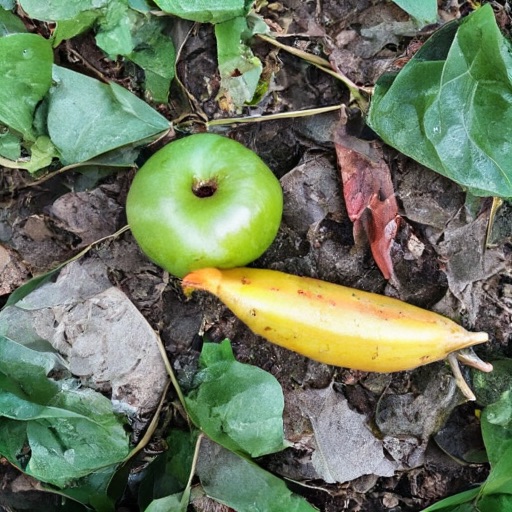} &
        \includegraphics[width=0.21\textwidth]{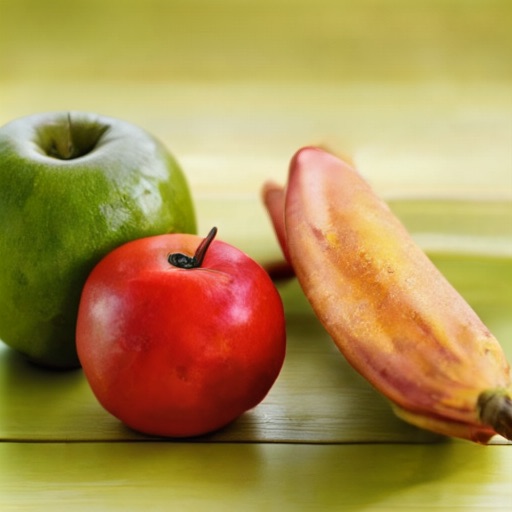} \\
        \multicolumn{4}{c}{\textit{A green apple and a red banana.}} \\

        \includegraphics[width=0.21\textwidth]{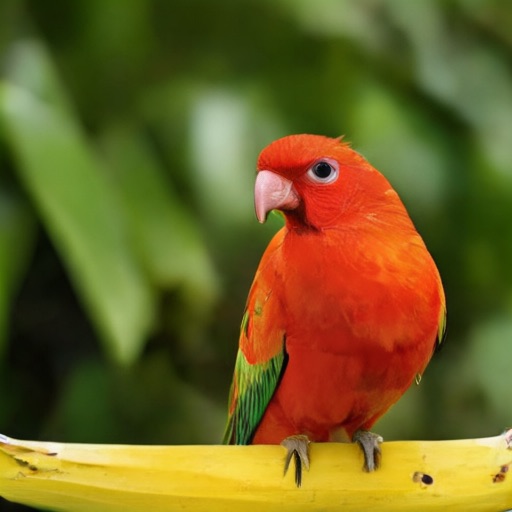} &  
        \includegraphics[width=0.21\textwidth]{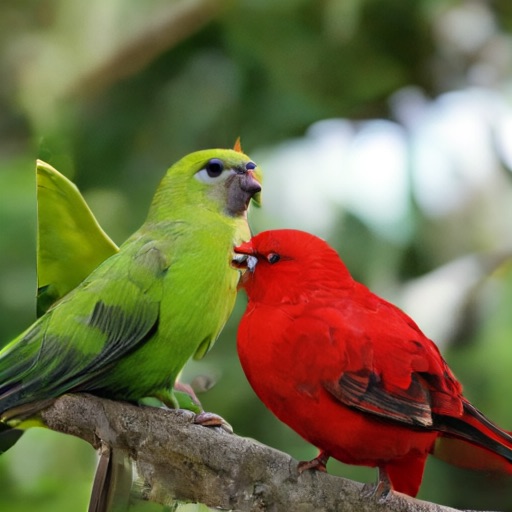} &
        \includegraphics[width=0.21\textwidth]{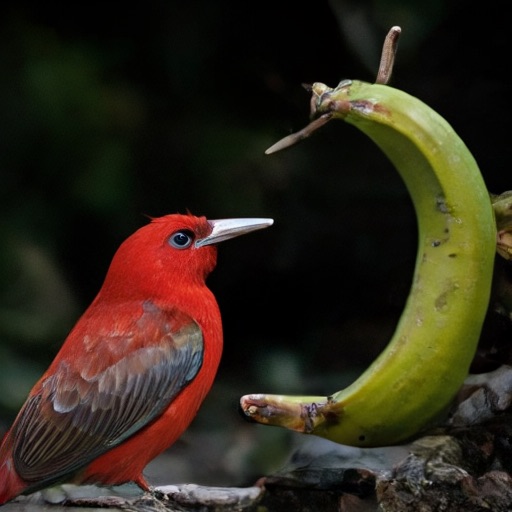} &
        \includegraphics[width=0.21\textwidth]{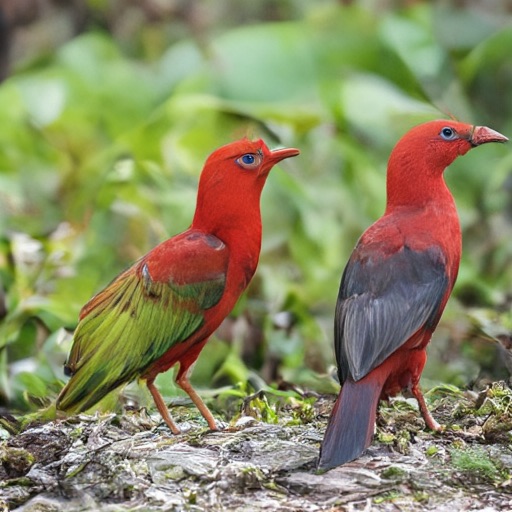} \\
        \multicolumn{4}{c}{\textit{A red bird and a green banana.}} \\

        \includegraphics[width=0.21\textwidth]{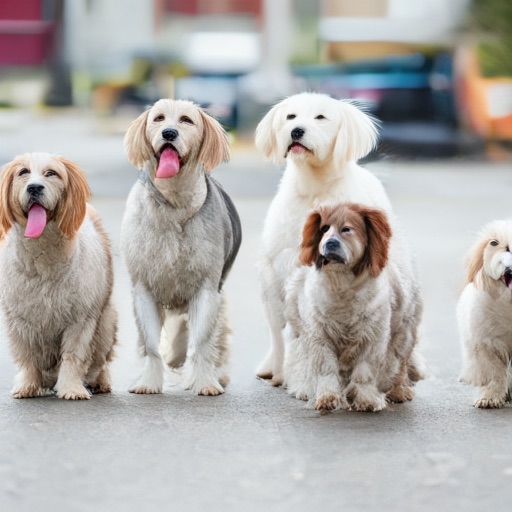} &  
        \includegraphics[width=0.21\textwidth]{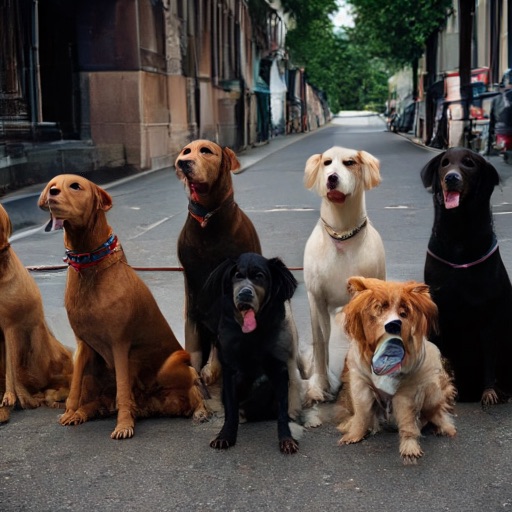} &
        \includegraphics[width=0.21\textwidth]{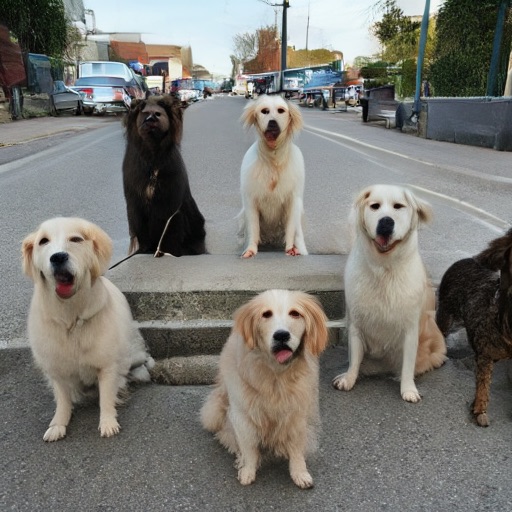} &
        \includegraphics[width=0.21\textwidth]{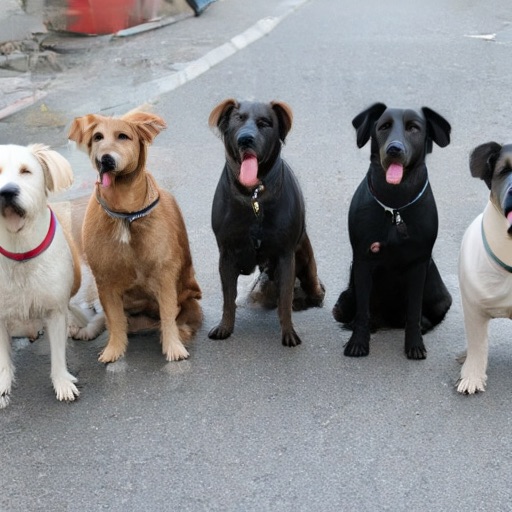} \\
        \multicolumn{4}{c}{\textit{Four dogs on the street.}} \\

    \end{tabular}
    \caption{Failure cases of UFOGen.}
    \label{tab:app_fail}
\end{table*}

\begin{figure*}[!t]
    \centering
    \includegraphics[width=0.9\textwidth]{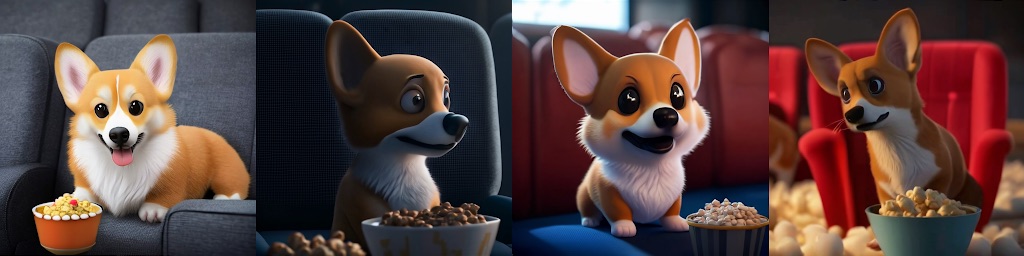} \\
    \textit{Cute small \underline{corgi} sitting in a movie theater eating popcorn, unreal engine.} \\
    \includegraphics[width=0.9\textwidth]{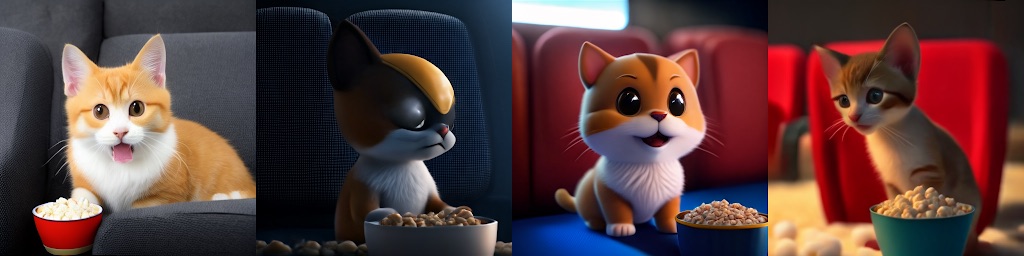} \\
    \textit{Cute small \underline{cat} sitting in a movie theater eating popcorn, unreal engine.} \\
    \includegraphics[width=0.9\textwidth]{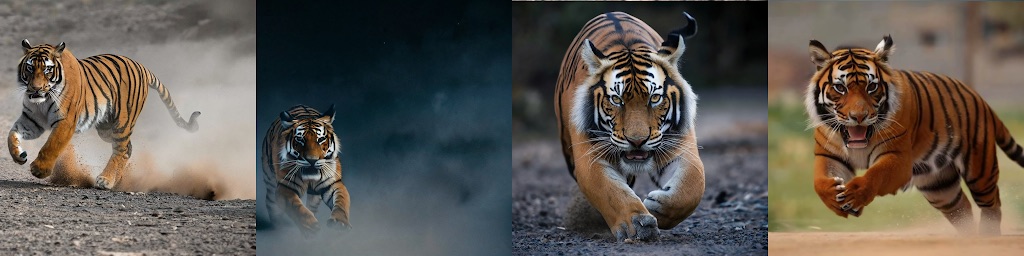} \\
    \textit{A \underline{tiger} is running.} \\
    \includegraphics[width=0.9\textwidth]{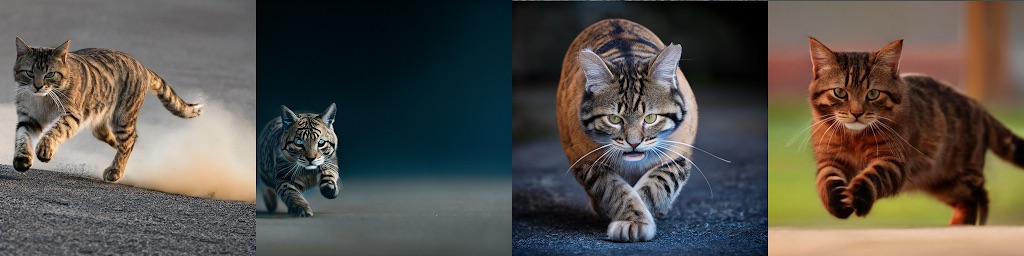} \\
    \textit{A \underline{cat} is running.} \\
    \caption{Extended results of image-to-image generation from UFOGen. For each group, we edit the images by adding noise and sightly modifying the prompt.}
    \label{fig:i2i_ext}
\end{figure*}

\begin{figure*}
    \centering
    \includegraphics[width=0.99\textwidth]{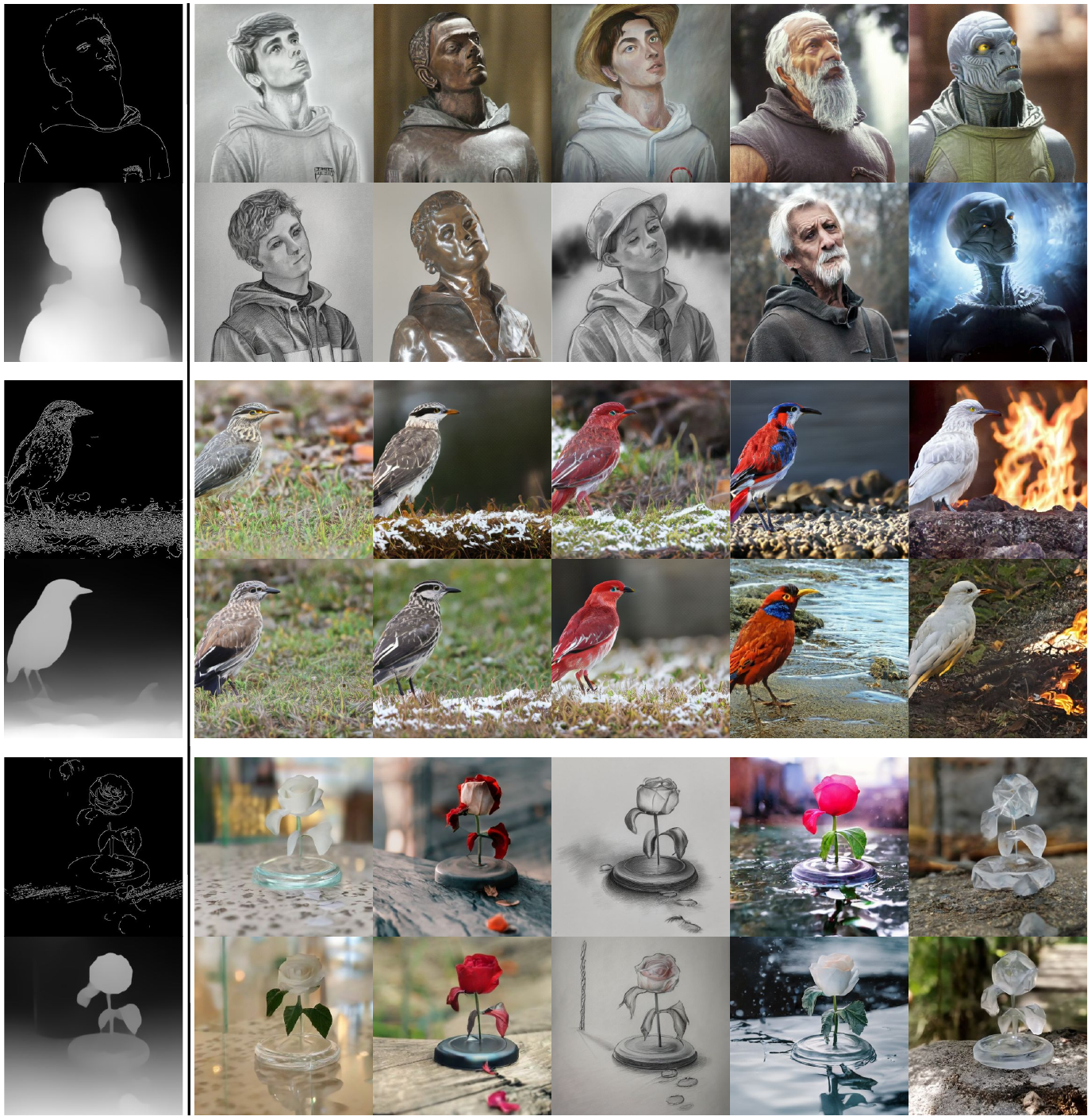}
    \caption{Extended results of controllable generation from UFOGen. For each group of canny edge and depth map images, we use same prompts per column.}
    \label{fig:plugin_ext}
\end{figure*}